\pdfoutput=1
\documentclass[11pt]{article}

\usepackage[preprint]{acl}

\usepackage{inconsolata}
\usepackage{times}
 
\usepackage{booktabs}
\usepackage{multirow}
\usepackage{tablefootnote}
\usepackage{array}
\usepackage{pdfpages}
\usepackage{colortbl}

\usepackage{amsmath}
\usepackage{amsfonts}

\usepackage{graphicx}
\usepackage{float}
\usepackage{subcaption}
\usepackage{paralist}
\usepackage{enumitem}
\usepackage{titlesec}

\usepackage[T1]{fontenc}
\usepackage[utf8]{inputenc}

\usepackage{microtype}

\usepackage{inconsolata}

\usepackage{longtable}
\usepackage{tabularx}

\newcommand{\ours}{\textcolor{black}{MiCEval {}}}

\title{MiCEval: Unveiling Multimodal Chain of Thought's Quality via Image Description and Reasoning Steps}

\author{
    Xiongtao Zhou$^1$\footnotemark[1] \quad 
    Jie He$^2$\thanks{\ \ Equal Contribution.} \quad 
    Lanyu Chen$^1$ \quad 
    Jingyu Li$^4$ \quad 
    \textbf{Haojing Chen}$^4$ \\
    \textbf{Víctor Gutiérrez-Basulto}$^3$\footnotemark[2] \, 
    \textbf{Jeff Z. Pan}$^2$\footnotemark[2] and \; 
    \textbf{Hanjie Chen}$^5$\thanks{ \, Corresponding author} \\
    $^1$ Waseda University, Japan  $^2$ School of Informatics, University of Edinburgh, UK  \\
    $^3$ School of Computer Science and Informatics, Cardiff University, UK \\
    $^4$ Independent  Researcher $^5$ Rice University, US\\
    \normalsize{\texttt{alenai.tao@ruri.waseda.jp, j.he@ed.ac.uk}} \\
    \normalsize{\texttt{j.z.pan@ed.ac.uk, hanjie@rice.edu}} \\
}

\begin{document}
\maketitle

\begin{abstract}

%\vic{Yes, the title is way too long and convoluted. You dont need to describe every single contribution}
%\victor{A title might be: {\bf On the Boundaries of Multimodal Chain of Thought}. Yes, the title does not give details of all contributions, but that is why we have the abstract and introduction }

% \xt{Multimodal Chain of Thought (MCoT) is a popular prompting strategy that enhances the performance of Multimodal Large Language Models (MLLMs) across various downstream tasks. However, the quality of MCoT answers of MLLMs remains unclear, without an automatic evaluation for granular assessment. To fill in this gap, we introduce \textsc{RGV}: \textbf{R}easoning \textbf{G}ranularity \textbf{V}erification, a framework that evaluates CoT answer by decomposing it into: (1) \textit{Description Step} and (2) \textit{Reasoning Step}, assessing both based on their correctness. The \textit{Description Step} focuses on describing the image, while the \textit{Reasoning Step} involves making inferences beyond the image by incorporating information from previous steps. We develop a comprehensive, multilevel MCoT dataset, with annotations for each step in terms of Correctness, Relevance and Informativeness. Extensive experiments with \textsc{RGV} on four state-of-the-art MLLMs demonstrate that, compared to existing  cosine-similarity based methods and fine-tuning based methods, step-wise evaluations of MLLM-generated answers better align with human judgements.}

\textbf{M}ultimodal \textbf{C}hain \textbf{o}f \textbf{T}hought (MCoT) is a popular prompting strategy for improving the performance of \textbf{m}ultimodal \textbf{l}arge \textbf{l}anguage \textbf{m}odels (MLLMs) across a range of complex reasoning tasks. Despite its popularity, there is a notable absence of automated methods for evaluating the quality of reasoning steps in MCoT. To address this gap, we propose \textbf{M}ult\textbf{i}modal \textbf{C}hain-of-Thought \textbf{Eval}uation (MiCEval), a  framework designed to assess the correctness of reasoning chains by evaluating the quality of both the \emph{description} and each \emph{reasoning step}. The evaluation of \emph{description} component focuses on the accuracy of the image descriptions, while the \emph{reasoning step} evaluates the quality of each step as it is conditionally generated based on the preceding steps. \ours is built upon a fine-grained dataset with annotations that rate each step according to \emph{correctness}, \emph{relevance}, and \emph{informativeness}. Extensive experiments on four state-of-the-art MLLMs show that step-wise evaluations using \ours align more closely with human judgments compared to existing methods based on cosine similarity or fine-tuning approaches. MiCEval datasets and code can be found in 
% \href{https://anonymous.4open.science/r/MiCEval-847F/README.md}{https://anonymous\_github/MicEval}.
\href{https://github.com/alenai97/MiCEval}{https://github.com/alenai97/MiCEval}.

\end{abstract}

\section{Introduction}
%\xt{Multimodal Chain-of-Thought (MCoT) enhances the performance of Multimodal Large Language Models (MLLMs) on downstream tasks by generating multi-step reasoning chain that culminates in an MCoT answer \cite{zhang2024multimodalchainofthoughtreasoninglanguage,NEURIPS2023_617ff527,chen-etal-2024-m3cot,shao2024visual}. These MCoT answers not only improve the interpretability of MLLMs but also provide a novel approach for evaluating their reasoning capabilities. In most existing works \cite{lu2022learn,zhang2024multimodalchainofthoughtreasoninglanguage,chen-etal-2024-m3cot}, the evaluation typically focuses solely on the correctness of the final prediction within the MCoT answer, neglecting the integrity of the entire reasoning chain. This one-step evaluation method, based on the final prediction, overlooks mistakes and unreliable information present in the reasoning chain \cite{lyu-etal-2023-faithful,NEURIPS2023_ed3fea90}, making it inadequate. To enhance the credibility of the evaluation and accurately reflect the quality of MCoT, it is necessary to assess each step of the MCoT answer.}

\textbf{M}ultimodal \textbf{c}hain-\textbf{o}f-\textbf{t}hought (MCoT) enhances the performance of \textbf{m}ultimodal \textbf{l}arge \textbf{l}anguage \textbf{m}odels (MLLMs) on complex reasoning tasks by generating explicit multi-step reasoning chains to achieve the final goal \cite{zhang2024multimodalchainofthoughtreasoninglanguage,NEURIPS2023_617ff527,chen-etal-2024-m3cot,shao2024visual}. %This approach 
It not only improves the interpretability of MLLM outputs but also offers a valuable framework for assessing their reasoning capabilities. Most existing evaluations \cite{lu2022learn,zhang2024multimodalchainofthoughtreasoninglanguage,chen-etal-2024-m3cot} focus exclusively on the correctness of the final prediction in MCoT, while overlooking the quality of the reasoning process within the chains, %This approach, which treats the reasoning chain as a whole, 
failing to capture errors and unreliable steps that may occur throughout the chain \cite{lyu-etal-2023-faithful,NEURIPS2023_ed3fea90}. To establish a principled framework to more accurately reflect the quality of MCoT, it is essential to evaluate each step in the reasoning chain individually.

\begin{figure}[!tpb]
    \centering
    \includegraphics[width=0.9\linewidth]{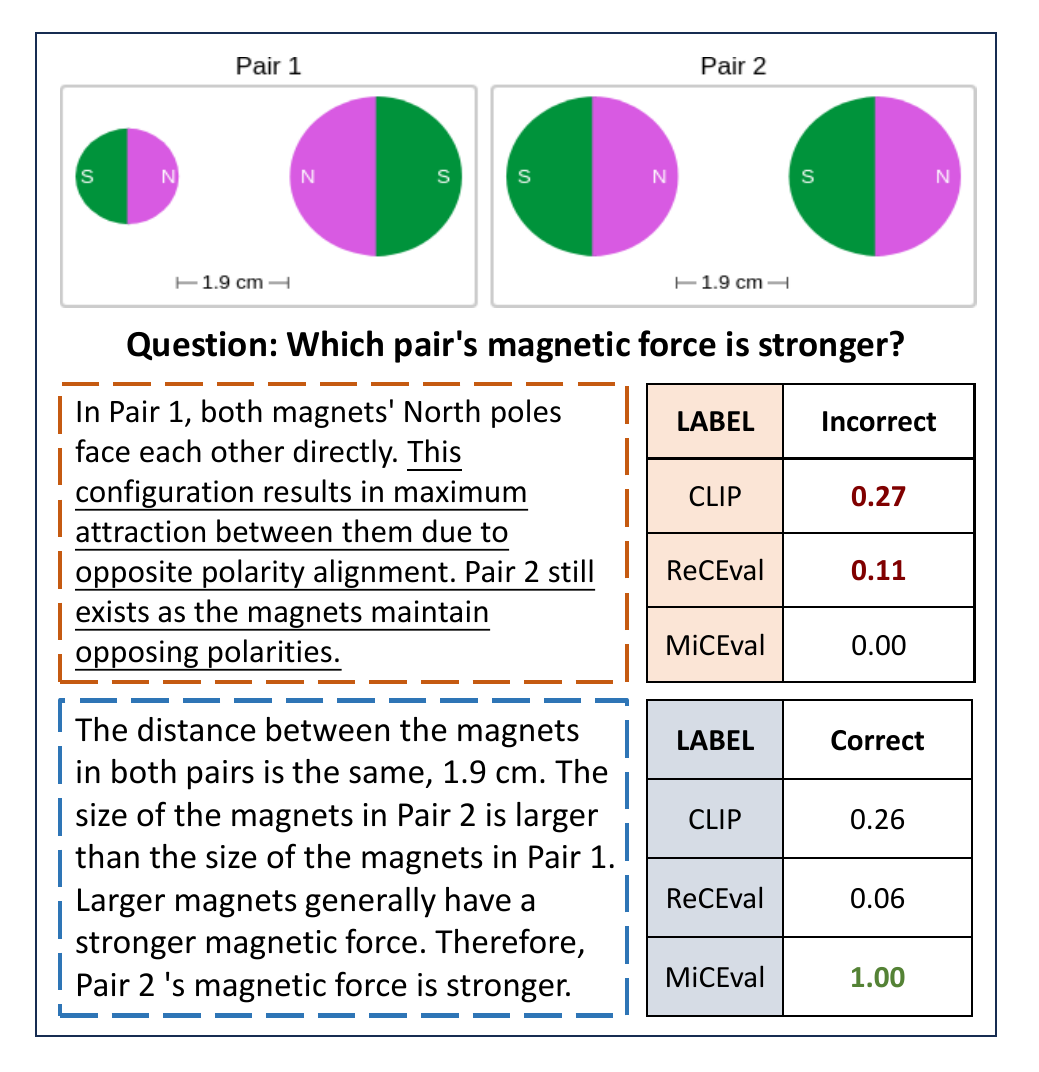}
    \caption{%We introduce MiCEval, a framework proposed to evaluate the quality of MCoT answer.
    We exemplify how CLIP and ReCEval did not choose the correct MCoT answer from the two model-generated MCoT answers, but MiCEval succeeded.}
    \label{fig:intro}
    \vspace{-0.6cm}
\end{figure}

A common approach for evaluating (unimodal) CoT is to compare model-generated reasoning chains with human-written reference chains \cite{clinciu-etal-2021-study,NEURIPS2022_1fc548a8,SaparovHe2023}. However, this method incurs substantial costs. As a result, recent research has shifted towards reference-free evaluation methods, fine-tuning models on human-annotated reasoning chain datasets \cite{golovneva2022roscoe,prasad-etal-2023-receval}. Yet, these methods often fall short in delivering robust evaluation metrics due to the inherent limitations of fine-tuning.  The use of \textbf{l}arge \textbf{l}anguage \textbf{m}odels (LLMs) as evaluators has emerged as a promising alternative \cite{he-etal-2024-socreval,xia_evaluating} %. However, note that these techniques have been developed specifically for CoT and 
but they cannot be directly applied to MCoT, as they are unable to effectively evaluate the image descriptions  in MCoT; e.g., %As shown in the first step of the first MCoT answer 
in Figure~\ref{fig:intro}, the statement \emph{``In Pair 1, both magnets' North poles face each other directly''} is an image description. Methods that focus solely on the textual modality cannot properly evaluate  the  correctness of image descriptions. In the multimodal domain, several evaluation metrics have been proposed to assess the alignment between images and descriptions by calculating their cosine similarity \cite{hessel-etal-2021-clipscore,li2023blip2}. However, their effectiveness in evaluating MCoT remains unexplored.

% \textcolor{red}{which not only involves reasoning steps but also includes image descriptions}.

% \victor{ Red Text  paragraph above: Ideally it would be good to more precisely say what the challenge of having image descriptions is . E.g. at each do reasoning and image description steps interact? The correctness of one depends on the correctness of the other? }

%\xt{MCoT answer involves both image descriptions and corresponding reasoning. Only if all steps are correct, the reasoning chain is correct \cite{li-etal-2023-making}. Therefore, a correct MCoT, all steps about image description should be correct and relevant to answering the question, while all steps contain reasoning should also be correct, relevant, and helpful in arriving at the final answer. Therefore, a method capable of evaluating both visual and language characteristics provides a more accurate reflection of the overall quality of MCoT. To conduct the comprehensive step-level evaluation, we need a multilevel MCoT verifier evaluation dataset similar to the one released by \citet{jacovi-etal-2024-chain}. A comprehensive MCoT verifier evaluation dataset will not only provide a detailed analysis of the ability of MCoT verifier, but also for the quality of MCoT.}

The answers generated by MCoT involve both image descriptions~\cite{CCZG+2023,LGFZ+2023} and their corresponding reasoning steps. Therefore, for an MCoT to be correct, all steps related to both image descriptions and reasoning must be accurate, relevant, and contribute effectively to reaching the final answer. This highlights the need for a method that evaluates both the visual and textual components, providing a comprehensive assessment of the overall quality of the MCoT output. 
%\xt{\citet{ChenCZWLZZ00024} used MLLMs as evaluators to assess their correlation with human judgments, which also serves as an evaluation of MCoT quality.} 
As noted in the context of CoT by \citet{jacovi-etal-2024-chain}, the creation of a multi-step evaluation dataset is essential for accurately assessing the abilities of MLLMs as MCoT verifiers. Such a comprehensive dataset will not only enable detailed evaluation of an MCoT verifier's effectiveness but will also offer insights into the overall quality of MCoT outputs.

%\xt{We present \textbf{R}easoning \textbf{G}ranularity \textbf{V}erification (\textsc{RGV}), a framework that evaluates MCoT by breaking MCoT answer down into (1) Description Step and (2) Reasoning Step, and assesses both for correctness, integrated with step-level human-annotated fine-grained data. \textsc{RGV} decomposes MCoT verification into five basic tasks from both visual and textual modalities: description correctness, image relevance, logic correctness, logic relevance, and informativeness, providing a comprehensive and in-depth evaluation of MCoT and its verifiers. \textsc{RGV} also includes a dataset with 903 MLLM-generated MCoT answers and 2,889 human annotated MCoT steps, which are utilized for our comprehensive step-level evaluations on MCoT and its verifiers.}

In this paper, we propose MiCEval, a framework designed to evaluate MCoT by breaking down MCoT chains into \emph{description} and \emph{reasoning} steps, and assessing the quality of each. \ours decomposes MCoT verification into five core tasks across both visual and textual modalities: description correctness, image relevance, logical correctness, logical relevance, and informativeness. Additionally, \ours introduces a dataset comprising 903 MLLM-generated MCoT answers and 2,889 human-annotated MCoT steps, which serve as the foundation for our comprehensive step-wise evaluations of both MCoT and its verifiers.

%\xt{Each MCoT step in \textsc{RGV} dataset is first labeled as either a “Description Step”, a “Reasoning Step”, or “Both Step” (an integration of description and reasoning). For the description steps, we provide unique annotations for correctness and relevance, whereas for the reasoning steps, we collect labels for correctness, relevance, and informativeness. Additionally, we further annotated the error types for steps classified as description incorrect and logic incorrect. At the MCoT-level, we annotated whether the MCoT answer is fully correct.}

Each MCoT step in the \ours dataset is initially categorized as either a “Description Step”, a “Reasoning Step”, or a “Both Step” (a combination of description and reasoning). For the description steps, we provide fine-grained annotations for correctness and relevance, while for the reasoning steps, we collect labels for (1) correctness: each step is generated based on previously valid information; (2) relevance: every step is related to either the image or answering the question; and (3) informativeness:  steps should provide new, relevant information. Furthermore, we annotate the specific error types for steps identified as description incorrect or logically incorrect. At the MCoT level, we annotate whether the overall MCoT answer is correct.

% \hanjie{we should briefly define these properties}
%\xt{Benefiting from the rich annotation tasks in the \textsc{RGV} dataset, \textsc{RGV} offers a wide range of evaluation dimensions. From the MCoT perspective, \textsc{RGV} not only supports step-level reasoning granularity MCoT evaluation but also enables the assessment of overall quality of MCoT. On the other hand, \textsc{RGV} enables a fine-grained evaluation of the reasoning capabilities of MLLMs. In our work, we primarily focus on two dimensions: (1) Evaluate the ability of MLLMs as verifiers, and (2) assess the correlation of \textsc{RGV} correctness metrics with human preferences.}

Leveraging the diverse annotation tasks in the \ours dataset, the \ours framework provides a comprehensive range of evaluation dimensions. From the MCoT perspective, \ours not only facilitates step-level reasoning evaluation but also assesses the overall quality of MCoT. Additionally, \ours enables a detailed, fine-grained evaluation of the reasoning capabilities of MLLMs. In this work, we focus primarily on evaluating: (1) the effectiveness of MLLMs as verifiers, and (2) the correlation between \ours correctness metrics and human preferences.

Our \textbf{main contributions}  are  as follows: 
% \begin{enumerate}
%     \item We propose a comprehensive step-level annotation protocol, and based on this protocol, we constructed  a high-quality, fine-grained, human-annotated MCoT dataset. 
%     \item Evaluating the MLLM verifier's performance on each fine-grained task to identify specific weaknesses, such as MLLMs perform poorly on complex reasoning tasks, highlighting a clear direction for improvement.
%     \item We propose \textsc{RGV} correctness metrics for MCoT answers, which evaluate MCoT answer from multiple dimensions. Furthermore, our research shows that \textsc{RGV} enables MLLM evaluator to align more closely with human performance, providing an evaluation approach for MCoT answer.
% \end{enumerate}
\begin{compactenum} \item We propose a comprehensive step-level annotation protocol to construct a high-quality, fine-grained, human-annotated MCoT dataset. \item We assess MLLM verifiers' capabilities on each fine-grained task, identifying specific weaknesses, such as poor performance on complex reasoning tasks. \item We introduce the MiCEval multidimensional correctness metrics for evaluating MCoT answers. Moreover, we demonstrate that MiCEval enables MLLM evaluators to better align with human judgment. \end{compactenum}

\section{Problem Definition and Formalization}
\label{sec2}
In this section, we define MCoT and formalize fine-grained tasks for validating and evaluating MCoT depending on the type of steps. An MCoT consists of an image $\mathcal I$, a question $\mathcal Q$, and a reasoning chain $\mathcal{RC} = r_1, ..., r_n$:
\[\textit{MCoT} = \textit{Prompt} (\mathcal{I}, \mathcal{Q}, \mathcal{RC})\]
 \textit{Prompt()} represents the transformation of multimodal inputs into task-specific instruction formats. The main difference between of CoT and MCoT is the inclusion of a visual input in the latter. In our study, we focus on \textit{MCoT answer}, namely the chain $\mathcal{RC}$ generated from  $\mathcal I$ and $\mathcal Q$. Each step $r_i, \, i \in [1,n]$, is a complete sentence, and $r_n$ is the final step that contains the prediction corresponding to the question.

% We define MCoT and formalize fine-grained tasks for validating and evaluating MCoT based on type of step in this section. An MCoT consists of an image $I$, a question $Q$, and a reasoning chain $RC$:
% \[\textit{MCoT} = \textit{Prompt} (\text{I}, \text{Q}, \text{RC})\]
%  \textit{Prompt}() represents the transformation of multimodal inputs into task-specific instruction formats. Compared to CoT, the main distinction of MCoT is the inclusion of the visual modality input. In our study, we focus on \textbf{MCoT answer}, namely the $RC$ generated based on the image and question of MCoT. $RC$ is composed of steps $r_1$, $r_2$, ..., $r_n$, with each step being a complete sentence, and $r_n$ being the final step that contains the prediction corresponding to the question.

\begin{figure*}[h]
    \centering
    \includegraphics[width=0.8\linewidth]{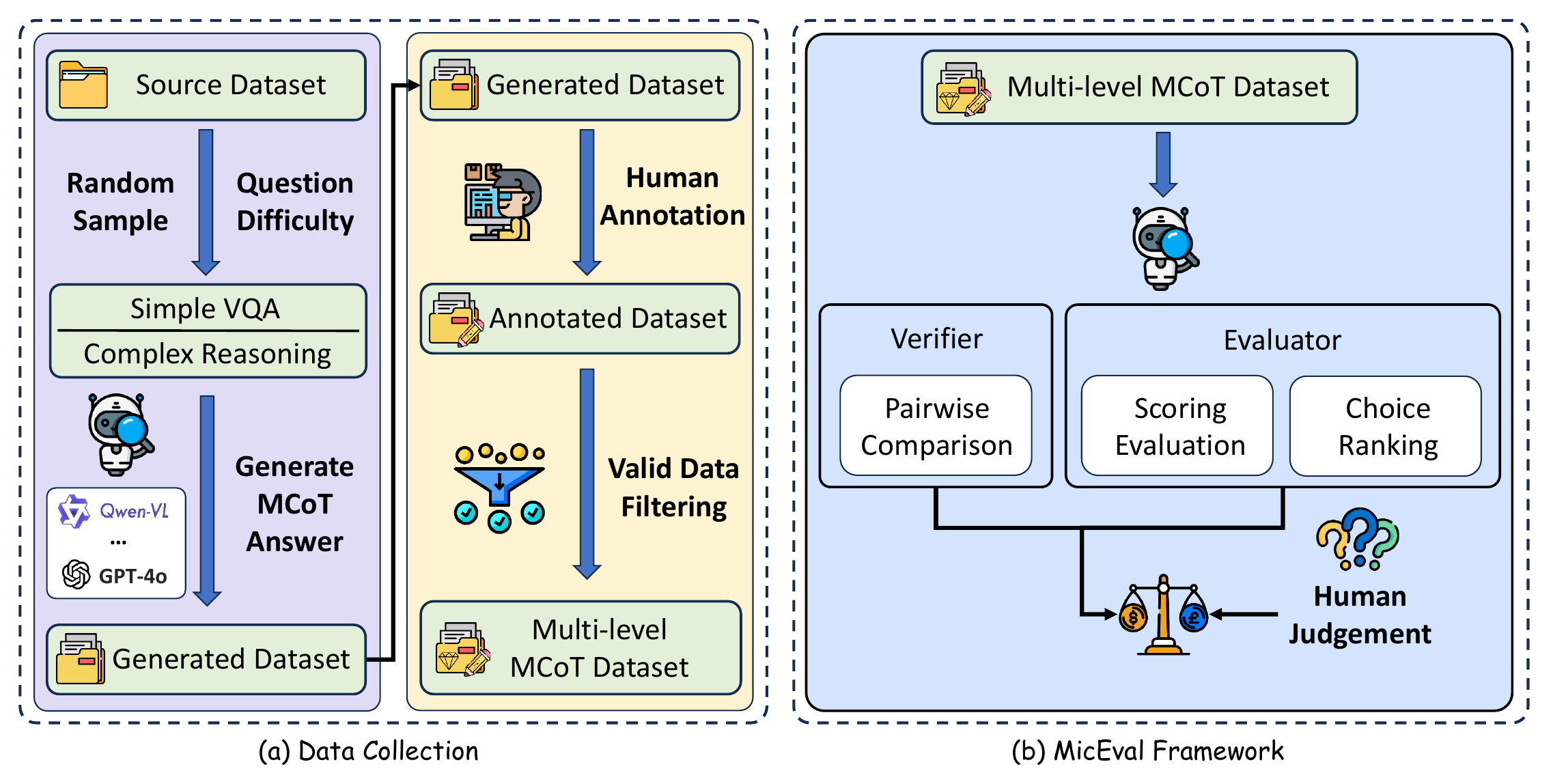}
    \caption{Our work consists of two main parts: (a) sampling questions from the source datasets, generating MCoT answers using four MLLMs, followed by high-quality human annotation and filtering to create the MiCEval dataset; (b) a detailed illustration of our MiCEval framework.}
    \label{fig:framework}
\vspace{-0.4cm}
\end{figure*}

\subsection{Verification and Evaluation Tasks}
In \ours we define two evaluation tasks: Verification \cite{jacovi-etal-2024-chain} and Evaluation \cite{ChenCZWLZZ00024}. Figure~\ref{fig:framework} illustrates the \ours framework.

% \victor{The quality of Figure~\ref{fig:framework} is bad. Raw Dataset is blurry. Dataset below Generated is not readable. Many words have an underlying red line from the editor's grammar corrector. It has spelling mistakes, it should be correlation. Also it would seem that different types of fonts have been used in similar cases. }

 \noindent \textbf{Verification.} %In this task, we aim to assess the ability of the current MLLMs to function as verifiers through the verifier task. We input the MCoT, and the MLLM outputs either $0$ (“Incorrect”) or $1$ (“Correct”). 
In this task, we aim to assess the ability of MLLMs as verifiers. Given an input MCoT, the MLLM outputs either $0$ (“Incorrect”) or $1$ (“Correct”).

 \noindent \textbf{Evaluation.} The goal of this task is to evaluate the correlation between \ours correctness metrics and human performance, using MLLMs as evaluators. Additionally, we aim to determine whether \ours enables MLLMs to achieve closer alignment with human judgments. For a given input MCoT, the MLLM generates a correctness score within the range of $[0,1]$.
 % \hanjie{What's the difference? Verification provides discrete scores while Evaluators gives continuous scores?}
 % \hanjie{In the intro, you only mentioned "verifiers", but here it seems you have a "verifier" and an "evaluator". Please double check the consistency.}

%\textbf{Evaluator.} Our goal in this task is to evaluate the correlation of the \textsc{RGV} correctness metrics with human performance using MLLM as a evaluator. Additionally, we aim to verify that the \textsc{RGV} helps the MLLM demonstrate closer alignment with human judgment. We input the MCoT and the MLLM outputs a correctness score between $[0,1]$.

\subsection{MCoT-level and Step-level Evaluation}
%There are two distinct evaluation settings in RGV framework: MCoT-level evaluation and step-level evaluation.
There are two distinct evaluation settings in \ours: MCoT-level and step-level evaluation.

\noindent \textbf{MCoT-level.} Given the input $\mathcal{I,Q,RC}$, where the $\mathcal{RC}$ comprising \{$r_1$,$r_2$,..., $r_n$\} is evaluated as a whole. MLLM generates a score between $[0,1]$ that reflects the correctness of the entire sequence $\mathcal{RC}$.

% \noindent \textbf{MCoT-level.} Given the input $\mathcal{I,Q,RC}$, the MLLM generates a  score between $[0,1]$ that reflects the  correctness of the entire sequence $\mathcal{RC}$.

%When we input the $I$, $Q$, and $RC$, the model generates a correctness score between [0,1] that reflects the overall correctness of the entire $RC$. 

\noindent \textbf{Step-level.} For the given input $\mathcal I$, $\mathcal Q$, the current step $r_i$, and the previous steps $\mathcal{RC}_{i-1}$ = \{$r_1$,$r_2$,..., $r_{i-1}$\}, only the current step $r_i$ is evaluated, the MLLM outputs a score between $[0,1]$ that reflects the correctness of  $r_i$.

% \noindent \textbf{Step-level.} For the given  input $\mathcal I$, $\mathcal Q$,  the current step $r_i$, and the previous steps $\mathcal{RC}_{i-1}$ = \{$r_1$,$r_2$,..., $r_{i-1}$\}, the MLLM outputs a score between $[0,1]$ that reflects the correctness of  $r_i$.

% based on $\mathcal{I, Q}$ and $\mathcal{RC}_{i-1}$
% \textbf{Step-level.} In this setting, given the $I$, $Q$, and the current step $r_i$, and the previous steps $RC_{i-1}$ = \{$r_1$,$r_2$,..., $r_{i-1}$\}, the model generates a correctness score between $[0,1]$. This score reflects the correctness of the current step $r_i$ based on $I$, $Q$, and $RC_{i-1}$.

\subsection{{Step Type}}
% \hanjie{Step Type and Property?}}

%\victor{I dont understand what Property refers to}

The definition of correctness varies depending on the type of step. For a \emph{description step}, the focus is on description correctness and relevance. For a \emph{reasoning step}, the emphasis is on logical correctness, relevance, and informativeness.

%Based on the type of step, we define correctness differently. Description Step: We focus on description correctness and description relevance. Reasoning Step: We focus on logic correctness, logic relevance, and informativeness.

In an MCoT sequence, the steps are generated from information extracted from both visual and textual modalities. We define steps derived solely from visual information (i.e., describing visual content) as \emph{description steps}. Steps that involve information inferred beyond the visual content, incorporating the question and previous steps, are referred to as \emph{reasoning steps}.

%(1) Step Type: In MCoT answer, steps are generated based on information extracted from both visual and language modalities. We define steps extracted solely from visual information (i.e., describe the image content) as “Description Step”. Steps contain the information beyond the image content and can be inferred from the question or previous steps are defined as “Reasoning Step”. Common-sense reasoning is also considered a reasoning step. If a step contains both Description and Reasoning, it is classified as “Both Step”.

\subsubsection{Description Step} For description steps, we assess their correctness and relevance.

\noindent \textbf{Description Correctness.} A step is labeled as “Fully Correct” if it contains no incorrect information about the image. If all the information in the step is incorrect, it is marked as “Unsupported.” If a step contains both correct and incorrect information, it is labeled as “Partially Correct.” Additionally, we further categorize error types for “Partially Correct” or “Unsupported” steps based on \citet{NEURIPS2023_f8ad010c}---Entity False: Incorrectly identifying the presence or type of entities; Attribute False: Incorrect description of entity attributes (e.g., color, shape, texture, count, state, or text recognition); Spatial Relationship False: Incorrect description of spatial relationships between entities (e.g., left, right); Non-Spatial Relationship False: Incorrect description of non-spatial relationships (e.g., speaking to).

 \noindent \textbf{Description Relevance.} A step is labeled “Image Relevant” if its description pertains to the content of the image. If the description is relevant to answering the question, it is labeled “Logic Relevant”. If it is relevant to both the image and the question, the step is marked as “Both”. If it is relevant to neither, the step is labeled “None”.

%(2) Correctness of Description Step: For description steps, we assess Description Correctness and Description Relevance. \textbf{Description Correctness}: A step is labeled “Fully Correct” if it contains no incorrect information regarding the image. If all information in the step is incorrect, it is marked as “Unsupported”. A step contains some incorrect information but also include some correct descriptions is labeled as “Partially Correct”. Furthermore, we define error types for “Partially Correct” or “Unsupported” steps based on \citet{NEURIPS2023_f8ad010c}'s work. “Entity False”: Incorrectly describing the presence or type of entities. “Attribute False”: Incorrect description related to the attributes of entities (color, shape, texture, count, state, text recognition). “Spatial Relationship False”: Incorrect description of spatial relationships between entities (e.g., left, right). “Non-spatial Relationship False”: Incorrect description of non-spatial relationships (e.g., speaking to, holding). \textbf{Description Relevance}: A step is labeled “Image Relevant” if the description is relevant to the image content. If the description is relevant to answering the question, it is labeled “Logic Relevant”. If relevant to both the image and answering the question, the step is labeled “Both”. If neither, the step is labeled “None”.

\subsubsection{Reasoning Step}
For reasoning steps, we focus on Logical Correctness, Logical Relevance, and Informativeness. 

\noindent \textbf{Logical Correctness.} A reasoning step is considered “Correct” if it is logically deduced from the facts and previous steps. A step is labeled “Incorrect” if it contains logical errors or contradicts the facts. Additionally, following \citet{prasad-etal-2023-receval}, we further categorize error types for logically incorrect steps---Inter-step Incorrect: When the step cannot be logically inferred or contradicts previous steps; Intra-step Incorrect: When there are internal contradictions within the step; Both: When both inter-step and intra-step errors are present.

 \noindent \textbf{Logical Relevance.} A step is labeled as “Relevant” if it contributes to answering the question; otherwise, it is marked as “Irrelevant”.

 \noindent \textbf{Informativeness.} A step is considered “Informative” if it introduces new information and does not repeat or provide redundant details from previous steps \cite{chen-etal-2023-rev,prasad-etal-2023-receval}. Otherwise, it is labeled as “Uninformative”.

%A step is labeled “Relevant” if it is relevant to answering the question, otherwise, it is “Irrelevant”. \textbf{Informativeness}: A step is “Informative” if it does not repeat or provide redundant information compared to previous steps \cite{chen-etal-2023-rev,prasad-etal-2023-receval}. Otherwise, it is “Uninformative”.

%(3) Correctness of Reasoning Step: For reasoning steps, we focus on Logic Correctness, Logic Relevance, and Informativeness. \textbf{Logic Correctness}: A reasoning step is “Correct” if it is logically deduced based on facts and previous steps, without internal contradictions. A step is “Incorrect” if it contains logical errors or conflicts with the facts. Following \citet{prasad-etal-2023-receval}'s definitions, we further categorize error types of the logic incorrect steps: “Inter-step Incorrect”: If the step can not be logically inferred from or contradicts previous steps. “Intra-step Incorrect”: If there are internal contradictions within the step. “Both”: If both inter-step and intra-step incorrect are present. \textbf{Logic Relevance}: A step is labeled “Relevant” if it is relevant to answering the question, otherwise, it is “Irrelevant”. \textbf{Informativeness}: A step is “Informative” if it does not repeat or provide redundant information compared to previous steps \cite{chen-etal-2023-rev,prasad-etal-2023-receval}. Otherwise, it is “Uninformative”.
\begin{figure}[t]
    \centering
    \includegraphics[width=0.8\columnwidth]{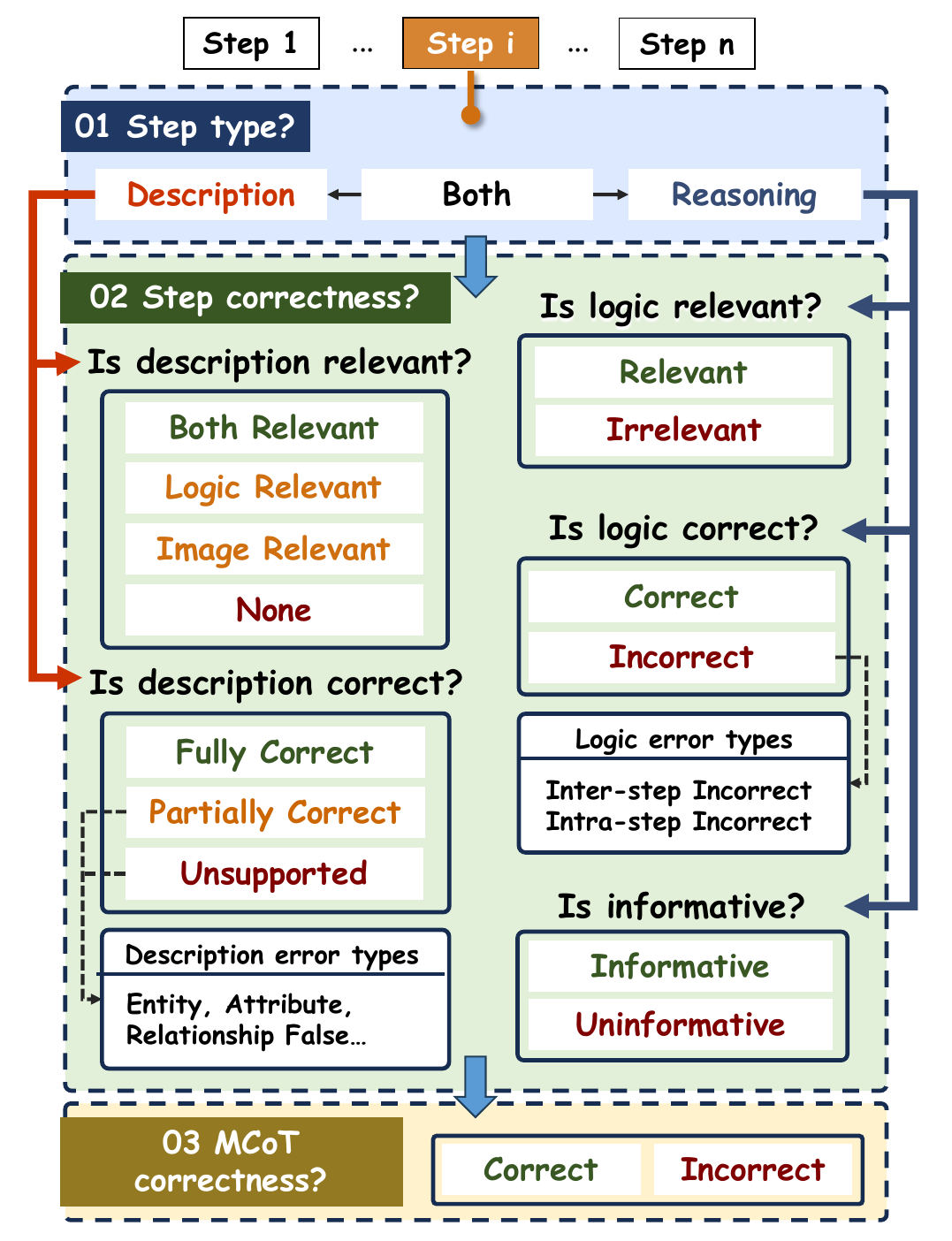}
    \caption{A complete flowchart of the MCoT annotation process. We first determine the type of each step and then annotate its correctness based on the type of step. Once all steps in an MCoT answer are annotated, we evaluate the correctness of the entire MCoT.}
    \label{fig:ann-process}
\vspace{-0.3cm}
\end{figure}

\subsubsection{Full MCoT Sequences}
When evaluating an MCoT, we also assess its overall correctness. An MCoT is considered high-quality (Good MCoT) if all description steps are correct and relevant, and all reasoning steps are correct, relevant, and informative \cite{li-etal-2023-making,jacovi-etal-2024-chain}. If these conditions are not met, it is classified as low-quality (Bad MCoT).

%When evaluating an MCoT, we assess its overall correctness. An MCoT is considered high-quality (Good MCoT) if all Description Steps are correct and relevant, and all Reasoning Steps are correct, relevant, and informative \cite{li-etal-2023-making,jacovi-etal-2024-chain}. If not, it is considered low-quality (Bad MCoT).

% (4) Correctness of MCoT: When evaluating an MCoT, we assess its overall Quality and the Correctness of the Final Answer. \textbf{MCoT Quality}: An MCoT is considered high-quality (Good MCoT) if all Description Steps are correct and relevant, and all Reasoning Steps are correct, relevant, and informative. If not, it is considered low-quality (Bad MCoT). \textbf{Final Answer Correctness}: The final answer is labeled "True" if it correctly answers the question, and "False" otherwise.

\section{Dataset Annotation Process}
\label{sec3}
In this section, we introduce the annotation process for the \ours dataset, which encompasses: \textit{type of step}, \textit{correctness of step}, and \textit{correctness of MCoT}. The complete data annotation process is illustrated in Figure~\ref{fig:ann-process}. Three annotation tasks are performed sequentially, with each task building on the results of the preceding one.

%In this section, we introduce the detailed annotation tasks of the \textsc{RGV} dataset, which includes: \textit{Type of Step} , \textit{Correctnesss of Step}, and \textit{Correctness of MCoT}. The complete data annotation process is illustrated in Figure~\ref{fig:ann-process}. Three annotation tasks are performed sequentially, with each step based on the annotation results from the preceding task.

%\subsection{Stage 1: Type of Step}

 \noindent \textbf{Stage 1: Type of Step.} In this stage, annotators will label the MCoT answers step-by-step according to the definitions of step types provided in Sec.~\ref{sec2}. Following this, the annotation proceeds to the corresponding labeling tasks based on the type of each step.

%In this stage, annotators will label the MCoT answer step-by-step based on the definitions of step types provided in Section 2. Following this, the annotation proceeds to the corresponding labeling task based on the type of each step. 

 \noindent \textbf{Stage 2: Correctness of Step.}
We annotate each step from different perspectives based on the results of Stage 1. For description steps, we label correctness and relevance. For reasoning steps, we label logical correctness, relevance, and informativeness. For steps classified as both description and reasoning, we annotate all aspects of each.

%\subsection{Stage 2: Correctness of Step}
%We annotate each step from different perspectives based on the results of the Stage 1. For Description steps, we label Description Correctness and Description Relevance. For Reasoning steps, we label Logic Correctness, Logic Relevance, and Informativeness. For Both steps, we annotate all the aspects of both Description and Reasoning steps.

When the previous steps contain incorrect information, the annotators need to make additional judgments about the current step. If the current step logically follows or references the incorrect information from the previous steps, it should also be labeled as “Incorrect”. For more details, refer to the annotation illustrated in Figure~\ref{fig:ann-eg}.

% Specifically, if the current step is independent of the previous incorrect information, it can still be marked as correct, even if the previous step was incorrect.
%\subsection{Stage 3: Correctness of MCoT}

 \noindent \textbf{Stage 3: Correctness of MCoT.} 
For MCoT, we evaluate the correctness of the MCoT answer. An MCoT answer is considered high-quality (Correct MCoT answer) when every step in the reasoning chain is a Correct Step. This means that each step, according to its respective type, is annotated as Correct, Relevant, or Informative, with no attribute marked as erroneous based on the criteria defined earlier. If any step contains an attribute that is annotated with an error label, the MCoT answer is deemed low-quality (Incorrect MCoT answer).

% For MCoT, we evaluate both the quality of the MCoT and the correctness of its final answer. An MCoT is considered high-quality (Good MCoT) when every step in the reasoning chain is a Correct Step. This means that each step, according to its respective type, is annotated as Correct, Relevant, or Informative, with no attribute marked as erroneous based on the criteria defined earlier. If any step contains an attribute that is annotated with an error label, the MCoT is deemed low-quality (Bad MCoT).

% As for the final answer correctness, it is determined by whether the MCoT correctly answers the question. The assessment is based on two criteria: (1) whether the answer is relevant to the question and (2) whether it contains no factual or logical errors. This ensures that the correctness is aligned with the task requirements and question context.

\section{Data Collection}
In this section, we present a comprehensive explanation of the construction process for the \ours dataset. The full data collection process is presented in Figure~\ref{fig:framework}, which consists of three stages: \emph{question collection for the dataset, MCoT answer collection}, and \emph{human annotation}.

%In this section, we offer a comprehensive explanation of the construction process for the \textsc{RGV} dataset. The full data collection process is presented in Figure~\ref{fig:framework}, which consists of three stages: Question Collection for the Dataset, MCoT Answer Collection, and Human Annotation.
\subsection{Sampling and Generation}
{We randomly sampled 700 questions from 8 multimodal datasets. Two authors manually conducted coarse-grained annotations based on the difficulty of the questions and source datasets to get \textsc{Hard} and \textsc{Normal} splits. We used four representative MLLMs to generate MCoT answers. Ultimately, we obtained 1,000 MCoT answers generated by the MLLMs. Detailed descriptions are provided in the Appendix~\ref{app1.2}.}

\subsection{High-Quality Annotation Protocol}
It is important to note that the requirement for fine-grained and multi-task annotations necessitates that annotators have relevant expertise. Therefore, all annotators possess the necessary background knowledge related to the annotation tasks.

% It is important to note that the requirement for fine-grained and multi-task annotations necessitates that annotators have relevant expertise. Therefore, all annotators possess the necessary background knowledge related to the annotation tasks. {For details on the preliminary training of annotators, please refer to the App.~\ref{app2}.}

\noindent \textbf{Preliminary Training.} To ensure high-quality data, four of the authors annotated the entire MCoTs (each CoT has three annotations). Previous to the generation of the official annotation of the MCoT answers in the dataset, all annotators underwent preliminary training. We randomly selected 30 questions from the remaining 800 questions not included in the dataset in Sec.~\ref{app1.2} and used GPT-4o and InstructBLIP to generate a total of 60 MCoT answers. The annotators then completed the annotation tasks on these 60 MCoT answers, and based on the annotation results, the annotators engaged in discussions with two additional authors, to reach an internal consensus on some of the more challenging cases.

%It is important to note that the fine-grained and multi-task annotations requirement annotators to have relevant expertise. Therefore, all annotators possess relevant background knowledge related to the annotation task.

\noindent \textbf{Human Annotation.} Each MCoT answer underwent annotation based on the tasks outlined in Sec.~\ref{sec3}. We established a rigorous process to filter for valid data. If any annotation for a task on a given step could not be determined through majority voting (i.e., a 1:1:1 or 1:1 tie), the CoT was considered invalid. In the end, we obtained 903 valid MCoTs. More details can be found in the App.~\ref{app2.2}. We computed Bennett, Alpert, and Goldstein's S inter-rater agreement for the MiCEval-\textsc{Hard} and MiCEval-\textsc{Normal}, with scores of 0.888 and 0.795, respectively. The S agreement for the description step and reasoning step annotations was 0.877 and 0.859, respectively.

\subsection{Data Statistics}
We provide the statistics of MiCEval-\textsc{Normal} and MiCEval-\textsc{Hard} in Table~\ref{tab:statistic}. MiCEval-\textsc{Normal} contains 1,144 MCoT steps and MiCEval-\textsc{Hard} contains 1,745 MCoT steps for evaluation. We observe significant differences between the two splits in the total number of reasoning steps, the average of MCoT steps per MCoT answer, and the number of fully correct MCoT answer. These differences reflect that our  splits based on question difficulty are meaningful. Further analysis of the dataset can be found in the Appendix \ref{app1.3}.

\begin{table}[t]
\small
\centering
\begin{tabular}{l|c|c}
  \toprule
    \textbf{Dataset(Eval)} & \textbf{Hard} & \textbf{Normal} \\
    \midrule
    {Question} & 323 & 320 \\
    {MCoT Answer} & 457 & 446 \\
    {MCoT Step} & 1,745 & 1,144 \\
    {Avg. step per MCoT} & 3.8 & 2.6 \\
    {Description Step} & 899 & 852 \\
    {Reasoning Step} & 769 & 288 \\
    {Description Fully Correct Step} & 679 & 745 \\
    {Logic Fully Correct Step} & 515 & 201 \\
    {Fully Correct MCoT} & 185 & 307 \\
  \bottomrule
\end{tabular}
\caption{Statistics on MiCEval-\textsc{Normal} and \textsc{Hard}.}
\label{tab:statistic}
\vspace{-0.3cm}
\end{table}

\section{\textsc{The \ours Framework }}
The \ours framework also introduces two evaluation metrics: one at the step-level and the other at the MCoT-level. The dataset and metrics are designed based on the correctness of the various MCoT step types.

%We now introduce the \textsc{RGV} framework, which includes a fine-grained step-level dataset and two metrics at step-level and MCoT-level. Both the dataset and metrics are established based on the correctness of MCoT step types.

\subsection{Evaluation of Step Correctness}
We propose two methods for measuring step correctness in MCoT answers.

%We propose two methods for measuring step correctness in MCoT answers, based on different types of MCoT answer step. 
% \hanjie{Can we use another word like "Step Quality"? It's a bit confusing as "correctness" has been overused for a specific property and the overall quality.}
\noindent \textbf{Description Step Correctness.}
Our goal is to instruct the MLLM to generate a correctness score for a description step $r_i$ using a prompt-based input $(\mathcal{I}, \mathcal{Q}, \mathcal{RC}_{i})$, where $\mathcal{RC}_{i}$ = \{$r_1$,$r_2$,..., $r_{i}$\}. Based on the definitions provided in Sec.~\ref{sec3}, the following metric is derived:
%We aim to instruct the MLLM in generating the correctness score for a description step $r_i$ through a prompt-based input $(\mathcal{I}, \mathcal{Q}, \mathcal{RC}_{(i)})$. According to the definitions outlined in section 3, the following metric is derived:
\[\textit{S}_{d\_\textit{correct}}, \textit{S}_{d\_\textit{relevant}} = \textit{M}_{\textit{prompt}}(\mathcal{I}, \mathcal{Q}, \mathcal{RC}_{i})\]
\[\textit{Correctness}^{(i)}_{D} = \textit{S}_{d\_\textit{correct}} \odot \textit{S}_{d\_\textit{relevant}}\]
where $M_{\textit{prompt}}$ represents the prompt-based MLLM, $\textit{S}_{d\_\textit{correct}}$ refers to the  description correctness score, and $\textit{S}_{d\_\textit{relevant}}$ represents the description relevance score. $\odot$ indicates the geometric mean operation.

% \victor{Notation in blue above is undefined.}

\noindent \textbf{Reasoning Step Correctness.} Similarly, using a prompt-based input $(\mathcal{I}, \mathcal{Q}, \mathcal{RC}_{i})$, we guide the MLLM for generating the correctness score for the reasoning step $r_i$. The corresponding metric is defined as:
%Similarly, using a prompt-based input $(\textit{I}, \textit{Q}, \textit{RC}_{(i)})$, we guide the MLLM in generating the correctness score for the reasoning step $r_i$. The corresponding metric is defined as:
\[\textit{S}_{l\_\textit{correct}}, \textit{S}_{l\_\textit{relevant}}, \textit{S}_{\textit{info}} = \textit{M}_{\textit{prompt}}(\mathcal{I}, \mathcal{Q}, \mathcal{RC}_{i})\]
\[\textit{Correctness}^{(i)}_{R} = \textit{S}_{l\_\textit{correct}} \odot \textit{S}_{l\_\textit{relevant}} \odot \textit{S}_{\textit{info}}\]
where $M_{\textit{prompt}}$ represents the prompt-based MLLM, $\textit{S}_{l\_\textit{correct}}$ denotes the logical correctness score, $\textit{S}_{l\_\textit{relevant}}$ refers to the logical relevance score, and $\textit{S}_{\textit{info}}$ captures the informativeness score. 

\subsection{Evaluation of MCoT Correctness}
\textbf{MCoT Correctness.} We compute the overall score for an MCoT answer by generating a score for each step using the prompt-based MLLM. {First, we need to obtain the type of each step based on MLLMs:} 
{\[\textit{Type}^{(i)} = \textit{M}_{\textit{prompt}}(\mathcal{I}, \mathcal{Q}, \mathcal{RC}_{i})\]}

Then, based on the type of each step, we calculate its correctness score and  the  score of the entire MCoT  as follows:
\[\resizebox{1\columnwidth}{!}{
$\textit{Correctness}^{(i)} = 
\begin{cases}
\textit{Correctness}^{(i)}_{D}, & \text{description step} \\
\textit{Correctness}^{(i)}_{R}, & \text{reasoning step}
\end{cases}$}\]
{\small \[\textit{Correctness}_{\textit{type}} = \sqrt[n]{ \prod_{i=1}^{n} \textit{Correctness}^{(i)} }\]}
% \[\textit{S}_{i} = \textit{M}_{prompt}(\textit{I}, \textit{Q}, \textit{RC}_{i})\]
% \[\textit{Correctness}_{MCoT} = \textit{S}_{d} \odot \textit{S}_{l}\]

\noindent{}where $\textit{Correctness}^{(i)}_{D}$ represents the correctness score for the description step $r_i$ and $\textit{Correctness}^{(i)}_{R}$ represents the correctness score for the reasoning step $r_i$. The overall MCoT score is computed as the geometric mean of all step correctness scores.

{Given the limitations of current research on MCoT step type classification, we also propose another method for calculating correctness that is not based on step type. This method involves calculating the description step correctness and reasoning step correctness for each step, and then deriving the overall score for the entire MCoT:}
\vspace{-0.1cm}
{\[\resizebox{1\columnwidth}{!}{ 
$\textit{Correctness}^{(i)} = \textit{Correctness}^{(i)}_{D} \odot \textit{Correctness}^{(i)}_{R}$}\]}
\vspace{-0.2cm}
{\small \[\textit{Correctness}_{\textit{all}} = \sqrt[n]{ \prod_{i=1}^{n} \textit{Correctness}^{(i)} }\]}
\vspace{-0.1cm}

{\noindent{}where $\textit{Correctness}^{(i)}_{D}$ represents the correctness score for the step $r_i$ and $\textit{Correctness}^{(i)}_{R}$ represents the correctness score for the step $r_i$, $\odot$ denotes the geometric mean operation. The overall MCoT score is computed as the geometric mean of all step correctness scores.}

\section{Experiments}
In the proposed \ours framework, we conducted MCoT verification and MCoT evaluation experiments on existing MLLMs at both the step-level and MCoT-level. 
%We conducted MCoT verification and MCoT evaluation experiments on existing MLLMs at both the step-level and MCoT-level based on the \textsc{RGV} framework.

\subsection{MLLM-as-a-Verifier}
\textbf{Pairwise comparison.} In this experiment, we integrate MLLM-as-a-verifier into the \ours framework to validate fine-grained tasks. The labels generated by the MLLMs serve as predictions, which are subsequently compared to human annotations to assess accuracy. Our evaluation focuses on two key aspects: the correlation between MLLM outputs and human annotations, and the effectiveness of MLLMs as MCoT verifiers.

%In this experiment, we utilize MLLM-as-a-verifier within the \textsc{RGV} framework to validate the fine-grained tasks. The labels generated by the MLLMs are used as predictions, which are then compared with human annotations to calculate accuracy. This experiment, conducted within the RGV framework, evaluates two aspects: the correlation between MLLMs and human annotations, and the ability of MLLMs as MCoT verifiers.

\subsubsection{Experimental Setting} 
{We assess the performance of seven mainstream MLLMs as MCoT verifiers in both zero-shot and few-shot settings. Details of the experimental setting and metrics can be found in Appendix~\ref{app3.1}.}

\subsubsection{Results}
\label{sec6.1.2}
\begin{table*}[h]
    \centering
    \resizebox{\textwidth}{!}{
    \begin{tabular}{c|l|c|ccc|cccc|c|c}
    \toprule
      \multirow{2}{*}{\textbf{Evaluation Setting}} &\multicolumn{1}{c|}{\multirow{2}{*}{\textbf{Model}}} & \multirow{2}{*}{\textbf{Step Type}} &\multicolumn{3}{c|}{\textbf{Description}} & \multicolumn{4}{c|}{\textbf{Logic}} & \multirow{2}{*}{\textbf{MCoT}} & \multirow{2}{*}{\textbf{Avg.}} \\ 
      
       ~ & ~ & ~ & {\textbf{Relevance}} & {\textbf{Correctness}} & {\textbf{Error Types}} & {\textbf{Relevance}} & {\textbf{Correctness}} & {\textbf{Informativeness}} & {\textbf{Error Types}} ~ &\\
        
        \midrule
\multirow{7}{*}{\centering{Zero-Shot}}& {LLaVA-1.6-Mistral-7B}    & 0.653 & 0.456 & 0.203 & 0.175 & 0.901 & 0.487 & \textbf{0.922} & 0.354 & 0.379 & 0.503 \\
~&{MiniCPM-V-2.6}  & 0.121 & 0.693 & 0.702 & 0.248 & 0.918 & 0.664 & 0.837 & \textbf{0.836} & 0.564 & 0.620 \\
~&{Llama-3.2-11B-Vision-Instruct}   & 0.137 & 0.806 & 0.028 & 0.309 & 0.851 & 0.541 & 0.375 & 0.534 & 0.703   &  0.493  \\
~&{LLaVA-1.6-Yi-34B}  & 0.348 & 0.575 & \textbf{0.704} & 0.443 & \textbf{0.949} & \textbf{0.693} & 0.917 & 0.690 & 0.508 & 0.647 \\
~&{GPT-4o}  & 0.672 & 0.768 & 0.631 & \textbf{0.545} & 0.903 & 0.649 & 0.732 & 0.765 & \textbf{0.726} & \textbf{0.710} \\
~&{Qwen-VL-Max}  & 0.433 & \textbf{0.853} & 0.630 & 0.455 & 0.931 & 0.691 & 0.910 & 0.227 & 0.580 & 0.634 \\
~&{Gemini-1.5-Pro}& \textbf{0.868} & 0.838 & 0.674 & 0.398 & 0.926 & 0.584 & 0.330 & 0.827 & 0.670 & 0.679 \\
        \midrule
\multirow{7}{*}{\centering{Few-Shot}}  & {LLaVA-1.6-Mistral-7B} & 0.452 & 0.310 & 0.255 & 0.371 & 0.934 & 0.688 & 0.812 & \textbf{0.863} & 0.584 & 0.585 \\
~& {MiniCPM-V-2.6}  & 0.104 & 0.413 & \textbf{0.719} & 0.401 & 0.873 & 0.679 & 0.764 & 0.811 & 0.547 & 0.590 \\
~& {Llama-3.2-11B-Vision-Instruct}& 0.044 & 0.257 & 0.601 & 0.004 & 0.638 & 0.670 & 0.069 & 0.220 & 0.376 & 0.320 \\
~& {LLaVA-1.6-Yi-34B} & 0.703 & 0.206 & 0.638 & 0.488 & \textbf{0.939} & 0.670 & \textbf{0.927} & 0.751 & 0.716 & 0.671 \\
~& {GPT-4o} & \textbf{0.871} & \textbf{0.455} & 0.642 & \textbf{0.687} & 0.891 & 0.671 & 0.784 & 0.773 & \textbf{0.742} & \textbf{0.724} \\
~& {Qwen-VL-Max} & 0.359 & 0.262 & 0.603 & 0.370 & 0.924 & \textbf{0.776} & 0.749 & 0.650 & 0.689 & 0.598 \\
~& {Gemini-1.5-Pro}& - & - & - & - & - & - & - & - & - & - \\

        \bottomrule
    \end{tabular}
    }
        \setlength{\belowcaptionskip}{0.2cm}
    \caption{The overall performance on Pairwise Comparison of different MLLMs in MiCEval-\textsc{Hard}. Due to limited funding, Gemini-1.5-Pro was not evaluated in the few-shot setting.}
    \label{tab:pair-hard}
    \vspace{-0.6cm}
\end{table*}

\begin{figure}[t]
    \centering
    \includegraphics[width=0.8\linewidth]{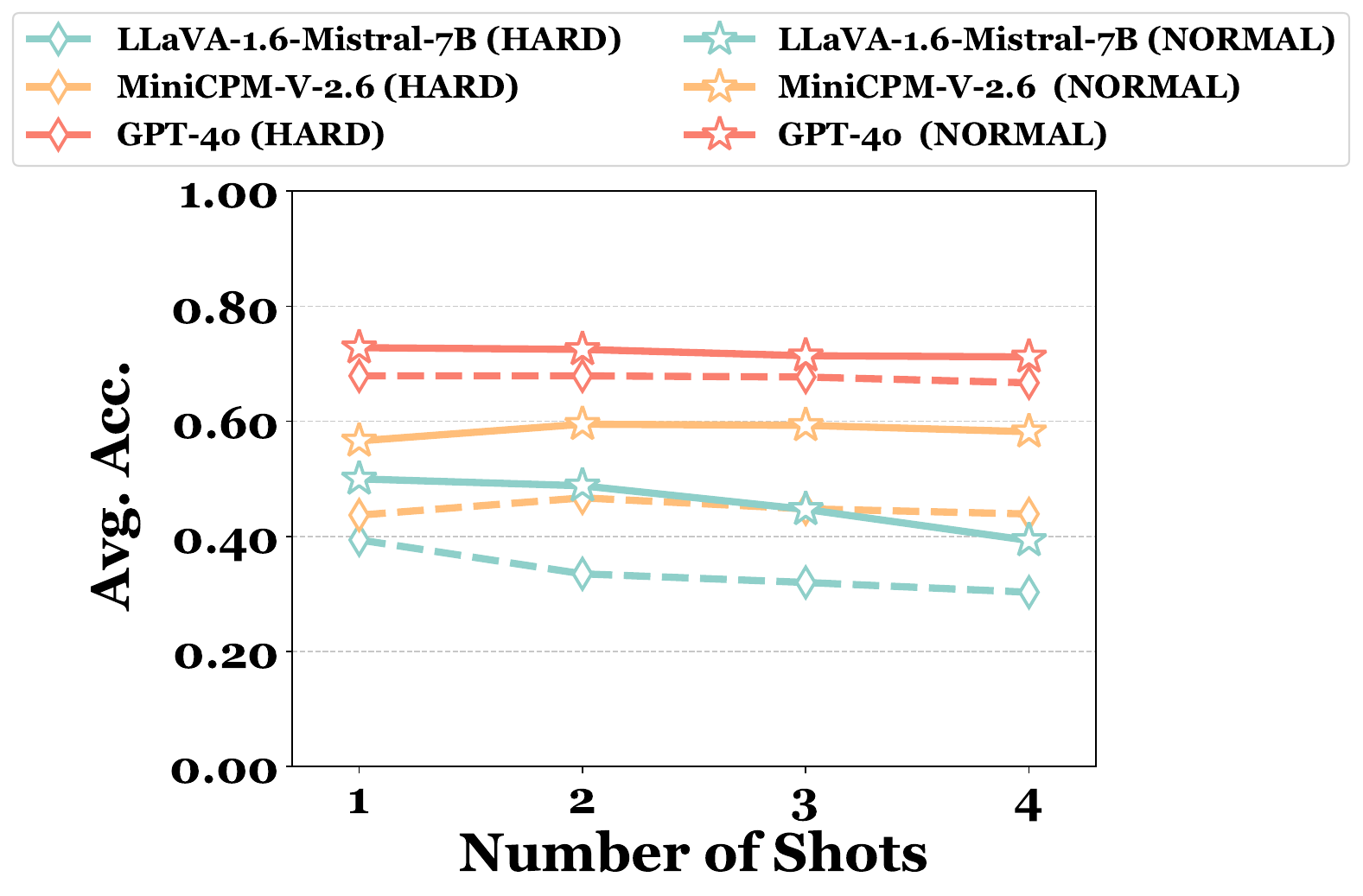}
    \caption{The relationship between the average accuracy of three MLLMs across all Pairwise Comparison tasks and the number of shots on two splits. }
    % \textbf{HARD}: The performance on MiCEval-\textsc{Hard}; \textbf{NORMAL}: The performance on MiCEval-\textsc{Normal}.
    \label{fig:shot-number}
    \vspace{-0.3cm}
\end{figure}

{The results are presented in Tables~\ref{tab:pair-hard} and~\ref{tab:pair-normal}. The results of the few-shot preliminary experiments are shown in Figure~\ref{fig:shot-number}, while the results of the exploration experiments for few-shot types and more results can be found in Appendix \ref{app3.3}}.

%MCoT answers in the \textsc{RGV-Hard} include more reasoning steps and are more complex compared to \textsc{RGV-Normal}. MLLM's performance on \textsc{RGV-Normal} consistently surpasses that on \textsc{RGV-Hard}, indicating that the correlation between current MLLMs and human judgments is more closely on simple reasoning tasks, and the MLLMs are not good at validating MCoTs that involve complex reasoning. 

\noindent \textit{ $\triangleright$ Few-shot does not always outperform zero-shot.} On MiCEval-\textsc{Normal}, only LLaVA-1.6-Mistral-7B demonstrated improvement, with a few-shot average accuracy of 0.657 compared to 0.575 in the zero-shot setting. The other MLLMs, however, showed performance declines, with Llama-3.2-11B-Vision-Instruct experiencing the most significant drop. A detailed analysis of Llama-3.2's performance is provided in  Appendix~\ref{app3.4}. On MiCEval-\textsc{Hard}, few-shot performance improved for LLaVA-1.6-7B, LLaVA-1.6-Yi-34B, and GPT-4o, while MiniCPM-V-2.6 and Gemini-1.5-Pro experienced declines.

%On \textsc{RGV-Normal}, only LLaVA-1.6-Mistral-7B shows an improvement, with a few-shot average accuracy 0.657 compared to 0.575 in zero-shot. Other MLLMs, however, saw performance declines, with Llama-3.2-11B-Vision-Instruct experiencing the most significant drop. A detailed analysis of Llama-3.2-11B-Vision-Instruct's performance can be found in the Appendix. On \textsc{RGV-Hard}, few-shot performance improved for LLaVA-1.6-Mistral-7B, LLaVA-1.6-Yi-34B, and GPT-4o, while MiniCPM-V-2.6 and Gemini-1.5-Pro showed a decline.

\noindent  \textit{$\triangleright$ MLLMs still exhibit shortcomings in handling complex reasoning tasks.} In comparison to the logical relevance and informativeness tasks on MiCEval-\textsc{Hard}, the average accuracy of most MLLMs across other tasks, in both zero-shot and few-shot settings, remains below 0.7, with a few notable exceptions: MiniCPM achieved 0.702 (zero-shot) and 0.719 (few-shot) in description correctness, while GPT-4o scored 0.726 (zero-shot) and 0.742 (few-shot) on the MCoT task. Most MLLMs demonstrated a low correlation with human judgments in description correctness and error type, indicating substantial room for improvement in the visual modality of current MLLMs.

%Compared to the Logic Relevance and Informativeness tasks on \textsc{RGV-Hard}, the average accuracy of most MLLMs in other tasks under both zero-shot and few-shot settings is generally below 0.7, with a few exceptions such as MiniCPM in Description Correctness is 0.702 (zero-shot) and 0.719 (few-shot), GPT-4o in MCoT task is 0.726 (zero-shot) and 0.742 (few-shot). Most MLLMs exhibited low correlation with human judgments in Description Correctness and Description error type further highlights the significant room for improvement in the visual modality of current MLLMs.

\noindent \textit{$\triangleright$ The number of shots in  few-shot settings does not always correlate with better model performance.}
As the number of shots increased, the performance of LLaVA-1.6-7B declined from 0.500 to 0.393 on MiCEval-\textsc{Hard} and from 0.394 to 0.303 on MiCEval-\textsc{Normal}. The other two MLLMs exhibited minimal changes in performance. Consequently, we selected 1-shot for all subsequent multimodal few-shot experiments.

%As the number of shots increased, the performance of LLaVA-1.6 decreased from 0.500 to 0.393 on the \textsc{RGV-hard} and from 0.394 to 0.303 on the \textsc{RGV-normal}. The other two MLLMs showed minimal change in performance. Therefore, we selected 1-shot for all subsequent Multimodal few-shot experiments.

% \noindent \textit{$\triangleright$ Good performance in textual few-shot evaluation can be achieved.} GPT-4o and MiniCPM showed similar performance when handling multimodal demonstrations and textual demonstrations. In contrast, LLaVA-1.6-Mistral-7B exhibited nearly 20\% improvement of accuracy in the textual few-shot evaluation compared to the multimodal few-shot evaluation.

\subsection{MLLM-as-a-Evaluator}
In this experiment, we have two distinct evaluation tasks: \emph{scoring evaluation} and \emph{choice ranking} \cite{zheng2023judging,ChenCZWLZZ00024}.

\noindent \textbf{Scoring Evaluation.} 
In this experiment, we utilize MLLMs as \ours evaluators to score the steps of MCoT answers on a scale from 0 to 10. Human annotations are mapped to 0 (``Incorrect''), 1 (``Correct''), or 0.5 (only for the ``Partially Correct'' label in the description correctness task). We then compute the correlation between human annotations and the scores generated by the \ours evaluators to assess the effectiveness of \ours metrics in detecting the quality of MCoT responses. The prompt templates used in this experiment can be found in App.~\ref{app4.2}.

%In this experiment, we leverage the MLLMs as \ours evaluators, scoring the steps of MCoT answer on a scale of 0 to 10. Human annotations will be mapped to 0 (“Incorrect”), 1 (“Correct”), or 0.5 (only for the “Partially Correct” label in the Description Correctness task). We then compute the correlation between human annotations and the scores generated by the \textsc{RGV} evaluators to evaluate the ability of \textsc{RGV} metrics in detecting the quality of MCoT. 

\noindent \textbf{Choice Ranking.} Using the common split adopted by four MLLMs in Sec.~\ref{app1.2}, we constructed a high-quality MCoT selection dataset consisting of 70 questions and 257 MCoT answers. Details about Choice Ranking are presented in the Appendix.

%Based on the common split,  used by four MLLMs in Section 4.2, we constructed a high-quality MCoT selection dataset comprising 70 questions and 257 MCoT answers. As Figure~\textcolor{blue}{PUT THE REFERENCE TO THE FIGURE} illustrates, in the dataset, each question contains at least one choice that is a fully correct MCoT and one choice that is not. We calculated the MCoT answer correctness scores for each choice, selecting the highest score one. We then computed the accuracy based on whether the selected choice was a fully correct MCoT. The MCoT evaluation metrics with the highest accuracy can be considered to possess a stronger capability for filtering high-quality MCoTs.

\subsubsection{Experimental Setting}
{We conduct correlation evaluations based on {two splits of  MiCEval: MiCEval-\textsc{Hard}, MiCEval-\textsc{Normal}}. Details of the experimental setting and metrics can be found in  Appendix \ref{app4.1}.}

\subsubsection{Results}

Table \ref{tab:somersd} presents the performance of various methods on the \textit{scoring evaluation}. Results for \textit{choice ranking} across different methods are shown in Table \ref{tab:choice-ranking}. For additional analysis, please refer to  Appendix~\ref{app4.3}.
% \hanjie{why do we use blue color?}

\begin{table}[h]
\scriptsize
\centering
\begin{tabular}{l|ccc}
\toprule
    \textbf{Metric} & \textbf{Normal} 
    & \textbf{Hard} &\textbf{MiCEval} \\
 \midrule
    CLIP & 0.079 & 0.019 & 0.060 \\
    BLIP2-ITM & 0.088 & -0.065 & 0.031 \\
    BLIP2-ITC & 0.072 & -0.006 & 0.075 \\
    ReCEval & -0.006 & 0.015 & 0.040 \\
    LLM-Score & 0.078 & 0.210 & 0.162 \\
 \midrule
    MiniCPM-V-2.6 (1) & 0.154 & 0.123 & 0.130 \\
    LLaVA-1.6-Mistral-7B (1) & 0.109 & -0.024 & 0.080 \\
    Llama-3.2-11B-Vision-Instruct (1) & 0.090 & 0.218 & 0.176 \\
    GPT-4o (1) & 0.154 & 0.256 & 0.208 \\
 \midrule
    MiniCPM-V-2.6 (2) & 0.112 & 0.123 & 0.169 \\
    LLaVA-1.6-Mistral-7B (2) & 0.083 & 0.178 & 0.140 \\
    Llama-3.2-11B-Vision-Instruct (2) & 0.031 & 0.090 & 0.075 \\
    GPT-4o (2) & 0.178 & \textbf{0.282} & 0.257 \\
 \midrule
    MiniCPM-V-2.6 (3) & 0.264 & 0.188 & 0.273 \\
    LLaVA-1.6-Mistral-7B (3) & 0.081 & 0.133 & 0.121 \\
    Llama-3.2-11B-Vision-Instruct (3) & 0.157 & 0.159 & 0.186 \\
    GPT-4o (3) & 0.229 & 0.255 & 0.265 \\
 \midrule
    MiniCPM-V-2.6 (4) & \textbf{0.270} & 0.191 & 0.277 \\
    LLaVA-1.6-Mistral-7B (4) & 0.083 & 0.141 & 0.129 \\
    Llama-3.2-11B-Vision-Instruct (4) & 0.156 & 0.172 & 0.194 \\
    GPT-4o (4) & 0.245 & 0.272 & \textbf{0.284} \\

\bottomrule
\end{tabular}
   \setlength{\belowcaptionskip}{0.2cm}
    \caption{Somer's-D scores of different evaluation metrics on two splits of MiCEval. (1) A holistic evaluation of the entire MCoT without the MiCEval; (2) “Step-by-Step” evaluation without MiCEval; (3) “Step-by-Step” evaluation with $\textit{Correctness}_{type}$-based MiCEval; (4) “Step-by-Step” evaluation with $\textit{Correctness}_{all}$-based MiCEval.}
    \label{tab:somersd}
    \vspace{-0.7cm}
\end{table}

\begin{table}[htb]
\scriptsize
\centering
\begin{tabular}{l|c}

\toprule
\textbf{Model} & \textbf{Acc.} \\
\midrule
CLIP & 0.700 \\
BLIP2-ITM & 0.657 \\
BLIP2-ITC & 0.657 \\
ReCEval & 0.600 \\
LLM-Score & 0.614 \\
\midrule
MiniCPM-V-2.6 (1) & 0.686 \\
LLaVA-1.6-Mistral-7B (1) & 0.686 \\
Llama-3.2-11B-Vision-Instruct (1) & 0.729 \\
GPT-4o (1) & 0.757\\
\midrule
MiniCPM-V-2.6 (2) & 0.700 \\
LLaVA-1.6-Mistral-7B (2) & 0.571 \\
Llama-3.2-11B-Vision-Instruct (2) & 0.757 \\
GPT-4o (2) & 0.771 \\
\midrule
MiniCPM-V-2.6 (3) & 0.800 \\
LLaVA-1.6-Mistral-7B (3) & 0.600  \\
Llama-3.2-11B-Vision-Instruct (3) & 0.743 \\
GPT-4o (3) & 0.786 \\
\midrule
MiniCPM-V-2.6 (4) & \textbf{0.814} \\
LLaVA-1.6-Mistral-7B (4) & 0.600 \\
Llama-3.2-11B-Vision-Instruct (4) & 0.757 \\
GPT-4o (4) & \textbf{0.814} \\

\bottomrule
\end{tabular}
    
\setlength{\belowcaptionskip}{0.2cm}
\caption{Accuracy of different evaluation metrics on Choice Ranking. Settings are provided in Table~\ref{tab:somersd}.}
\label{tab:choice-ranking}
\vspace{-0.3cm}
\end{table}

\noindent\textit{$\triangleright$ Existing metrics are not well-suited for evaluating MCoT answers.} MLLM-based methods show a stronger correlation with human judgment in MCoT answers compared to existing approaches. Among current metrics, LLM-Score achieves the highest performance on the entire \ours dataset, with a Somer’s-D score of 0.162. In general, the MLLM-based methods outperform LLM-Score, with the exception of LLaVA-1.6-Mistral-7B.

%Compared to existing methods, the MLLM-based methods demonstrate better correlation with human judgment in MCoT answers. Among the current metrics, LLM-Score shows the best performance on the whole \textsc{RGV}, with a Somer’s-D score of 0.162. The performance of the \textsc{RGV} MLLM-based methods exceeds that of LLM-Score, except for LLaVA. 

\noindent \textit{$\triangleright$ \ours brings the MLLM evaluator closer to human preferences.}
The evaluation metric based on the description and reasoning correctness of \ours shows better performance and more closely aligns with human judgments compared to evaluations without \ours. We observe that MiCEval-based methods, specifically $\textit{Correctness}_{type}$ and $\textit{Correctness}^{(i)}_{all}$ using MiniCPM-V-2.6, Llama-3.2-11B-Vision-Instruct, and GPT-4o, exhibit a stronger correlation in MCoT answer evaluation than the other two MLLM-based approaches. Notably, GPT-4o with $\textit{Correctness}^{(i)}_{all}$ achieved the highest Somer’s-D score of 0.284 across the entire \ours dataset.

\noindent\textit{$\triangleright$ MiCEval  helps  filtering out high-quality MCoTs.} 
In the choice ranking task, CLIP achieved the best result with an accuracy of 0.700. The performance of the three MLLM-based methods varied: MiniCPM (CoT, w/o MiCEval) reached an accuracy of only 0.686 and all LLaVA-based evaluation metrics were below 0.700. However, the remaining MLLM-based methods outperformed CLIP. Compared to the other two approaches (without MiCEval), MiniCPM, Llama-3.2, and GPT-4o performed better using the MiCEval metrics evaluation, with GPT-4o and MiniCPM achieving the highest result.

%In the Choice Ranking task, CLIP achieved the best result with a accuracy of 0.700. The performance of the three MLLM-based methods varied: MiniCPM-V-2.6 (CoT, w/o MiCEval) reached an accuracy of only 0.686 and all LLaVA-based evaluation metrics were below 0.700. However, the remaining MLLM-based methods outperformed CLIP. Compared to the other two approaches (without MiCEval), MiniCPM, Llama, and GPT-4o performed better using the MiCEval metrics evaluation, with GPT-4o and MiniCPM-V-2.6 achieving the highest result.

\subsection{Analysis}

%We observed that, whether on \textsc{RGV-Hard} or \textsc{RGV-Normal}, all MLLMs performed worse on Description Relevance in the few-shot setting compared to zero-shot. We analyzed the accuracy of each model across the labels in the Description Relevance task, and the results are shown in Table~\ref{tab:f1-hard} and Table~\ref{tab:f1-normal}. Our analysis revealed that the prediction distribution of each model shifted in different ways during few-shot evaluation. We believe this performance drop may be due to the larger number of label classes in Description Relevance compared to other tasks, as well as the generally weaker In-context learning abilities of MLLMs in this task. For the performance of each MLLM on various labels for each task, please refer to the Appendix.

\noindent \textbf{Instruction-following ability of MLLMs.}
Despite our efforts to guide the model's predictions within the defined label categories by experimenting with various instructions, prompt templates, and multiple iterations, a small percentage of invalid outputs persisted due to the current limitations of MLLMs in following instructions. We conducted a statistical analysis of the invalid predictions across all models in the \textbf{Pairwise Comparison} experiments. The overall proportion of invalid outputs was 2.93\%, with the zero-shot evaluation yielding 3.48\% invalid outputs, compared to 2.34\% in the few-shot evaluation. This suggests that few-shot evaluation improves the model's ability to follow instructions. The distribution of invalid output proportions for each model across different settings and datasets is shown in Table~\ref{tab:invalid-output}. For additional analysis, please refer to Appendix~\ref{app3.4} and~\ref{app4.4}.

\begin{table}
\tiny
\centering
\begin{tabular}{l|cc|cc}
\toprule
\multicolumn{1}{c|}{\multirow{2}{*}{\textbf{Model}}} & \multicolumn{2}{c|}{\textbf{Zero Shot}} & \multicolumn{2}{c}{\textbf{Few Shot}} \\
                       & \textbf{NORMAL} & \textbf{HARD}  & \textbf{NORMAL} & \textbf{HARD}  \\
\midrule
LLaVA-1.6-Mistral-7B         & 1.29\%  & 3.94\% & 2.46\% & \textbf{8.59\%} \\
MiniCPM-V-2.6                & 0.00\%  & 1.08\%  & 0.53\%  & 3.49\% \\
LLaVA-1.6-Yi-34B             & 0.18\%  & 0.32\%  & 0.00\%  & 0.87\% \\
GPT-4o                       & 0.74\%  & 1.40\%  & 0.64\%  & 0.91\% \\
Qwen-VL-Max                  & \textbf{14.27\%} & \textbf{18.61\%} & 0.96\% & 5.40\% \\
Llama-3.2-11B-Vision-Instruct & 7.48\%  & 7.12\%  & \textbf{3.08\%} & 2.87\% \\
Gemini-1.5-Pro               &  0.25\% &  0.19\% & - & - \\
\bottomrule
\end{tabular}
\setlength{\belowcaptionskip}{0.2cm}
\caption{Invalid outputs proportions of each MLLMs in zero shot and few shot evaluations under all tasks of \textsc{Normal} and \textsc{Hard} splits.}
\label{tab:invalid-output}
\vspace{-0.6cm}
\end{table}

% \begin{table*} [h]
% \scriptsize
% \centering
% \begin{tabular}{lcccc}
% \toprule
% Model & Zero Shot (NORMAL) & Zero Shot (HARD) & Few Shot (NORMAL) & Few Shot (HARD) \\
% \midrule
% LLaVA-1.6-Mistral-7B         & 5.30\%  & 16.20\% & 30.80\% & 48.60\% \\
% MiniCPM-V-2.6                & 0.00\%  & 2.70\%  & 6.60\%  & 19.70\% \\
% LLaVA-1.6-Yi-34B             & 0.80\%  & 1.30\%  & 0.00\%  & 4.90\% \\
% GPT-4o                       & 3.00\%  & 3.50\%  & 8.10\%  & 5.20\% \\
% Qwen-VL-Max                  & 58.90\% & 46.20\% & 15.90\% & 5.40\% \\
% Llama-3.2-11B-Vision-Instruct & 30.90\%  & 29.30\%  & 38.60\% & 16.20\% \\
% Gemini-1.5-Pro               &  1.00\% &  0.80\% & - & - \\
% \bottomrule
% \end{tabular}
% \setlength{\belowcaptionskip}{0.2cm}
% \caption{Error proportions of models in zero shot and few shot settings with different task difficulties}
% \label{tab:invalid-output}
% \end{table*}

\section{Related Work}
\noindent \textbf{Reasoning Chain Construction and Evaluations.} Chain of Thought (CoT) is an innovative method for solving reasoning tasks using large language models (LLMs) \cite{NEURIPS2022_9d560961}. \citet{NEURIPS2022_8bb0d291} showed  that adding a simple prompt, such as ``Let's think step by step'', significantly improves LLMs performance in zero-shot settings. \citet{jung-etal-2022-maieutic} demonstrated that even erroneous reasoning chains can lead to correct final predictions, which has increasingly shifted the focus towards verifying the correctness of the entire reasoning chain \cite{NEURIPS2022_1fc548a8,chen-etal-2024-reconcile}. To address the  challenge of  automatically generating good quality COTs, one idea is to connect explicit knowledge, such as knowledge graphs (KG)~\cite{PVGW2017,PCEH+2017}, and parametric knowledge from LLMs~\cite{PRKS+2023}.  \citet{WHHH+2024} proposed to generate COTs from relevant  knowledge graphs~\cite{HZWV+2024}. \citet{WCHY+2024} proposed to use knowledge graph patterns, such as thouse widely used for query answering over KGs, to generate effective COT, without having to rely on the content of KGs.
\citet{golovneva2022roscoe} introduced a comprehensive set of CoT evaluation metrics, validating CoT’s logical correctness, grammar, and informativeness. \citet{prasad-etal-2023-receval} evaluates the informativeness and logical correctness of CoTs, while \citet{jacovi-etal-2024-chain} focuses on verifying both attribution and logical correctness of CoT answers.

Broadly speaking, reasoning is also related to text entailments~\cite{SPK2023}, entailment graphs~\cite{journals/tacl/HosseiniCRHCJS18,LLLCP2024,ZLVSP2024}, NL2SQL~\cite{VPZT+2023,SVDV+2024,ZLP2024}, as well as planning~\cite{WWXX+2024,VGYM+2025} and API~\cite{WQLPW2024}.  

% \noindent \textbf{Image-Text Matching.} Multimodal Chain of Thought (MCoT) involves not only logical reasoning but also image description. Several metrics have been developed for evaluating image descriptions, such as CLIP-Score \cite{hessel-etal-2021-clipscore} and BLIP2-Score \cite{li2023blip2}, based on cosine similarity between image and text, \citet{chen-etal-2024-unveiling} combines visual models with LLMs and performs well in image-text matching tasks.

\noindent \textbf{Evaluation based on MLLMs.} Recent works leverage LLM's instruction-following capabilities as an evaluator \cite{fu-etal-2024-gptscore,zheng2023judging,liu-etal-2023-g}. \citet{NEURIPS2023_f8ad010c} used MLLMs as evaluators to assess the degree of alignment between images and text. \citet{ChenCZWLZZ00024} explored the correlation between MLLM judgments as evaluators and human assessments.

\section{Conclusion}
In this paper, we introduce MiCEval, a novel automated evaluation framework designed to assess the correctness of MCoT and evaluate the capabilities of MLLMs in judging the quality of different reasoning steps. We create a multilevel human-annotated MCoT dataset and conduct three distinct experiments to evaluate the alignment between MLLMs' judgments and human agreement: Pairwise Comparison, Scoring Evaluation, and Choice Ranking. The experimental results demonstrate that \ours metrics align more closely with human preferences. Additionally, current MLLMs show notable weaknesses in both visual and language modalities, especially in handling complex reasoning tasks, highlighting an area in need of further improvement.
%In this paper, we propose a new automated evaluation framework called RGV, which assesses the correctness of MCoT and the capabilities of MLLMs as evaluators by evaluating the quality of different steps. We construct a multilevel human-annotated MCoT dataset and establish three distinct experiments based on this dataset to evaluate MLLMs' judgments and human agreement: Pairwise Comparison, Scoring Evaluation, and Choice Ranking. The final experimental results demonstrate that our RGV metrics align more closely with human preferences, helping MLLMs make evaluations that are more in line with human assessments. Furthermore, current MLLMs exhibit shortcomings in both visual and language models, particularly in their performance on complex reasoning tasks, indicating an area that requires improvement.

%\clearpage

\section*{Acknowledgement}
The computations detailed in this research were facilitated by the Edinburgh International Data Facility (EIDF) and the Data-Driven Innovation Programme at the University of Edinburgh.

\section*{Limitation}
\noindent \textbf{Gemini-1.5-Pro under few-shot evaluation.} Compared to zero-shot evaluation, the token consumption in few-shot evaluation is approximately four times higher. In the zero-shot evaluation, the cost of using Gemini-1.5-Pro is twenty times that of GPT-4o. Due to budget constraints, we were unable to conduct a few-shot evaluation for Gemini-1.5-Pro.

\noindent \textbf{Baseline in Pairwise Comparison.} Due to the limitations of the Pairwise Comparison outputs, it is challenging to map the existing method's outputs to 0 and 1 without setting a threshold. However, introducing a threshold could significantly affect the accuracy of the evaluation. Additionally, given the wide range of tasks in Pairwise Comparison, the current method is unable to effectively handle all tasks. Therefore, we did not test other baselines on Pairwise Comparison.

\noindent {\textbf{Step type classification.} In the current zero-shot evaluation, the two best performing  MLLMs in the step type task are GPT-4o and Gemini-1.5-Pro. Therefore, in our experiments, we used Gemini-1.5-Pro as the classifier for step type. From the experimental results, it is evident that the MLLM's classification of step type affects the MiCEval performance based on the $\textit{Correctness}_{type}$. This is an issue we need to address in future work.}

\noindent \textbf{The Instruction-following abilities of current MLLMs.} The instruction-following abilities of current MLLMs still have limitations and are not yet able to handle complex instruction tasks effectively. Despite our best efforts to improve the performance of each model in the evaluation, Qwen-VL-Max and Llama-3.2-11B-Vision-Instruct produced more invalid outputs compared to other models.

\bibliography{acl_latex}

\begin{thebibliography}{63}
\providecommand{\natexlab}[1]{#1}

\bibitem[{Bai et~al.(2023)Bai, Bai, Yang, Wang, Tan, Wang, Lin, Zhou, and Zhou}]{Qwen-VL}
Jinze Bai, Shuai Bai, Shusheng Yang, Shijie Wang, Sinan Tan, Peng Wang, Junyang Lin, Chang Zhou, and Jingren Zhou. 2023.
\newblock Qwen-vl: A versatile vision-language model for understanding, localization, text reading, and beyond.
\newblock \emph{arXiv preprint arXiv:2308.12966}.

\bibitem[{Chen et~al.(2024{\natexlab{a}})Chen, Chen, Zhang, Wang, Liu, Zhou, Zhang, Wan, Zhou, and Sun}]{ChenCZWLZZ00024}
Dongping Chen, Ruoxi Chen, Shilin Zhang, Yaochen Wang, Yinuo Liu, Huichi Zhou, Qihui Zhang, Yao Wan, Pan Zhou, and Lichao Sun. 2024{\natexlab{a}}.
\newblock \href {https://openreview.net/forum?id=dbFEFHAD79} {{MLLM}-as-a-judge: Assessing multimodal {LLM}-as-a-judge with vision-language benchmark}.
\newblock In \emph{Forty-first International Conference on Machine Learning}.

\bibitem[{Chen et~al.(2023{\natexlab{a}})Chen, Brahman, Ren, Ji, Choi, and Swayamdipta}]{chen-etal-2023-rev}
Hanjie Chen, Faeze Brahman, Xiang Ren, Yangfeng Ji, Yejin Choi, and Swabha Swayamdipta. 2023{\natexlab{a}}.
\newblock \href {https://aclanthology.org/2023.acl-long.112} {{REV}: Information-theoretic evaluation of free-text rationales}.
\newblock In \emph{Proceedings of the 61st Annual Meeting of the Association for Computational Linguistics (Volume 1: Long Papers)}, pages 2007--2030, Toronto, Canada. Association for Computational Linguistics.

\bibitem[{Chen et~al.(2024{\natexlab{b}})Chen, Saha, and Bansal}]{chen-etal-2024-reconcile}
Justin Chen, Swarnadeep Saha, and Mohit Bansal. 2024{\natexlab{b}}.
\newblock \href {https://doi.org/10.18653/v1/2024.acl-long.381} {{R}e{C}oncile: Round-table conference improves reasoning via consensus among diverse {LLM}s}.
\newblock In \emph{Proceedings of the 62nd Annual Meeting of the Association for Computational Linguistics (Volume 1: Long Papers)}, pages 7066--7085, Bangkok, Thailand. Association for Computational Linguistics.

\bibitem[{Chen et~al.(2024{\natexlab{c}})Chen, Li, Dong, Zhang, Zang, Chen, Duan, Wang, Qiao, Lin et~al.}]{chen2024we}
Lin Chen, Jinsong Li, Xiaoyi Dong, Pan Zhang, Yuhang Zang, Zehui Chen, Haodong Duan, Jiaqi Wang, Yu~Qiao, Dahua Lin, et~al. 2024{\natexlab{c}}.
\newblock Are we on the right way for evaluating large vision-language models?
\newblock \emph{arXiv preprint arXiv:2403.20330}.

\bibitem[{Chen et~al.(2024{\natexlab{d}})Chen, Qin, Zhang, Chen, Xu, and Che}]{chen-etal-2024-m3cot}
Qiguang Chen, Libo Qin, Jin Zhang, Zhi Chen, Xiao Xu, and Wanxiang Che. 2024{\natexlab{d}}.
\newblock M$^3$cot: A novel benchmark for multi-domain multi-step multi-modal chain-of-thought.
\newblock In \emph{Proc. of ACL}.

\bibitem[{Chen et~al.(2024{\natexlab{e}})Chen, Zhang, Luo, D{'}Haro, Tan, and Li}]{chen-etal-2024-unveiling}
Yiming Chen, Chen Zhang, Danqing Luo, Luis~Fernando D{'}Haro, Robby Tan, and Haizhou Li. 2024{\natexlab{e}}.
\newblock \href {https://doi.org/10.18653/v1/2024.findings-acl.80} {Unveiling the achilles{'} heel of {NLG} evaluators: A unified adversarial framework driven by large language models}.
\newblock In \emph{Findings of the Association for Computational Linguistics ACL 2024}, pages 1359--1375, Bangkok, Thailand and virtual meeting. Association for Computational Linguistics.

\bibitem[{Chen et~al.(2023{\natexlab{b}})Chen, Chen, Zhang, Guo, Fang, Huang, Zhang, Geng, Pan, Song, and Chen}]{CCZG+2023}
Zhuo Chen, Jiaoyan Chen, Wen Zhang, Lingbing Guo, Yin Fang, Yufeng Huang, Yichi Zhang, Yuxia Geng, Jeff~Z. Pan, Wenting Song, and Huajun Chen. 2023{\natexlab{b}}.
\newblock {MEAformer: Multi-modal Entity Alignment Transformer for Meta Modality Hybrid}.
\newblock In \emph{In Proc. of the 31st ACM International Conference on Multimedia (MM 2023)}.

\bibitem[{Chen et~al.(2023{\natexlab{c}})Chen, Guo, Fang, Zhang, Chen, Pan, Li, Chen, and Zhang}]{LGFZ+2023}
Zhuo Chen, Lingbing Guo, Yin Fang, Yichi Zhang, Jiaoyan Chen, Jeff~Z. Pan, Yangning Li, Huajun Chen, and Wen Zhang. 2023{\natexlab{c}}.
\newblock {Rethinking Uncertainly Missing and Ambiguous Visual Modality in Multi-Modal Entity Alignment}.
\newblock In \emph{In the 22nd International Semantic Web Conference (ISWC 2023)}, pages 121--139.

\bibitem[{Clinciu et~al.(2021)Clinciu, Eshghi, and Hastie}]{clinciu-etal-2021-study}
Miruna-Adriana Clinciu, Arash Eshghi, and Helen Hastie. 2021.
\newblock \href {https://doi.org/10.18653/v1/2021.eacl-main.202} {A study of automatic metrics for the evaluation of natural language explanations}.
\newblock In \emph{Proceedings of the 16th Conference of the European Chapter of the Association for Computational Linguistics: Main Volume}, pages 2376--2387, Online. Association for Computational Linguistics.

\bibitem[{Dai et~al.(2023)Dai, Li, Li, Tiong, Zhao, Wang, Li, Fung, and Hoi}]{dai2023instructblip}
Wenliang Dai, Junnan Li, Dongxu Li, Anthony Meng~Huat Tiong, Junqi Zhao, Weisheng Wang, Boyang Li, Pascale Fung, and Steven Hoi. 2023.
\newblock \href {https://arxiv.org/abs/2305.06500} {Instructblip: Towards general-purpose vision-language models with instruction tuning}.
\newblock \emph{Preprint}, arXiv:2305.06500.

\bibitem[{Dubey et~al.(2024)Dubey, Jauhri, Pandey, Kadian, Al-Dahle, Letman, Mathur, Schelten, Yang, , and et~al.}]{dubey2024llama3herdmodels}
Abhimanyu Dubey, Abhinav Jauhri, Abhinav Pandey, Abhishek Kadian, Ahmad Al-Dahle, Aiesha Letman, Akhil Mathur, Alan Schelten, Amy Yang, , and et~al. 2024.
\newblock \href {https://arxiv.org/abs/2407.21783} {The llama 3 herd of models}.
\newblock \emph{Preprint}, arXiv:2407.21783.

\bibitem[{Fu et~al.(2024)Fu, Ng, Jiang, and Liu}]{fu-etal-2024-gptscore}
Jinlan Fu, See-Kiong Ng, Zhengbao Jiang, and Pengfei Liu. 2024.
\newblock \href {https://doi.org/10.18653/v1/2024.naacl-long.365} {{GPTS}core: Evaluate as you desire}.
\newblock In \emph{Proceedings of the 2024 Conference of the North American Chapter of the Association for Computational Linguistics: Human Language Technologies (Volume 1: Long Papers)}, pages 6556--6576, Mexico City, Mexico. Association for Computational Linguistics.

\bibitem[{Gao et~al.(2023)Gao, Wu, Blair, and Pagnucco}]{NEURIPS2023_617ff527}
Jingying Gao, Qi~Wu, Alan Blair, and Maurice Pagnucco. 2023.
\newblock \href {https://proceedings.neurips.cc/paper_files/paper/2023/file/617ff5271b2b41dfb217a3b0f1b3d1be-Paper-Datasets_and_Benchmarks.pdf} {Lora: A logical reasoning augmented dataset for visual question answering}.
\newblock In \emph{Advances in Neural Information Processing Systems}, volume~36, pages 30579--30591. Curran Associates, Inc.

\bibitem[{Golovneva et~al.(2022)Golovneva, Chen, Poff, Corredor, Zettlemoyer, Fazel-Zarandi, and Celikyilmaz}]{golovneva2022roscoe}
Olga Golovneva, Moya Chen, Spencer Poff, Martin Corredor, Luke Zettlemoyer, Maryam Fazel-Zarandi, and Asli Celikyilmaz. 2022.
\newblock {ROSCOE}: A suite of metrics for scoring step-by-step reasoning.
\newblock In \emph{In The Eleventh International Conference on Learning Representations.}

\bibitem[{Goyal et~al.(2017)Goyal, Khot, Summers{-}Stay, Batra, and Parikh}]{balanced_vqa_v2}
Yash Goyal, Tejas Khot, Douglas Summers{-}Stay, Dhruv Batra, and Devi Parikh. 2017.
\newblock Making the {V} in {VQA} matter: Elevating the role of image understanding in {V}isual {Q}uestion {A}nswering.
\newblock In \emph{Conference on Computer Vision and Pattern Recognition (CVPR)}.

\bibitem[{Gurari et~al.(2018)Gurari, Li, Stangl, Guo, Lin, Grauman, Luo, and Bigham}]{Gurari_2018_CVPR}
Danna Gurari, Qing Li, Abigale~J. Stangl, Anhong Guo, Chi Lin, Kristen Grauman, Jiebo Luo, and Jeffrey~P. Bigham. 2018.
\newblock Vizwiz grand challenge: Answering visual questions from blind people.
\newblock In \emph{Proceedings of the IEEE Conference on Computer Vision and Pattern Recognition (CVPR)}.

\bibitem[{He et~al.(2024)He, Zhang, and Roth}]{he-etal-2024-socreval}
Hangfeng He, Hongming Zhang, and Dan Roth. 2024.
\newblock \href {https://doi.org/10.18653/v1/2024.findings-naacl.175} {{S}oc{RE}val: Large language models with the socratic method for reference-free reasoning evaluation}.
\newblock In \emph{Findings of the Association for Computational Linguistics: NAACL 2024}, pages 2736--2764, Mexico City, Mexico. Association for Computational Linguistics.

\bibitem[{Hessel et~al.(2021)Hessel, Holtzman, Forbes, Le~Bras, and Choi}]{hessel-etal-2021-clipscore}
Jack Hessel, Ari Holtzman, Maxwell Forbes, Ronan Le~Bras, and Yejin Choi. 2021.
\newblock \href {https://doi.org/10.18653/v1/2021.emnlp-main.595} {{CLIPS}core: A reference-free evaluation metric for image captioning}.
\newblock In \emph{Proceedings of the 2021 Conference on Empirical Methods in Natural Language Processing}, pages 7514--7528, Online and Punta Cana, Dominican Republic. Association for Computational Linguistics.

\bibitem[{Hosseini et~al.(2018)Hosseini, Chambers, Reddy, Holt, Cohen, Johnson, and Steedman}]{journals/tacl/HosseiniCRHCJS18}
Mohammad~Javad Hosseini, Nathanael Chambers, Siva Reddy, Xavier~R. Holt, Shay~B. Cohen, Mark Johnson, and Mark Steedman. 2018.
\newblock \href {http://dblp.uni-trier.de/db/journals/tacl/tacl6.html#HosseiniCRHCJS18} {Learning typed entailment graphs with global soft constraints.}
\newblock \emph{Trans. Assoc. Comput. Linguistics}, 6:703--717.

\bibitem[{Huang et~al.(2023)Huang, Sun, Xie, Li, and Liu}]{NEURIPS2023_f8ad010c}
Kaiyi Huang, Kaiyue Sun, Enze Xie, Zhenguo Li, and Xihui Liu. 2023.
\newblock \href {https://proceedings.neurips.cc/paper_files/paper/2023/file/f8ad010cdd9143dbb0e9308c093aff24-Paper-Datasets_and_Benchmarks.pdf} {T2i-compbench: A comprehensive benchmark for open-world compositional text-to-image generation}.
\newblock In \emph{Advances in Neural Information Processing Systems}, volume~36, pages 78723--78747. Curran Associates, Inc.

\bibitem[{Huang et~al.(2024)Huang, Zhou, Wang, Vougiouklis, Lapata, and Pan}]{HZWV+2024}
Wenyu Huang, Guancheng Zhou, Hongru Wang, Pavlos Vougiouklis, Mirella Lapata, and Jeff~Z. Pan. 2024.
\newblock {Less is More: Making Smaller Language Models Competent Subgraph Retrievers for Multi-hop KGQA}.
\newblock In \emph{In the Findings of the Association for Computational Linguistics: EMNLP 2024 (EMNLP 2024 Findings)}, pages 15787--15803.

\bibitem[{Jacovi et~al.(2024)Jacovi, Bitton, Bohnet, Herzig, Honovich, Tseng, Collins, Aharoni, and Geva}]{jacovi-etal-2024-chain}
Alon Jacovi, Yonatan Bitton, Bernd Bohnet, Jonathan Herzig, Or~Honovich, Michael Tseng, Michael Collins, Roee Aharoni, and Mor Geva. 2024.
\newblock \href {https://doi.org/10.18653/v1/2024.acl-long.254} {A chain-of-thought is as strong as its weakest link: A benchmark for verifiers of reasoning chains}.
\newblock In \emph{Proceedings of the 62nd Annual Meeting of the Association for Computational Linguistics (Volume 1: Long Papers)}, pages 4615--4634, Bangkok, Thailand. Association for Computational Linguistics.

\bibitem[{Jung et~al.(2022)Jung, Qin, Welleck, Brahman, Bhagavatula, Le~Bras, and Choi}]{jung-etal-2022-maieutic}
Jaehun Jung, Lianhui Qin, Sean Welleck, Faeze Brahman, Chandra Bhagavatula, Ronan Le~Bras, and Yejin Choi. 2022.
\newblock \href {https://doi.org/10.18653/v1/2022.emnlp-main.82} {Maieutic prompting: Logically consistent reasoning with recursive explanations}.
\newblock In \emph{Proceedings of the 2022 Conference on Empirical Methods in Natural Language Processing}, pages 1266--1279, Abu Dhabi, United Arab Emirates. Association for Computational Linguistics.

\bibitem[{Kojima et~al.(2022)Kojima, Gu, Reid, Matsuo, and Iwasawa}]{NEURIPS2022_8bb0d291}
Takeshi Kojima, Shixiang~(Shane) Gu, Machel Reid, Yutaka Matsuo, and Yusuke Iwasawa. 2022.
\newblock \href {https://proceedings.neurips.cc/paper_files/paper/2022/file/8bb0d291acd4acf06ef112099c16f326-Paper-Conference.pdf} {Large language models are zero-shot reasoners}.
\newblock In \emph{Advances in Neural Information Processing Systems}, volume~35, pages 22199--22213. Curran Associates, Inc.

\bibitem[{Li et~al.(2024)Li, Li, Li, Chai, and Pan}]{LLLCP2024}
Juncai Li, Ru~Li, Xiaoli Li, Qinghua Chai, and Jeff~Z. Pan. 2024.
\newblock {Inference Helps PLMs Conceptual Understanding: Improving the Abstract Inference Ability with Hierarchical Conceptual Entailment Graphs}.
\newblock In \emph{In Proc. of the 2024 Conference on Empirical Methods in Natural Language Processing (EMNLP 2024)}, pages 22088--22104.

\bibitem[{Li et~al.(2023{\natexlab{a}})Li, Li, Savarese, and Hoi}]{li2023blip2}
Junnan Li, Dongxu Li, Silvio Savarese, and Steven Hoi. 2023{\natexlab{a}}.
\newblock {BLIP-2:} bootstrapping language-image pre-training with frozen image encoders and large language models.
\newblock In \emph{ICML}.

\bibitem[{Li et~al.(2023{\natexlab{b}})Li, Lin, Zhang, Fu, Chen, Lou, and Chen}]{li-etal-2023-making}
Yifei Li, Zeqi Lin, Shizhuo Zhang, Qiang Fu, Bei Chen, Jian-Guang Lou, and Weizhu Chen. 2023{\natexlab{b}}.
\newblock \href {https://doi.org/10.18653/v1/2023.acl-long.291} {Making language models better reasoners with step-aware verifier}.
\newblock In \emph{Proceedings of the 61st Annual Meeting of the Association for Computational Linguistics (Volume 1: Long Papers)}, pages 5315--5333, Toronto, Canada. Association for Computational Linguistics.

\bibitem[{Liu et~al.(2023{\natexlab{a}})Liu, Emerson, and Collier}]{Liu2022VisualSR}
Fangyu Liu, Guy Edward~Toh Emerson, and Nigel Collier. 2023{\natexlab{a}}.
\newblock Visual spatial reasoning.
\newblock \emph{Transactions of the Association for Computational Linguistics}.

\bibitem[{Liu et~al.(2024)Liu, Li, Li, Li, Zhang, Shen, and Lee}]{liu2024llavanext}
Haotian Liu, Chunyuan Li, Yuheng Li, Bo~Li, Yuanhan Zhang, Sheng Shen, and Yong~Jae Lee. 2024.
\newblock \href {https://llava-vl.github.io/blog/2024-01-30-llava-next/} {Llava-next: Improved reasoning, ocr, and world knowledge}.

\bibitem[{Liu et~al.(2023{\natexlab{b}})Liu, Iter, Xu, Wang, Xu, and Zhu}]{liu-etal-2023-g}
Yang Liu, Dan Iter, Yichong Xu, Shuohang Wang, Ruochen Xu, and Chenguang Zhu. 2023{\natexlab{b}}.
\newblock \href {https://doi.org/10.18653/v1/2023.emnlp-main.153} {{G}-eval: {NLG} evaluation using gpt-4 with better human alignment}.
\newblock In \emph{Proceedings of the 2023 Conference on Empirical Methods in Natural Language Processing}, pages 2511--2522, Singapore. Association for Computational Linguistics.

\bibitem[{Lu et~al.(2022)Lu, Mishra, Xia, Qiu, Chang, Zhu, Tafjord, Clark, and Kalyan}]{lu2022learn}
Pan Lu, Swaroop Mishra, Tony Xia, Liang Qiu, Kai-Wei Chang, Song-Chun Zhu, Oyvind Tafjord, Peter Clark, and Ashwin Kalyan. 2022.
\newblock Learn to explain: Multimodal reasoning via thought chains for science question answering.
\newblock In \emph{The 36th Conference on Neural Information Processing Systems (NeurIPS)}.

\bibitem[{Lyu et~al.(2023)Lyu, Havaldar, Stein, Zhang, Rao, Wong, Apidianaki, and Callison-Burch}]{lyu-etal-2023-faithful}
Qing Lyu, Shreya Havaldar, Adam Stein, Li~Zhang, Delip Rao, Eric Wong, Marianna Apidianaki, and Chris Callison-Burch. 2023.
\newblock \href {https://doi.org/10.18653/v1/2023.ijcnlp-main.20} {Faithful chain-of-thought reasoning}.
\newblock In \emph{Proceedings of the 13th International Joint Conference on Natural Language Processing and the 3rd Conference of the Asia-Pacific Chapter of the Association for Computational Linguistics (Volume 1: Long Papers)}, pages 305--329, Nusa Dua, Bali. Association for Computational Linguistics.

\bibitem[{OpenAI et~al.(2023)OpenAI, Achiam, Adler, Agarwal, Ahmad, Akkaya, Aleman, Almeida, Altenschmidt, Altman, Anadkat, Avila, and et~al.}]{openai2023gpt4}
OpenAI, :~Josh Achiam, Steven Adler, Sandhini Agarwal, Lama Ahmad, Ilge Akkaya, Florencia~Leoni Aleman, Diogo Almeida, Janko Altenschmidt, Sam Altman, Shyamal Anadkat, Red Avila, and et~al. 2023.
\newblock \href {https://arxiv.org/abs/2303.08774} {Gpt-4 technical report}.
\newblock \emph{Preprint}, arXiv:2303.08774.

\bibitem[{Pan et~al.(2023)Pan, Razniewski, Kalo, Singhania, Chen, Dietze, Jabeen, Omeliyanenko, Zhang, Lissandrini, Biswas, de~Melo, Bonifati, Vakaj, Dragoni, and Graux}]{PRKS+2023}
Jeff~Z. Pan, Simon Razniewski, Jan-Christoph Kalo, Sneha Singhania, Jiaoyan Chen, Stefan Dietze, Hajira Jabeen, Janna Omeliyanenko, Wen Zhang, Matteo Lissandrini, Russa Biswas, Gerard de~Melo, Angela Bonifati, Edlira Vakaj, Mauro Dragoni, and Damien Graux. 2023.
\newblock {Large Language Models and Knowledge Graphs: Opportunities and Challenges}.
\newblock pages 1--38.

\bibitem[{Pan et~al.(2017{\natexlab{a}})Pan, Calvanese, Eiter, Horrocks, Kifer, Lin, and Zhao}]{PCEH+2017}
J.Z. Pan, D.~Calvanese, T.~Eiter, I.~Horrocks, M.~Kifer, F.~Lin, and Y.~Zhao, editors. 2017{\natexlab{a}}.
\newblock \emph{{Reasoning Web: Logical Foundation of Knowledge Graph Construction and Querying Answering}}.
\newblock Springer.

\bibitem[{Pan et~al.(2017{\natexlab{b}})Pan, Vetere, Gomez-Perez, and Wu}]{PVGW2017}
J.Z. Pan, G.~Vetere, J.M. Gomez-Perez, and H.~Wu, editors. 2017{\natexlab{b}}.
\newblock \emph{{Exploiting Linked Data and Knowledge Graphs for Large Organisations}}.
\newblock Springer.

\bibitem[{Prasad et~al.(2023)Prasad, Saha, Zhou, and Bansal}]{prasad-etal-2023-receval}
Archiki Prasad, Swarnadeep Saha, Xiang Zhou, and Mohit Bansal. 2023.
\newblock \href {https://doi.org/10.18653/v1/2023.emnlp-main.622} {{R}e{CE}val: Evaluating reasoning chains via correctness and informativeness}.
\newblock In \emph{Proceedings of the 2023 Conference on Empirical Methods in Natural Language Processing}, pages 10066--10086, Singapore. Association for Computational Linguistics.

\bibitem[{Saadat-Yazdi et~al.(2023)Saadat-Yazdi, Pan, and Kokciyan}]{SPK2023}
Ameer Saadat-Yazdi, Jeff~Z. Pan, and Nadin Kokciyan. 2023.
\newblock {Uncovering Implicit Inferences for Improved Relational Argument Mining}.
\newblock In \emph{In the 17th Conference of the European Chapter of the Association for Computational Linguistics (EACL 2023)}, pages 2484--2495.

\bibitem[{Saparov and He(2023)}]{SaparovHe2023}
Abulhair Saparov and He~He. 2023.
\newblock Language models are greedy reasoners: A systematic formal analysis of chain-of-thought.
\newblock In \emph{International Conference on Learning Representations}.

\bibitem[{Shao et~al.(2024)Shao, Qian, Xiao, Song, Zong, Wang, Liu, and Li}]{shao2024visual}
Hao Shao, Shengju Qian, Han Xiao, Guanglu Song, Zhuofan Zong, Letian Wang, Yu~Liu, and Hongsheng Li. 2024.
\newblock \href {https://arxiv.org/abs/2403.16999} {Visual cot: Unleashing chain-of-thought reasoning in multi-modal language models}.
\newblock \emph{Preprint}, arXiv:2403.16999.

\bibitem[{Shen et~al.(2024)Shen, Vougiouklis, Diao, Vyas, Ji, and Pan}]{SVDV+2024}
Zhili Shen, Pavlos Vougiouklis, Chenxin Diao, Kaustubh Vyas, Yuanyi Ji, and Jeff~Z. Pan. 2024.
\newblock {Improving Retrieval-augmented Text-to-SQL with AST-based Ranking and Schema Pruning}.
\newblock In \emph{In Proc. of the 2024 Conference on Empirical Methods in Natural Language Processing (EMNLP 2024)}, pages 7865--7879.

\bibitem[{Somers(1962)}]{Somers1962ANA}
Robert~H. Somers. 1962.
\newblock \href {https://api.semanticscholar.org/CorpusID:144232738} {A new asymmetric measure of association for ordinal variables.}
\newblock \emph{American Sociological Review}, 27:799.

\bibitem[{Spearman(1904)}]{ca468a70-0be4-389a-b0b9-5dd1ff52b33f}
C.~Spearman. 1904.
\newblock \href {http://www.jstor.org/stable/1412159} {The proof and measurement of association between two things}.
\newblock \emph{The American Journal of Psychology}, 15(1):72--101.

\bibitem[{Team et~al.(2024)Team, Georgiev, Lei, Burnell, Bai, Gulati, Tanzer, Vincent, Pan, Wang, Mariooryad, Ding, Geng, Alcober, and et~al.}]{geminiteam2024gemini15unlockingmultimodal}
Gemini Team, Petko Georgiev, Ving~Ian Lei, Ryan Burnell, Libin Bai, Anmol Gulati, Garrett Tanzer, Damien Vincent, Zhufeng Pan, Shibo Wang, Soroosh Mariooryad, Yifan Ding, Xinyang Geng, Fred Alcober, and et~al. 2024.
\newblock \href {https://arxiv.org/abs/2403.05530} {Gemini 1.5: Unlocking multimodal understanding across millions of tokens of context}.
\newblock \emph{Preprint}, arXiv:2403.05530.

\bibitem[{Tong et~al.(2024)Tong, Liu, Zhai, Ma, LeCun, and Xie}]{tong2024eyes}
Shengbang Tong, Zhuang Liu, Yuexiang Zhai, Yi~Ma, Yann LeCun, and Saining Xie. 2024.
\newblock \href {https://arxiv.org/abs/2401.06209} {Eyes wide shut? exploring the visual shortcomings of multimodal llms}.
\newblock \emph{Preprint}, arXiv:2401.06209.

\bibitem[{Turpin et~al.(2023)Turpin, Michael, Perez, and Bowman}]{NEURIPS2023_ed3fea90}
Miles Turpin, Julian Michael, Ethan Perez, and Samuel Bowman. 2023.
\newblock \href {https://proceedings.neurips.cc/paper_files/paper/2023/file/ed3fea9033a80fea1376299fa7863f4a-Paper-Conference.pdf} {Language models don\textquotesingle t always say what they think: Unfaithful explanations in chain-of-thought prompting}.
\newblock In \emph{Advances in Neural Information Processing Systems}, volume~36, pages 74952--74965. Curran Associates, Inc.

\bibitem[{Vougiouklis et~al.(2023)Vougiouklis, Papasarantopoulos, Zheng, Tuckey, Diao, Shen, and Pan}]{VPZT+2023}
Pavlos Vougiouklis, Nikos Papasarantopoulos, Danna Zheng, David Tuckey, Chenxin Diao, Zhili Shen, and Jeff~Z Pan. 2023.
\newblock {FastRAT: Fast and Efficient Cross-lingual Text-to-SQL Semantic Parsing}.
\newblock In \emph{In Proc. of the 13th International Joint Conference on Natural Language Processing and the 3rd Conference of the Asia-Pacific Chapter of the Association for Computational Linguistics (JCNLP-AACL 2023)}.

\bibitem[{Vyas et~al.(2025)Vyas, Graux, Yang, Montella, Diao, Zhou, Vougiouklis, Lai, Ren, Li, and Pan}]{VGYM+2025}
Kaustubh Vyas, Damien Graux, Yijun Yang, Sébastien Montella, Chenxin Diao, Wendi Zhou, Pavlos Vougiouklis, Ruofei Lai, Yang Ren, Keshuang Li, and Jeff~Z. Pan. 2025.
\newblock {From An LLM Swarm To A PDDL-Empowered HIVE: Planning Self-Executed Instructions In A Multi-Modal Jungle}.
\newblock In \emph{In Proc. of the International Conference on Learning Representations (ICLR 2025).}

\bibitem[{Wang et~al.(2024{\natexlab{a}})Wang, Qin, Lin, Pan, and Wong}]{WQLPW2024}
Hongru Wang, Yujia Qin, Yankai Lin, Jeff~Z. Pan, and Kam-Fai Wong. 2024{\natexlab{a}}.
\newblock {Empowering Large Language Models: Tool Learning for Real-world Interaction}.
\newblock In \emph{In Proc. of the 47th International ACM SIGIR Conference on Research and Development in Information Retrieval (SIGIR '24)}, pages 2983 -- 2986.

\bibitem[{Wang et~al.(2024{\natexlab{b}})Wang, Wang, Xue, Xia, Cao, Liu, Pan, and Wong}]{WWXX+2024}
Hongru Wang, Rui Wang, Boyang Xue, Heming Xia, Jingtao Cao, Zeming Liu, Jeff~Z. Pan, and Kam-Fai Wong. 2024{\natexlab{b}}.
\newblock {MetaBench: Planning of Multiple APIs from Various APPs for Complex User Instruction}.
\newblock In \emph{In Proc. of the 2024 Conference on Empirical Methods in Natural Language Processing (EMNLP 2024)}, pages 15322--15336.

\bibitem[{Wang et~al.(2024{\natexlab{c}})Wang, Chen, Hu, Yang, Liu, Shen, Wei, Zhang, Gu, Zhou, Pan, Zhang, and Chen}]{WCHY+2024}
Junjie Wang, Mingyang Chen, Binbin Hu, Dan Yang, Ziqi Liu, Yue Shen, Peng Wei, Zhiqiang Zhang, Jinjie Gu, Jun Zhou, Jeff~Z. Pan, Wen Zhang, and Huajun Chen. 2024{\natexlab{c}}.
\newblock {Learning to Plan for Retrieval-Augmented Large Language Models from Knowledge Graphs}.
\newblock In \emph{In the Findings of the Association for Computational Linguistics: EMNLP 2024 (EMNLP 2024 Findings)}, pages 7813--7835.

\bibitem[{Wei et~al.(2022)Wei, Wang, Schuurmans, Bosma, ichter, Xia, Chi, Le, and Zhou}]{NEURIPS2022_9d560961}
Jason Wei, Xuezhi Wang, Dale Schuurmans, Maarten Bosma, brian ichter, Fei Xia, Ed~Chi, Quoc~V Le, and Denny Zhou. 2022.
\newblock \href {https://proceedings.neurips.cc/paper_files/paper/2022/file/9d5609613524ecf4f15af0f7b31abca4-Paper-Conference.pdf} {Chain-of-thought prompting elicits reasoning in large language models}.
\newblock In \emph{Advances in Neural Information Processing Systems}, volume~35, pages 24824--24837. Curran Associates, Inc.

\bibitem[{Welleck et~al.(2022)Welleck, Liu, Lu, Hajishirzi, and Choi}]{NEURIPS2022_1fc548a8}
Sean Welleck, Jiacheng Liu, Ximing Lu, Hannaneh Hajishirzi, and Yejin Choi. 2022.
\newblock \href {https://proceedings.neurips.cc/paper_files/paper/2022/file/1fc548a8243ad06616eee731e0572927-Paper-Conference.pdf} {Naturalprover: Grounded mathematical proof generation with language models}.
\newblock In \emph{Advances in Neural Information Processing Systems}, volume~35, pages 4913--4927. Curran Associates, Inc.

\bibitem[{Wu et~al.(2024)Wu, Huang, Hu, Hua, Qi, Chen, and Pan}]{WHHH+2024}
Yike Wu, Yi~Huang, Nan Hu, Yuncheng Hua, Guilin Qi, Jiaoyan Chen, and Jeff~Z. Pan. 2024.
\newblock {CoTKR: Chain-of-Thought Enhanced Knowledge Rewriting for Complex Knowledge Graph Question Answering}.
\newblock In \emph{In Proc. of the 2024 Conference on Empirical Methods in Natural Language Processing (EMNLP 2024)}, pages 3501--3520.

\bibitem[{Xia et~al.(2024)Xia, Li, Liu, Wu, and Liu}]{xia_evaluating}
Shijie Xia, Xuefeng Li, Yixin Liu, Tongshuang Wu, and Pengfei Liu. 2024.
\newblock Evaluating mathematical reasoning beyond accuracy.
\newblock \emph{arXiv preprint arXiv:2404.05692}.

\bibitem[{Yao et~al.(2024)Yao, Yu, Zhang, Wang, Cui, Zhu, Cai, Li, Zhao, He et~al.}]{yao2024minicpm}
Yuan Yao, Tianyu Yu, Ao~Zhang, Chongyi Wang, Junbo Cui, Hongji Zhu, Tianchi Cai, Haoyu Li, Weilin Zhao, Zhihui He, et~al. 2024.
\newblock Minicpm-v: A gpt-4v level mllm on your phone.
\newblock \emph{arXiv preprint arXiv:2408.01800}.

\bibitem[{Yu et~al.(2024)Yu, Yang, Li, Wang, Lin, Liu, Wang, and Wang}]{yu2024mm}
Weihao Yu, Zhengyuan Yang, Linjie Li, Jianfeng Wang, Kevin Lin, Zicheng Liu, Xinchao Wang, and Lijuan Wang. 2024.
\newblock Mm-vet: Evaluating large multimodal models for integrated capabilities.
\newblock In \emph{International conference on machine learning}. PMLR.

\bibitem[{Zhang et~al.(2024)Zhang, Zhang, Li, Zhao, Karypis, and Smola}]{zhang2024multimodalchainofthoughtreasoninglanguage}
Zhuosheng Zhang, Aston Zhang, Mu~Li, Hai Zhao, George Karypis, and Alex Smola. 2024.
\newblock \href {https://arxiv.org/abs/2302.00923} {Multimodal chain-of-thought reasoning in language models}.
\newblock \emph{Preprint}, arXiv:2302.00923.

\bibitem[{Zheng et~al.(2024)Zheng, Lapata, and Pan}]{ZLP2024}
Danna Zheng, Mirella Lapata, and Jeff~Z. Pan. 2024.
\newblock {Archer: A Human-Labeled Text-to-SQL Dataset with Arithmetic, Commonsense and Hypothetical Reasoning}.
\newblock In \emph{In Proc. of the 18th Conference of the European Chapter of the Association for Computational Linguistics (EACL 2024)}.

\bibitem[{Zheng et~al.(2023)Zheng, Chiang, Sheng, Zhuang, Wu, Zhuang, Lin, Li, Li, Xing, Zhang, Gonzalez, and Stoica}]{zheng2023judging}
Lianmin Zheng, Wei-Lin Chiang, Ying Sheng, Siyuan Zhuang, Zhanghao Wu, Yonghao Zhuang, Zi~Lin, Zhuohan Li, Dacheng Li, Eric.~P Xing, Hao Zhang, Joseph~E. Gonzalez, and Ion Stoica. 2023.
\newblock \href {https://arxiv.org/abs/2306.05685} {Judging llm-as-a-judge with mt-bench and chatbot arena}.
\newblock \emph{Preprint}, arXiv:2306.05685.

\bibitem[{Zhou et~al.(2024)Zhou, Li, Vougiouklis, Steedman, and Pan}]{ZLVSP2024}
Wendi Zhou, Tianyi Li, Pavlos Vougiouklis, Mark Steedman, and Jeff~Z. Pan. 2024.
\newblock {A Usage-centric Take on Intent Understanding in E-Commerce}.
\newblock In \emph{In Proc. of the 2024 Conference on Empirical Methods in Natural Language Processing (EMNLP 2024)}, pages 228--236.

\bibitem[{Zhu et~al.(2016)Zhu, Groth, Bernstein, and Fei-Fei}]{zhu2016cvpr}
Yuke Zhu, Oliver Groth, Michael Bernstein, and Li~Fei-Fei. 2016.
\newblock {Visual7W: Grounded Question Answering in Images}.
\newblock In \emph{{IEEE Conference on Computer Vision and Pattern Recognition}}.

\end{thebibliography}

\clearpage
\appendix

\section{MiCEval Dataset}
\label{app1}

\subsection{Source Datasets}
\label{app1.1}
\begin{figure}[h]
    \centering
    \includegraphics[width=1\linewidth]{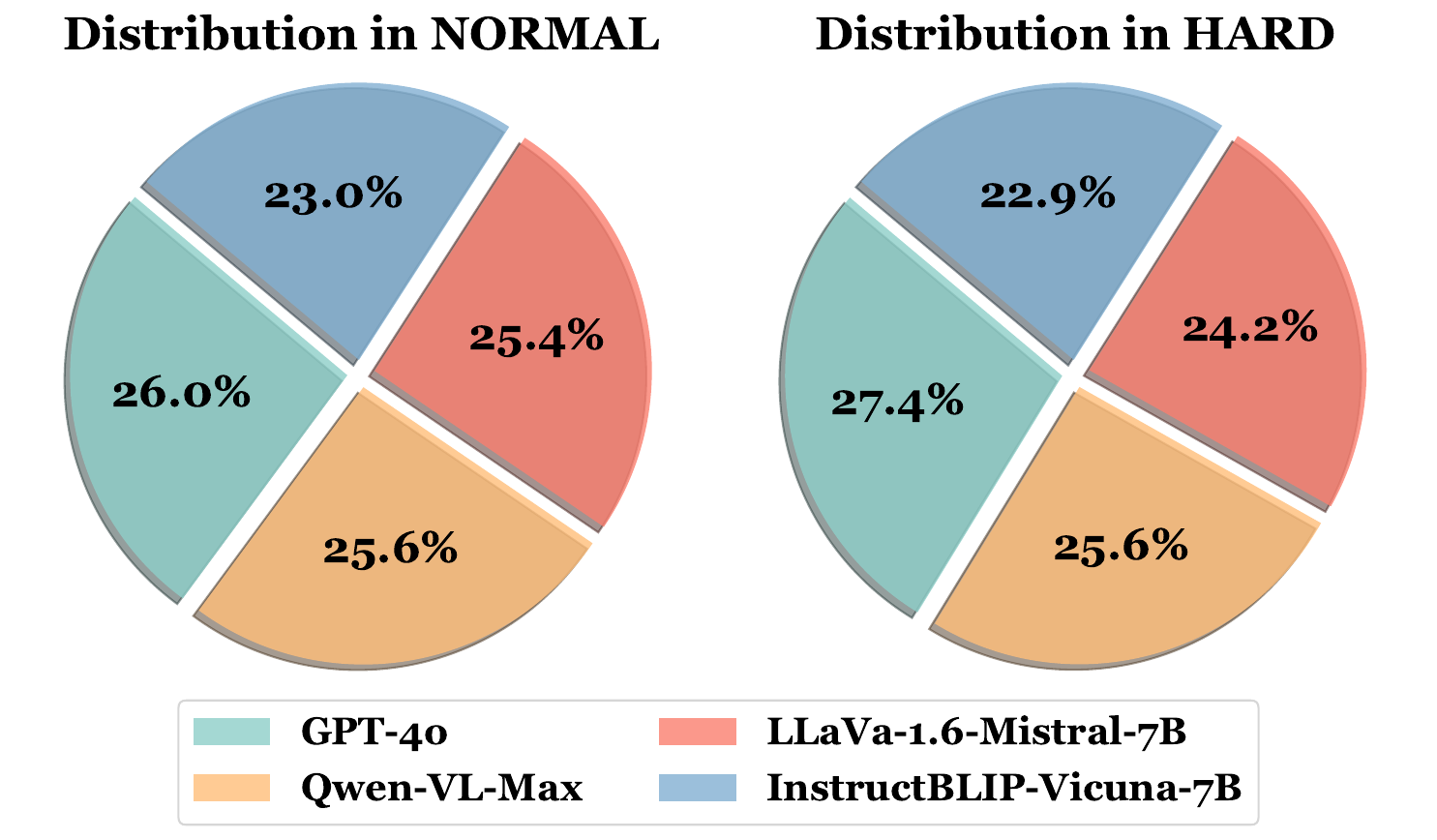}
    \caption{The MCoT generators distribution of each splits.}
    \label{fig:generator-source}
\end{figure}

\begin{figure*}[!t]
    \centering
    \includegraphics[width=0.8\linewidth]{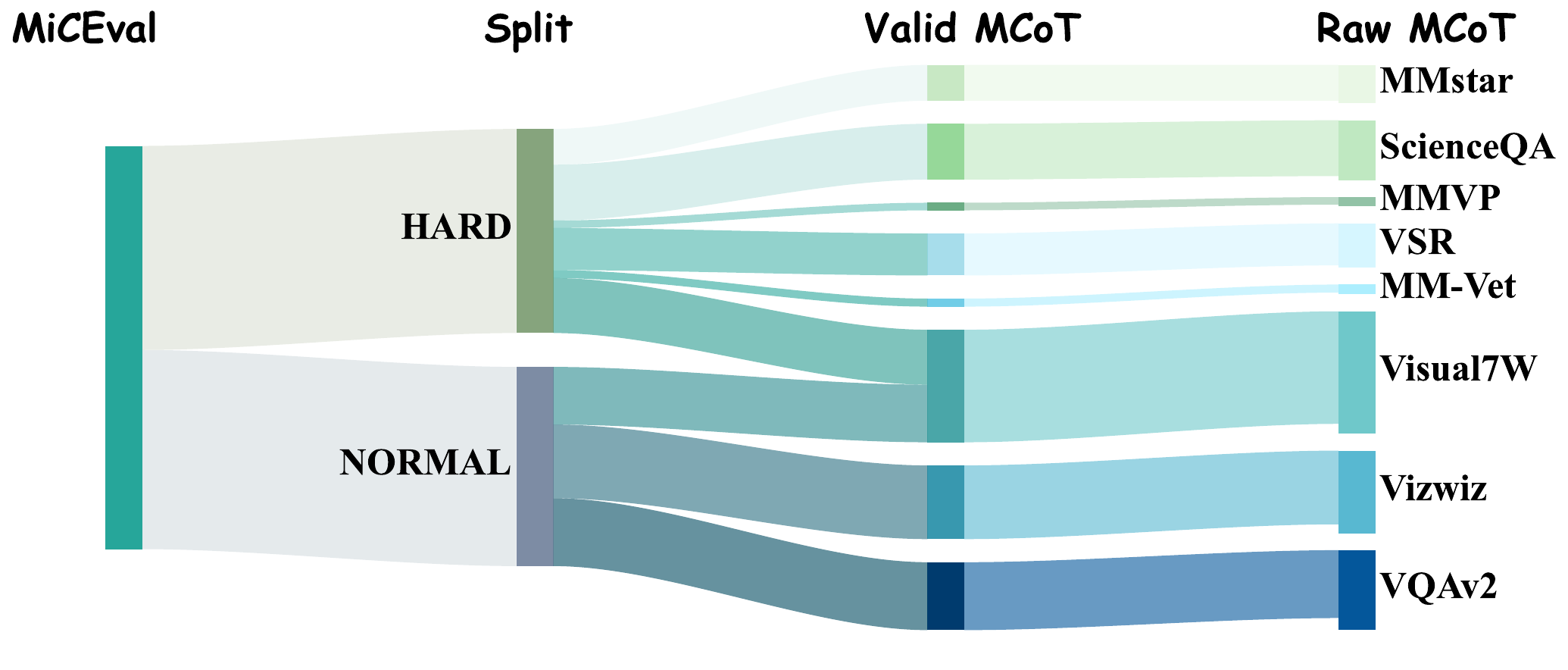}
    \caption{The MCoT generators distribution of each splits.}
    \label{fig:question-source}
\end{figure*}

Detailed information of the all source datasets for MiCEval are provided in Table~\ref{tab:sample-source}. The distribution of source datasets are shown in Figure~\ref{fig:question-source}, and the distribution of MCoT-generators is also provided in Figure~\ref{fig:generator-source}. Table~\ref{tab:hard-sample} and Table~\ref{tab:normal-sample} shows randomly sampled questions from both MiCEval-\textsc{Hard} and MiCEval-\textsc{Normal}.

\subsection{Supplement to Data Collection}
\label{app1.2}
\noindent \textbf{Split Based on Question Difficulty.}
1,500 open-ended questions were randomly sampled from 8 datasets: Visual7W dataset \cite{zhu2016cvpr}, VSR dataset \cite{Liu2022VisualSR}, ScienceQA dataset \cite{lu2022learn}, VQAv2 dataset \cite{balanced_vqa_v2}, Vizwiz dataset \cite{Gurari_2018_CVPR}, MMVP dataset \cite{tong2024eyes}, MM-Vet dataset \cite{yu2024mm}, and MMstar dataset \cite{chen2024we}. Two authors manually filtered out duplicated questions and conducted coarse-grained annotations based on the difficulty of the questions and source datasets. As a result, the dataset was divided into two splits, \textsc{Hard} and \textsc{Normal}, with 350 questions in each split.

\noindent \textbf{MLLMs as MCoT Answer Generators.}
We used four representative MLLMs to generate MCoT answers: Qwen-VL-Max \cite{Qwen-VL}, GPT-4o \cite{openai2023gpt4}, LLaVA-1.6-Mistral-7B \cite{liu2024llavanext}, and InstructBLIP-Vicuna-7B \cite{dai2023instructblip}. By selecting models with varying performance levels, we aimed to capture a diverse range of CoTs. Following \citet{zhang2024multimodalchainofthoughtreasoninglanguage}, we manually crafted several MCoT answer generation prompts tailored to each dataset. Inspired by \citet{jacovi-etal-2024-chain}, we also randomly divided each split into five parts, with four MLLMs answering one part independently, and one part being answered by each MLLMs. Our aim was to enhance the dataset's flexibility and its potential for analysis and evaluation across a wide range of MLLMs. Ultimately, we obtained 1,000 MCoT answers generated by the MLLMs.

\begin{table}[!tpb]
\centering
\resizebox{\columnwidth}{!}{
    \begin{tabular}{l|ccc}
    \toprule
      \multicolumn{1}{l}{\textbf{\shortstack{Complexity Error Types\\~}}} & \textbf{\shortstack{{Description Step}\\{(N=1,751)}}} & \textbf{\shortstack{{Reasoning Step}\\{(N=1,057)}}} & \textbf{\shortstack{{Both Step}\\{(N=81)}}}\\
      \midrule
        Entity False & 7.2\% & - & 2.5\%  \\  
        Attribute False & 10.5\% & - & 9.9\%  \\ 
        Spatial Relationship False & 1.7\% & - & 0.0\% \\ 
        Non-spatial Relationship False & 0.9\% & - & 0.0\%  \\ 
        \midrule
        Inter-step Incorrect & - & 29.6\% & 4.9\%  \\ 
        Intra-step Incorrect & - & 2.5\% & 13.6\%   \\ 
        \midrule
        Image Irrelevant & 1.6\% & - & 0.0\%  \\ 
        Logic Irrelevant & 4.9\% & 4.4\% & 12.3\%  \\ 
        \midrule
        Uninformative & - & 8.7\% & 4.9\% \\  
    \bottomrule
    \end{tabular}
}
\setlength{\belowcaptionskip}{0.2cm}
\caption{The percentage of complexity error types on different MCoT steps.}
\label{tab:step-category}
\end{table}

\subsection{Dataset Analysis}
\label{app1.3}
\noindent \textbf{{Visual and Language Modalities of MLLMs.}} In Table~\ref{tab:step-category}, we summarize the proportions of fine-grained error categories at the step level. For description steps, 7.2\% are classified as entity false, while 10.5\% are labeled as attribute false. This suggests that current MLLMs still have room for improvement in accurately extracting information from the visual modality, particularly when detecting entities and their attributes. For reasoning steps, 29.6\% contain inter-step incorrect errors, indicating that MLLMs continue to struggle with reasoning, especially in maintaining logical consistency between steps. Although the number of both steps is relatively small compared to description and reasoning steps, the proportion of intra-step incorrect steps reaches 13.6\%, suggesting that MLLMs are more prone to internal logical errors when generating mixed steps that combine both visual information and logical reasoning, compared to the other two types of steps.

\begin{figure}[h]
    \centering
    \begin{subfigure}{0.49\linewidth}
        \centering
        \includegraphics[width=\linewidth]{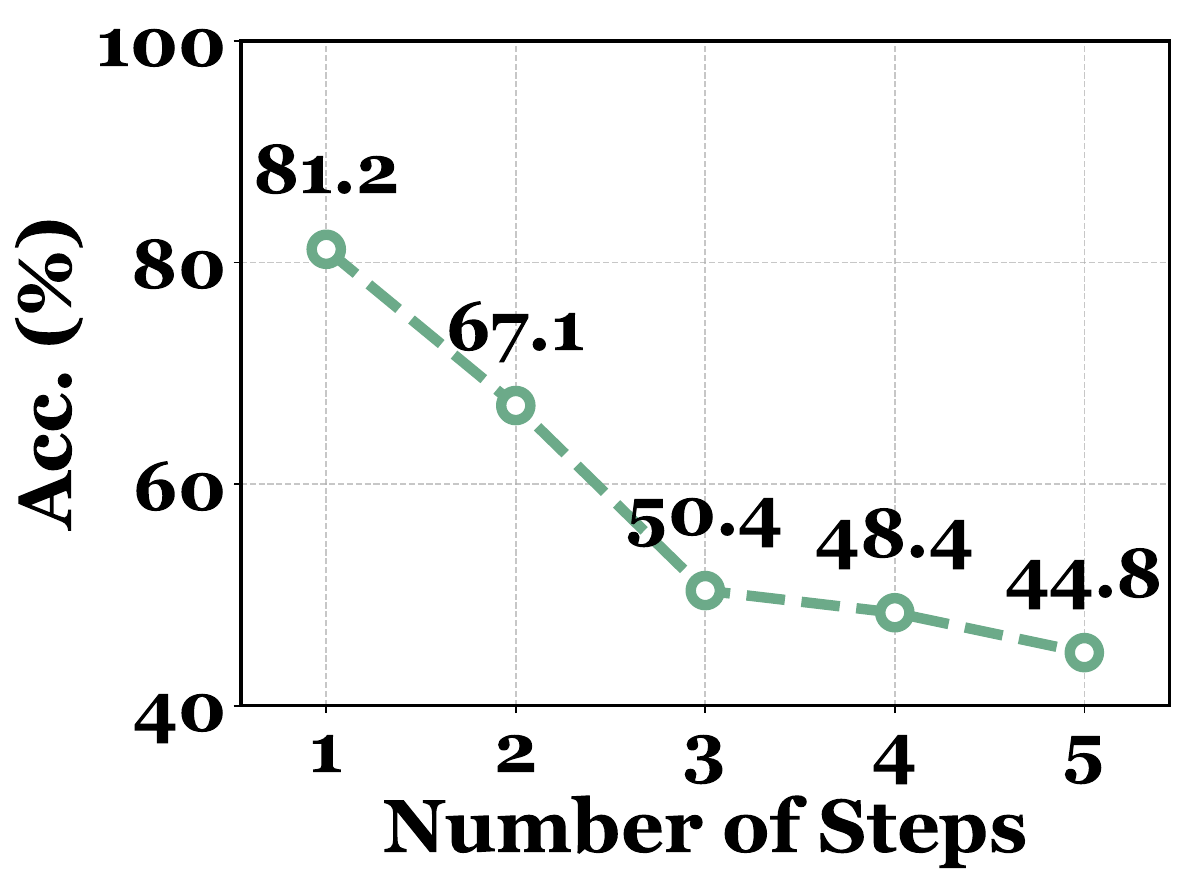}
        \caption{Step Correctness}
        \label{fig:step-correctness}
    \end{subfigure}
    \hfill
    \begin{subfigure}{0.49\linewidth}
        \centering
        \includegraphics[width=\linewidth]{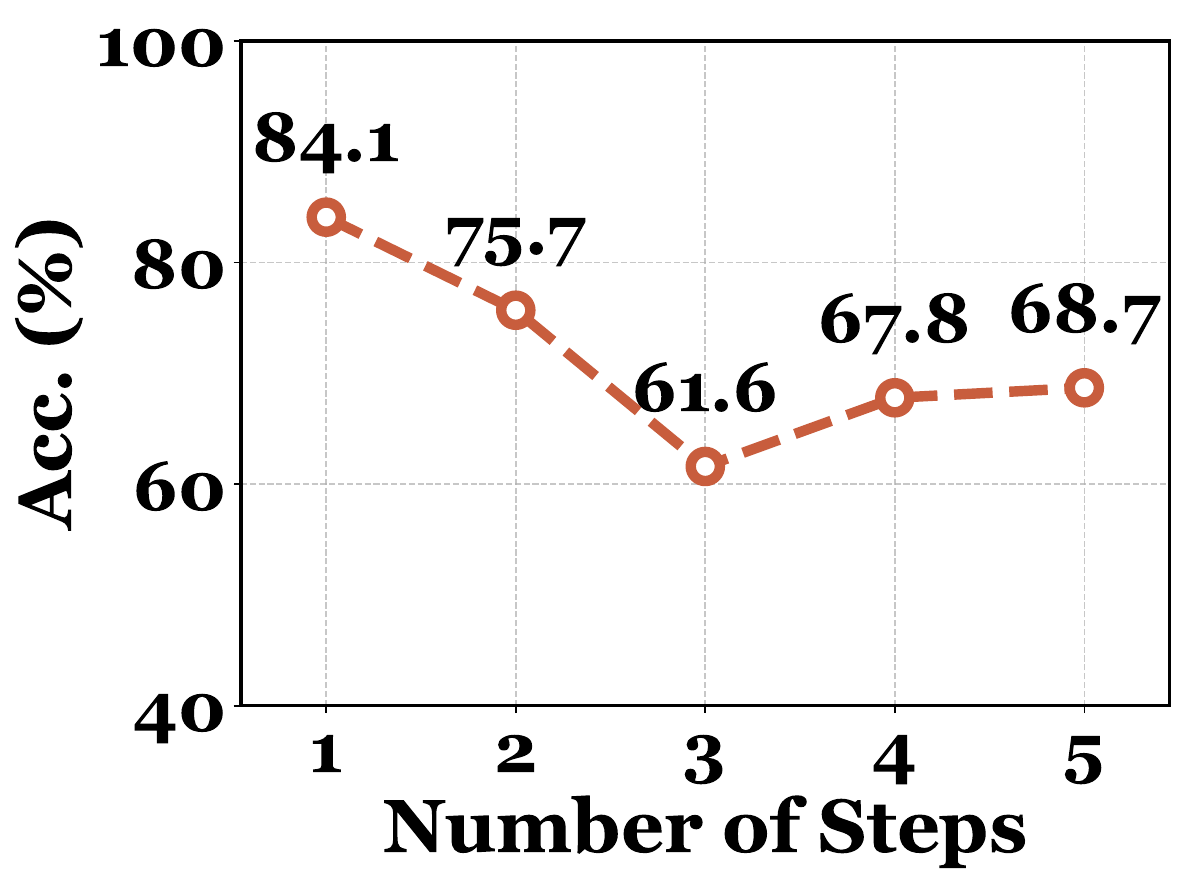}
        \caption{Step Prediction}
        \label{fig:mcot-step}
    \end{subfigure}

    \caption{(a) The Relationship between MCoT answers Correctness and the number of steps. (b) The Relationship between Prediction Correctness and the number of steps.}
    \label{fig:steps-curve}
\end{figure}

\noindent \textbf{{Single-step MCoT Answers.}} 
In Figure~\ref{fig:step-correctness}, we present the relationship between MCoT correctness and the number of steps in an MCoT. It is observed that MLLM achieves the highest quality when generating single-step MCoTs, with 81.2\% being classified as High-Quality MCoTs. However, as the number of steps increases, the proportion of High-Quality MCoTs gradually declines, eventually to 44.8\%.

\noindent \textbf{{High-quality MCoT Answers \& Final Predictions.}} Inspired by \citet{jung-etal-2022-maieutic}, we aim to analyze the relationship between MCoT correctness and the final prediction of the MCoT, specifically its answer to the question. To this end, we conducted additional annotations for the MCoT final predictions. For detailed information about this annotation task and inter-rater agreement, please refer to the App.~\ref{app2.3}. Based on our annotations of MCoT answer correctness and final prediction correctness, we analyzed their relationship. As shown in Table~\ref{tab:correctness-prediction}, our results align with the findings of \citet{jung-etal-2022-maieutic}, where high-quality MCoT answers lead to a final prediction accuracy of 90.4\%, while 42.8\% of MCoTs with errors still produce correct final predictions.

\begin{table}[!tpb]
\scriptsize
    \centering
    \begin{tabular}{c|c|c}
        \toprule
        & Correct Prediction & Incorrect Prediction \\
        \midrule
        High-quality MCoT answer & 90.4\% & 9.6\% \\
        Low-quality MCoT answer & 42.8\% & 57.2\% \\
        \bottomrule
    \end{tabular}
    \caption{The correlation between the correctness of MCoT answer and Prediction. High-quality MCoT answer presents fully correct.}
    \label{tab:correctness-prediction}
\end{table}

\noindent\textbf{Errors \& Final Prediction.} We calculated the proportions of fine-grained error categories when the final prediction is incorrect.  Table~\ref{tab:prediction-categopry} shows that Inter-step Incorrect, Attribute False, and Entity False are more likely to lead to incorrect final predictions in MCoT answer. In particular, the proportion of wrong predictions for MCoT answers with Inter-step Incorrect reaches as high as 60.3\%.

\begin{table}[!tpb]
\centering
\resizebox{\columnwidth}{!}{
    \begin{tabular}{l|cc}
    \toprule
      \multicolumn{1}{l}{\textbf{\shortstack{Complexity Category}}} & \textbf{Correct Prediction} & \textbf{Incorrect Prediction}\\
      \midrule
        Entity False & {3.86\%} & {26.60\%} \\  
        Attribute False & {8.70\%} & {36.50\%} \\ 
        Spatial Relationship False & {0.97\%} & {7.45\%}  \\ 
        Non-spatial Relationship False & {0.81\%} & {2.48\%}  \\ 
        \midrule
        Inter-step Incorrect & {7.89\%} & {60.30\%}   \\ 
        Intra-step Incorrect & {1.13\%} & {9.57\%}    \\ 
        \midrule
        Image Irrelevant & {0.64\%} & {6.74\%} \\ 
        Logic Irrelevant & {4.51\%} & {11.70\%} \\ 
        \midrule
        Uninformative & {10.60\%} & {9.57\%}  \\  
        \bottomrule
    \end{tabular}
}
\setlength{\belowcaptionskip}{0.2cm}
\caption{The percentage of complexity error types on correct and incorrect predictions.}
\label{tab:prediction-categopry}
\end{table}

\noindent \textbf{Number of Steps \& Final Prediction.}
Figure~\ref{fig:mcot-step} illustrates the relationship between the number of MCoT steps and the accuracy of the final predictions. We observe that MLLMs achieve their highest accuracy, 84.1\%, when generating single-step MCoT answers. As the number of steps increases, the accuracy declines gradually. However, when the number of steps reaches four or more, the accuracy of the final predictions begins to improve. We attribute this improvement to the positive impact of more detailed and comprehensive reasoning, which eventually outweighs the negative effects of accumulating errors from additional steps.

\begin{figure}[h]
    \centering
    \begin{subfigure}{0.49\columnwidth}
        \centering
        \includegraphics[width=\linewidth]{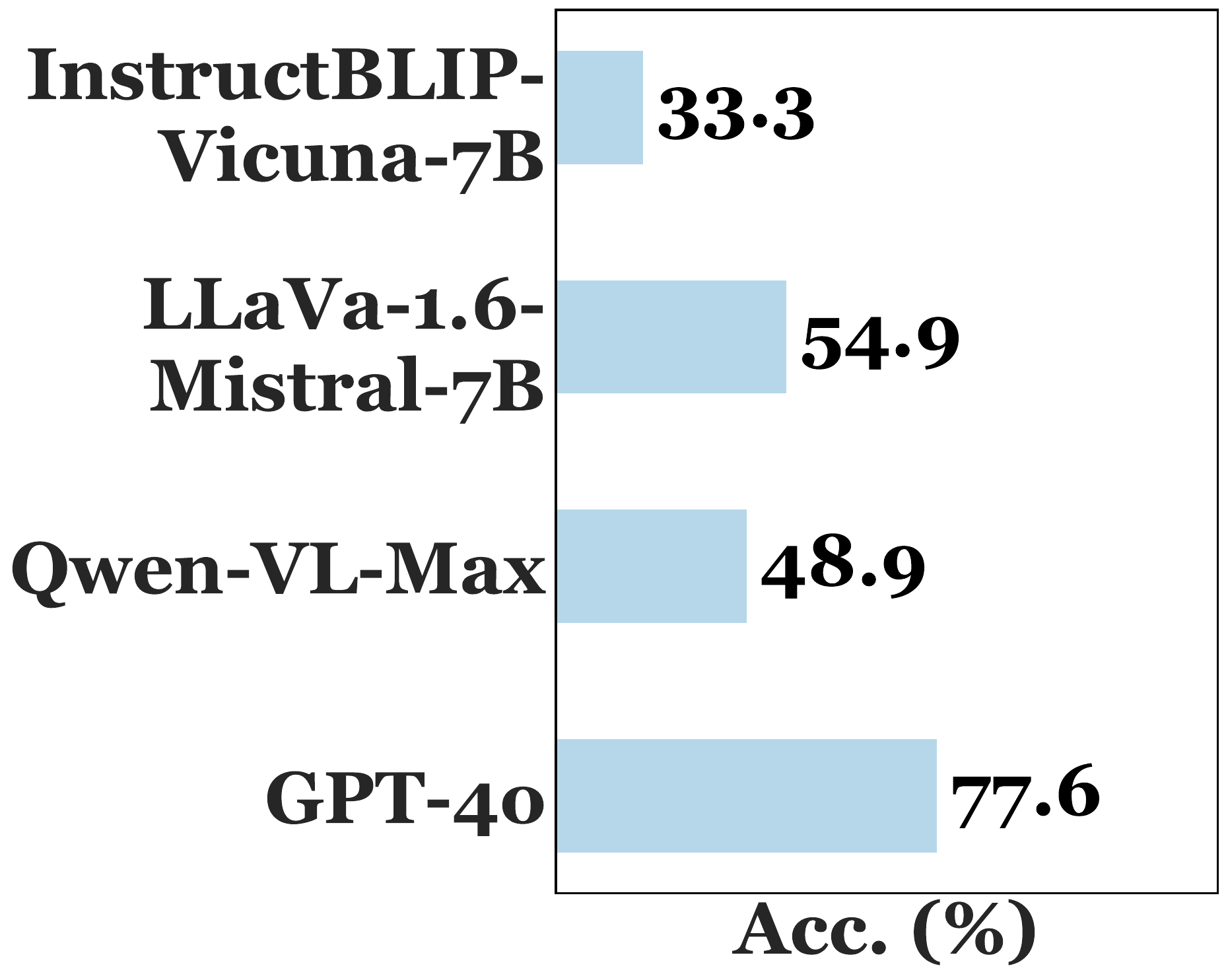}
        \caption{MCoT Answer Correctness}
        \label{fig:correct-mcot}
    \end{subfigure}
    \hfill
    \begin{subfigure}{0.49\columnwidth}
        \centering
        \includegraphics[width=\linewidth]{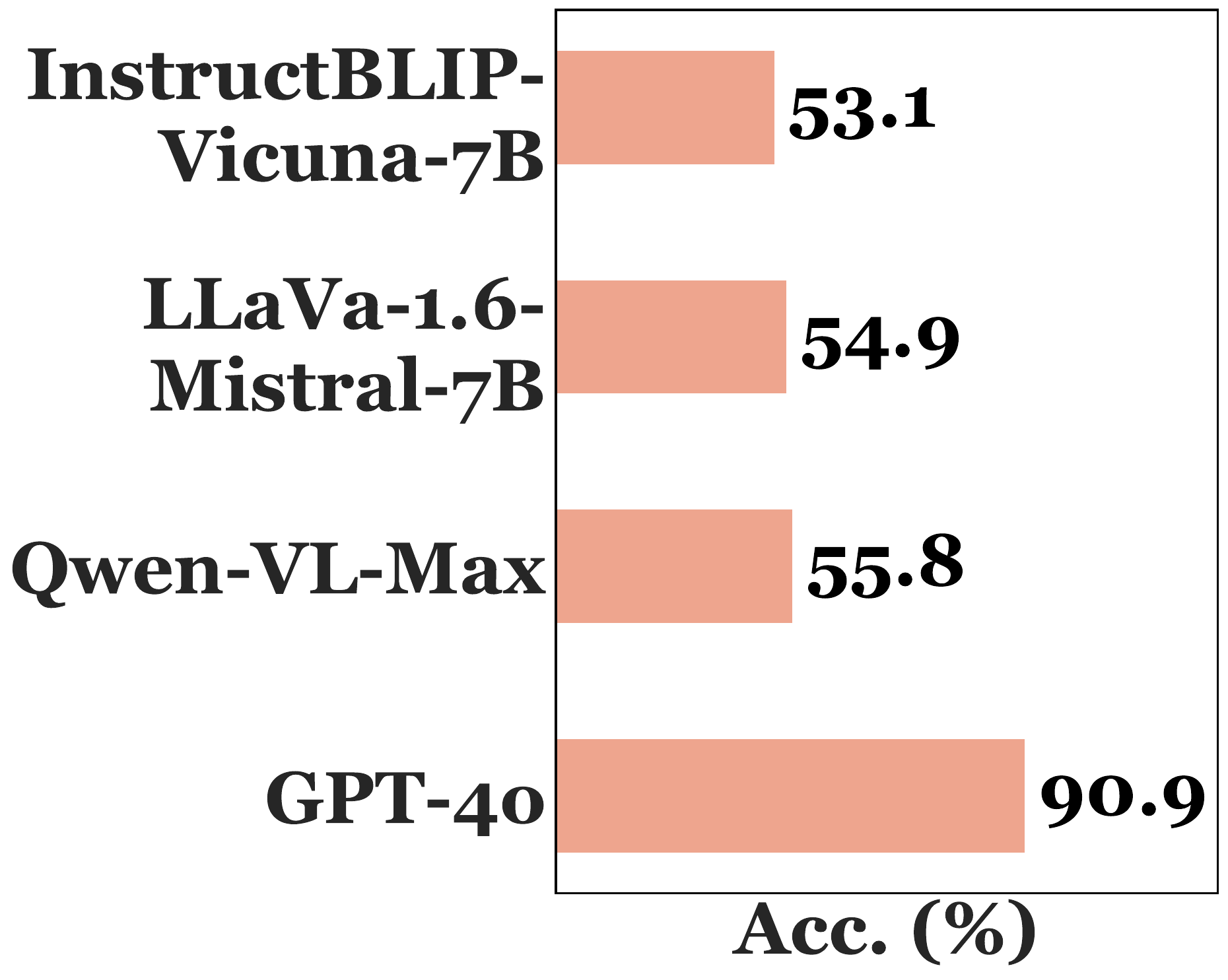}
        \caption{Prediction Correctness}
        \label{fig:mcot-prediction}
    \end{subfigure}   
    \caption{MCoT answer Correctness and Prediction Correctness across different MCoT generators.}
    \label{fig:mcot-acc}
\end{figure}

\noindent \textbf{MLLM generators.} Figure~\ref{fig:mcot-acc} illustrates the accuracy of MCoT answers and Prediction generated by each MLLM. From these statistics, GPT-4o outperform the other three MLLMs in generating the correct MCoT answers.

\section{Annotation}
\label{app2}
% \noindent \textbf{Preliminary Training.} To ensure high-quality data, four of the authors annotated the entire MCoTs (each CoT has three annotations). Previous to the generation of the official annotation of the MCoT answers in the dataset, all annotators underwent preliminary training. We randomly selected 30 questions from the remaining 800 questions not included in the dataset in Sec.~\ref{app1.2} and used GPT-4o and InstructBLIP to generate a total of 60 MCoT answers. The annotators then completed the annotation tasks on these 60 MCoT answers, and based on the annotation results, the annotators engaged in discussions with two additional authors, to reach an internal consensus on some of the more challenging cases.

\subsection{Detailed Definitions on Annotation Tasks}
\label{app2.1}
We provide the labels for each annotation task along with their detailed reference definitions as follows:

\begin{enumerate}
    \item \textbf{Step Type}:
    \begin{itemize}
        \item \textbf{Description}: A step is entailed from the image.
        \item \textbf{Reasoning}: A step is entailed from previous steps.
        \item \textbf{Both}: A step is entailed from both the image and previous steps.
    \end{itemize}

    \item \textbf{Description Correctness}:
    \begin{itemize}
        \item \textbf{Fully correct}: A step without any incorrect information.
        \item \textbf{Partially correct}: A step contains some incorrect information, but there is still some correct information as well.
        \item \textbf{Unsupported}: All information is incorrect.
    \end{itemize}

    \item \textbf{Description Relevance}:
    \begin{itemize}
        \item \textbf{Image relevant}: A step is relevant to the image.
        \item \textbf{Logic relevant}: A step is relevant to answering the question.
        \item \textbf{Both}: A step is relevant to both the image and answering the question.
        \item \textbf{None}: A step is irrelevant to both the image and answering the question.
    \end{itemize}

    \item \textbf{Logic Correctness}:
    \begin{itemize}
        \item \textbf{Correct}: A step can be logically inferred from the previous steps and without any logical errors or conflicts between its internal clauses.
        \item \textbf{Incorrect}: A step cannot be logically inferred from the previous steps or contains logical errors or conflicts between its internal clauses.
    \end{itemize}

    \item \textbf{Logic Relevance}:
    \begin{itemize}
        \item \textbf{Relevant}: A step is relevant to answering the question.
        \item \textbf{Irrelevant}: A step is not relevant to answering the question.
    \end{itemize}

    \item \textbf{Informativeness}:
    \begin{itemize}
        \item \textbf{Uninformative}: A step is repetitive or redundant. 
        \item \textbf{Informative}: A step without repetition or redundancy. 
    \end{itemize}

    \item \textbf{Description Error Types}:
    \begin{itemize}
        \item \textbf{Entity false}: Some entities mentioned in this step do not exist in the image.
        \item \textbf{Attribute false}: The attributes of an entity are incorrectly described, including color, shape, texture, count, state, text recognition.
        \item \textbf{Spatial Relationship false}: The spatial relationship between two objects is incorrectly described.
        \item \textbf{Non-Spatial Relationship false}: The active, passive, or action relationship between two objects is incorrectly described.
    \end{itemize}

    \item \textbf{Logic Error Type}:
    \begin{itemize}
        \item \textbf{Inter-step Incorrect}: A step can not be logically inferred from the previous steps.
        \item \textbf{Intra-step Incorrect}: A step contains logical errors or conflicts between its internal clauses.
        \item \textbf{Both}: A step can not be logically inferred from the previous steps and contains logical errors or conflicts between its internal clauses.
    \end{itemize}

    \item \textbf{MCoT Answer}:
    \begin{itemize}
        \item \textbf{Correct}: The MCoT answer is fully correct, free of any descriptive or logical errors, fully relevant to both the image and the answer, and contains no repetition or redundancy.
        \item \textbf{Incorrect}: The MCoT answer is either irrelevant to the image or the answer, contains descriptive or logically incorrect, or includes repetition or redundancy.
    \end{itemize}
\end{enumerate}

\subsection{Valid Data Filtering Process}
\label{app2.2}
In our valid data filtering process, we filter based on the results of all annotation tasks in Section C.1, defining invalid steps and invalid MCoTs at both the step-level and MCoT-level, respectively.

\vspace{0.2cm}

\noindent\textbf{Invalid Step}
\vspace{-0.2cm}
\begin{itemize}
    \item If the annotation result of a step's step type is 3:0, we further assess the annotation results of other tasks. If there is one task with an annotation result of 1:1:1, this step is deemed an invalid step; otherwise, it is considered a valid step. 
    \vspace{-0.3cm}
    \item If a step's step type is 2:1, meaning each annotation task for step correctness has only two valid annotation results, and if one task has a result of 1:1, then this step is classified as an invalid step.
\end{itemize}

\noindent\textbf{Invalid MCoT}
\vspace{-0.2cm}
\begin{itemize}
    \item If an MCoT answer contains one invalid step, the MCoT is considered invalid data.
    \vspace{-0.3cm}
    \item If half or more of the steps in an MCoT answer have a step type annotation result of 2:1, this MCoT is also deemed invalid data.
\end{itemize}

\begin{figure*}[h]
    \centering
    \includegraphics[width=1\linewidth]{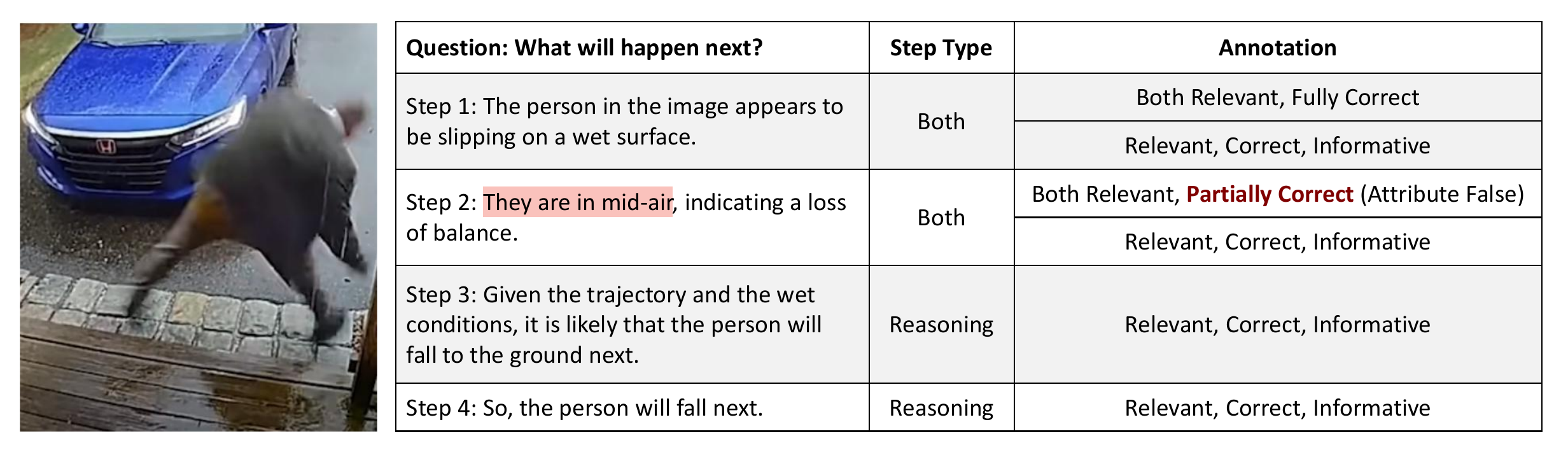}
    \caption{An detailed example of annotation: The Description Correctness of Step 2 is labeled as Partially Correct. However, the subsequent Step 3 and Step 4 do not rely on the incorrect information from Step 2 for their reasoning. Since the logic in these two steps is correct, their Logic Correctness is labeled as Correct.}
    \label{fig:ann-eg}
\end{figure*}

\begin{figure*}[htpb]
    \centering
    \includegraphics[width=1\linewidth]{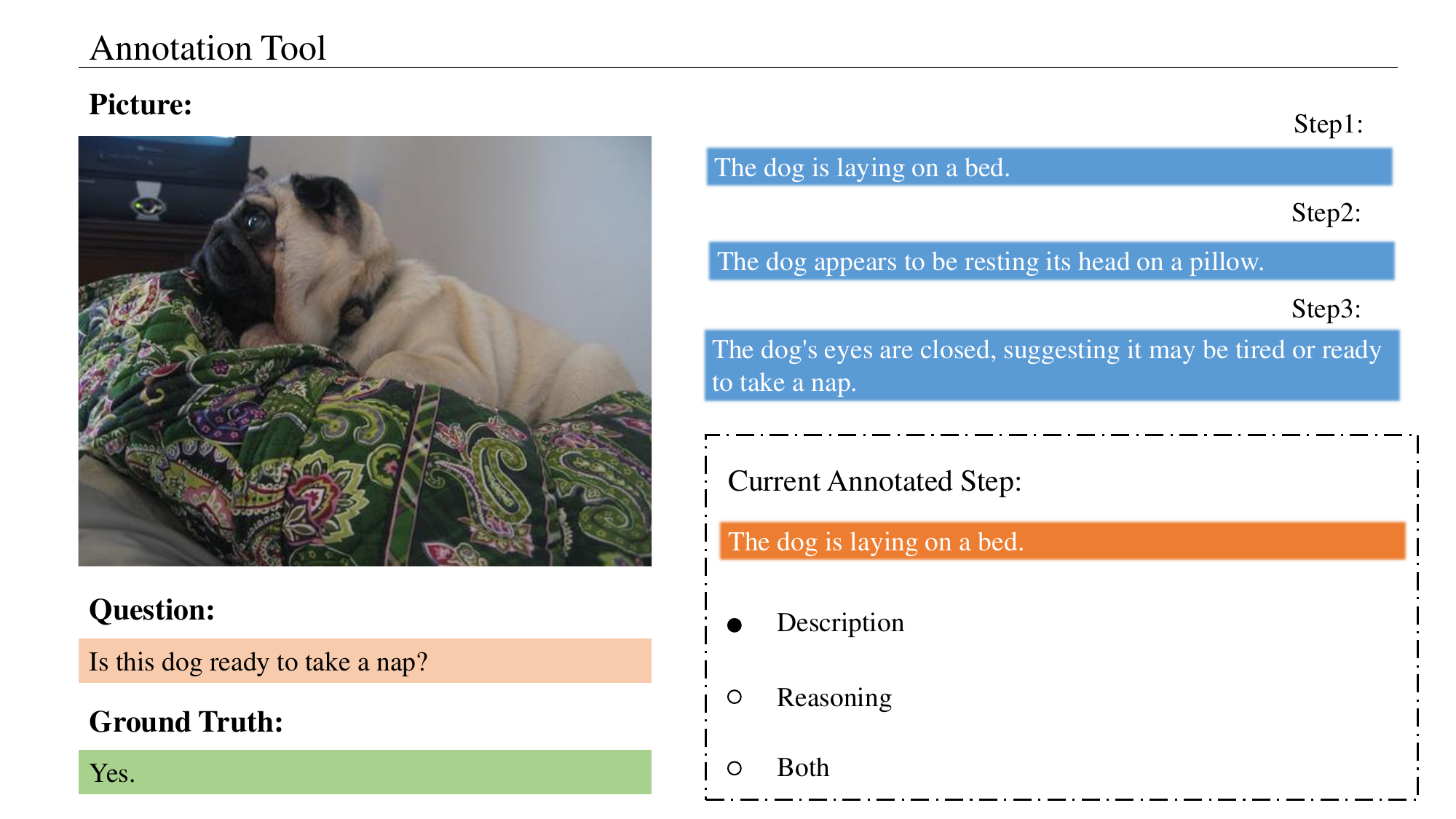}
    \caption{The annotation tool screenshot for step type task.}
    \label{fig:anntool1}
\end{figure*}

\begin{figure*}[htpb]
    \centering
    \includegraphics[width=1\linewidth]{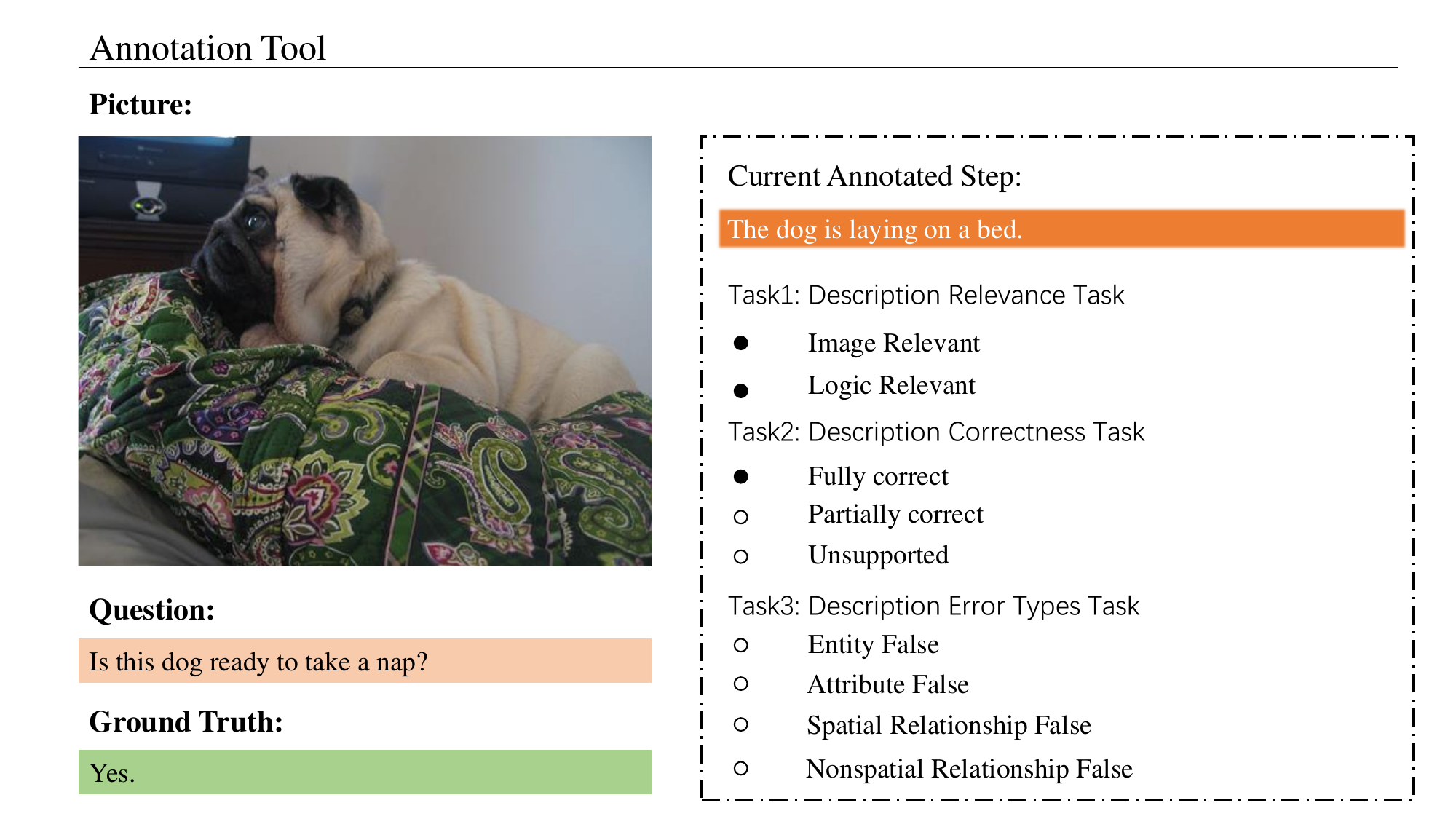}
    \caption{The annotation tasks related to the Description step.}
    \label{fig:anntool2}
\end{figure*}

\begin{figure*}[h]
    \centering
    \includegraphics[width=1\linewidth]{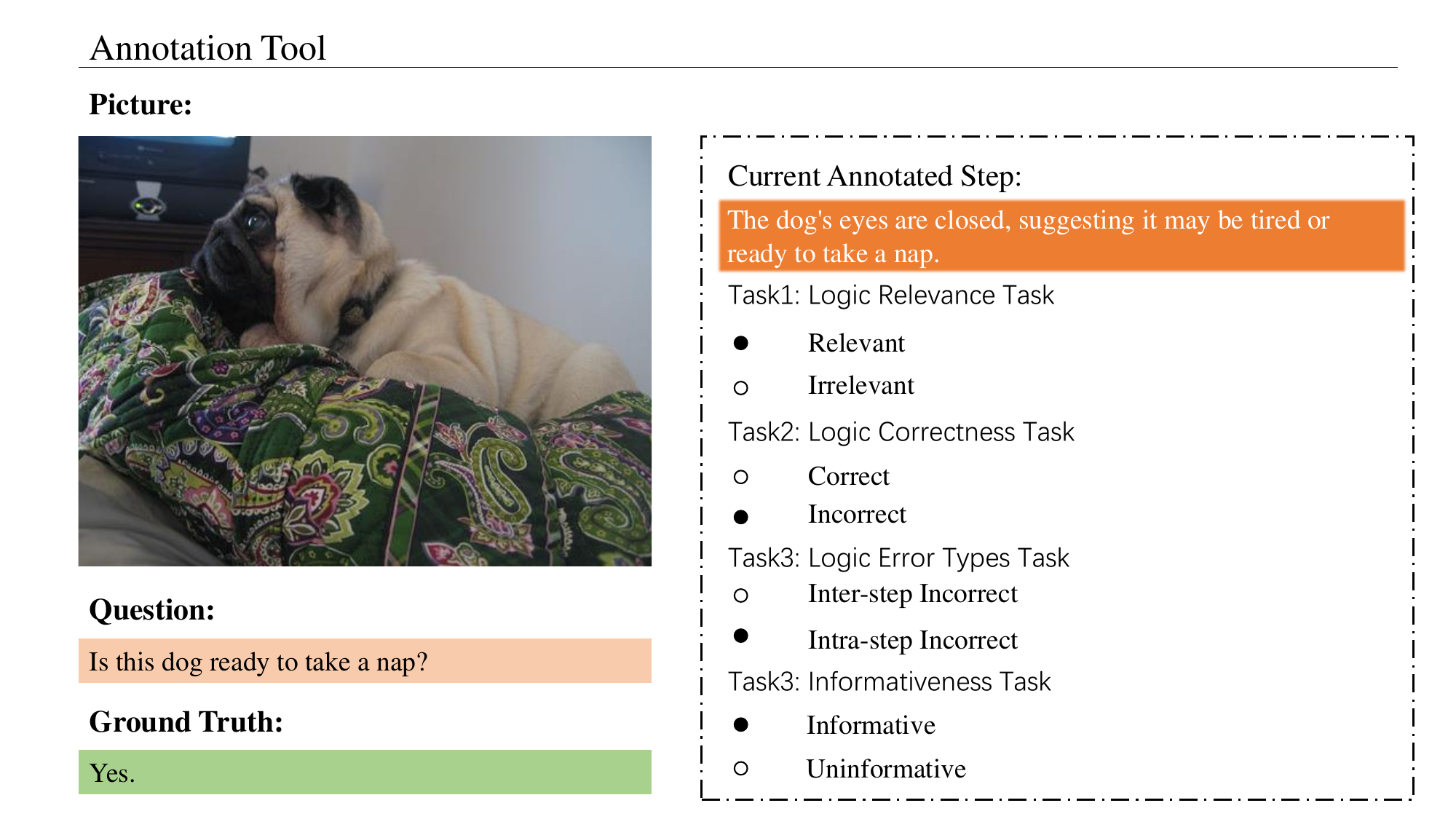}
    \caption{The annotation logic tasks related to the Both step.}
    \label{fig:anntool3}
\end{figure*}

After filtering, our MiCEval dataset contains 903 valid MCoT answers and 2,889 valid MCoT steps. Figure~\ref{fig:ann-eg} provides a detailed label of the MiCEval annotated data and Figures~\ref{fig:anntool1} to \ref{fig:anntool3} show screenshots of our annotation tool.

\subsection{Additional Task: Correctness of Prediction}
\label{app2.3}
During the analysis process, we observed that despite providing a complete reasoning chain in the prompt, some MCoT answers generated by MLLMs were incomplete, with the final prediction missing. Additionally, since the questions are open-ended and do not have a single correct answer, we introduced an extra annotation step: We annotate the final prediction of the MCoT answer (the last step of the reasoning chain) based on whether it correctly answered the question. If the prediction is correct, it will be labeled as \textbf{Correct}; otherwise, it will be labeled as \textbf{Incorrect}. The Bennett, Alpert, and Goldstein's S inter-rater agreement of this additional task is 0.827.

\section{Verifier Experiment}
\label{app3}
\subsection{Detail Experimental Setting and Metric}
\label{app3.1}
The MLLMs evaluated include open-source MLLMs such as LLaVA-1.6-Mistral-7B, LLaVA-1.6-Yi-34B \cite{liu2024llavanext}, MiniCPM-V-2.6 \cite{yao2024minicpm}, and Llama-3.2-11B-Vision-Instruct \cite{dubey2024llama3herdmodels}, as well as API-accessible models like Qwen-VL-Max, GPT-4o, and Gemini-1.5-Pro \cite{geminiteam2024gemini15unlockingmultimodal}. For each task, we conducted three trials and reported the average performance across the runs. In the few-shot evaluation, we designed nine different 4-shot prompts for all tasks, ensuring each set was label-balanced. These demonstrations were randomly drawn from the source datasets.

For the few-shot evaluation, we conducted preliminary experiments using a subset of models, specifically LLaVA-1.6-Mistral-7B, MiniCPM-V-2.6, and GPT-4o, representing both open-source and API-accessible MLLMs. We tested these models with varying shot numbers (1-4) and selected the best-performing configuration for the full few-shot evaluation. Inspired by \citet{chen-etal-2024-m3cot}, we further investigated the performance of these MLLMs across five tasks involving image inputs, examining both multimodal and textual few-shot types.

Given the limitations in MLLMs' instruction-following capabilities, a small percentage of their predictions fell outside the predefined label set. As a result, we use accuracy as the evaluation metric, treating any predictions that do not match the defined labels as incorrect.

\subsection{Prompt Template}
\label{app3.2}
In addition to following the prompt templates used in \citet{jacovi-etal-2024-chain}'s work, we also experimented with various prompt templates to identify one that optimally enhances the performance of each MLLM. The specific prompt templates are as follows:

\noindent\textbf{Step Type Task}\\
\hspace*{2em}Image: [image]\\
\hspace*{2em}Question: [question]\\
\hspace*{2em}Rationale: [rationale] \\ 
\hspace*{2em}Step: [current step] \\ 
\hspace*{2em}Step Type: \{Description, Reasoning, Both\}\\

\noindent\textbf{Description Correctness Task}\\
\hspace*{2em}Image: [image]\\
\hspace*{2em}Step: [description step]\\
\hspace*{2em}Output: \{Fully Correct, Partially Correct, Unsupported\}\\

\noindent\textbf{Description Relevance Task}\\
\hspace*{2em}Image: [image]\\
\hspace*{2em}Question: [question]\\
\hspace*{2em}Rationale: [rationale] \\ 
\hspace*{2em}Step: [description step] \\ 
\hspace*{2em}Output: \{Both, Image Relevant, Logic Relevant, None\}\\

\noindent\textbf{Description Error Types Task}\\
\hspace*{2em}Image: [image]\\
\hspace*{2em}Step: [incorrect description step] \\ 
\hspace*{2em}Output: \{Entity False, Attribute False, Spatial Relationship False, Non-spatial Relationship False\}\\

\noindent\textbf{Logic Correctness}\\
\hspace*{2em}Question: [question]\\
\hspace*{2em}Premise: [previous steps] \\ 
\hspace*{2em}Hypothesis: [reasoning step] \\ 
\hspace*{2em}Output: \{Correct, Incorrect\}\\

\noindent\textbf{Logic Relevance Task}\\
\hspace*{2em}Question: [question]\\
\hspace*{2em}Rationale: [rationale] \\ 
\hspace*{2em}Step: [reasoning step] \\ 
\hspace*{2em}Output: \{Relevant, Irrelevant\}\\

\noindent\textbf{Logic Error Types Task}\\
\hspace*{2em}Premise: [previous steps]\\
\hspace*{2em}Hypothesis: [incorrect reasoning step] \\ 
\hspace*{2em}Output: \{Inter-step Incorrect, Intra-step Incorrect, Both\}\\

\noindent\textbf{Informativeness Task}\\
\hspace*{2em}Previous: [previous steps]\\
\hspace*{2em}Step: [incorrect reasoning step] \\ 
\hspace*{2em}Output: \{Informative, Uninformative\}\\

\noindent\textbf{MCoT Correctness Task}\\
\hspace*{2em}Image: [image]\\
\hspace*{2em}Question: [question]\\
\hspace*{2em}Rationale: [rationale]\\
\hspace*{2em}Is this a good rationale or not? Output: \{Yes, No\}

\subsection{Detailed Results}
\label{app3.3}

\begin{table*}[h]
    \centering
    \resizebox{\textwidth}{!}{
    \begin{tabular}{c|l|c|ccc|cccc|c|c}
    \toprule
      \multirow{2}{*}{\textbf{Evaluation Setting}} &\multicolumn{1}{c|}{\multirow{2}{*}{\textbf{Model}}} & \multirow{2}{*}{\textbf{Step Type}} &\multicolumn{3}{c|}{\textbf{Description}} & \multicolumn{4}{c|}{\textbf{Logic}} & \multirow{2}{*}{\textbf{MCoT}} & \multirow{2}{*}{\textbf{Avg.}} \\ 
      
       ~ & ~ & ~ & {\textbf{Relevance}} & {\textbf{Correctness}} & {\textbf{Error Types}} & {\textbf{Relevance}} & {\textbf{Correctness}} & {\textbf{Informativeness}} & {\textbf{Error Types}} ~ &\\
        
        \midrule
\multirow{7}{*}{\centering{Zero-Shot}} & {LLaVA-1.6-Mistral-7B} & 0.719 & 0.484 & 0.144 & 0.260 & 0.866 & 0.541 & \textbf{0.860} & 0.729 & 0.574 & 0.575 \\
~ & {MiniCPM-V-2.6} & 0.184 & 0.731 & 0.782 & 0.480 & 0.950 & 0.774 & 0.789 & 0.931 & 0.756 & 0.709 \\
~ & {Llama-3.2-11B-Vision-Instruct}& 0.380 & 0.757 & 0.137 & 0.453 & 0.842 & 0.580 & 0.345 & 0.861 & 0.632 & 0.554\\
~ & {LLaVA-1.6-Yi-34B}  & 0.580 & 0.512 & \textbf{0.804} & 0.560 & \textbf{0.962} & 0.808 & 0.849 & 0.812 & 0.729 & 0.735 \\
~ & {GPT-4o} & 0.767 & 0.763 & 0.748 & \textbf{0.650} & 0.938 & 0.753 & 0.568 & 0.812 & \textbf{0.765} & \textbf{0.752} \\
~ & {Qwen-VL-Max} & 0.420 & 0.910 & 0.714 & 0.580 & 0.952 & \textbf{0.815} & 0.805 & 0.271 & 0.740 & 0.690 \\
~ & {Gemini-1.5-Pro}& \textbf{0.838} & \textbf{0.921} & 0.778 & 0.490 & 0.935 & 0.671 & 0.291 & \textbf{0.979} & 0.646 & 0.728 \\
        \midrule
\multirow{7}{*}{\centering{Few-Shot}} & {LLaVA-1.6-Mistral-7B}  & 0.705 & 0.313 & 0.374 & 0.423 & 0.914 & 0.801 & 0.712 & \textbf{0.979} & 0.686 & 0.656 \\
~ & {MiniCPM-V-2.6}  & 0.277 & 0.428 & \textbf{0.795} & 0.607 & 0.879 & 0.807 & 0.717 & 0.930 & 0.724 & 0.685 \\
~ & {Llama-3.2-11B-Vision-Instruct} & 0.005 & 0.223 & 0.716 & 0.010 & 0.805 & 0.760 & 0.130 & 0.271 & 0.251 & 0.352 \\
~ & {LLaVA-1.6-Yi-34B}  & 0.744 & 0.164 & 0.752 & 0.590 & \textbf{0.955} & 0.805 & \textbf{0.866} & 0.938 & 0.735 & 0.728 \\
~ & {GPT-4o}& \textbf{0.856} & \textbf{0.530} & 0.739 & \textbf{0.730} & 0.908 & 0.726 & 0.623 & 0.773 & \textbf{0.785} & \textbf{0.741} \\
~ & {Qwen-VL-Max} & 0.679 & 0.437 & 0.725 & 0.530 & 0.921 & \textbf{0.808} & 0.654 & 0.521 & 0.704 & 0.664 \\
~ & {Gemini-1.5-Pro}& - & - & - & - & - & - & - & - & - & - \\

        \bottomrule
    \end{tabular}
    }
        \setlength{\belowcaptionskip}{0.2cm}
    \caption{The overall performance on Pairwise Comparison of different MLLMs in MiCEval-Normal.}
    \label{tab:pair-normal}
\end{table*}

\begin{figure*}[h]
    \centering
    \begin{subfigure}[b]{\linewidth}
        \centering
        \includegraphics[width=\linewidth]{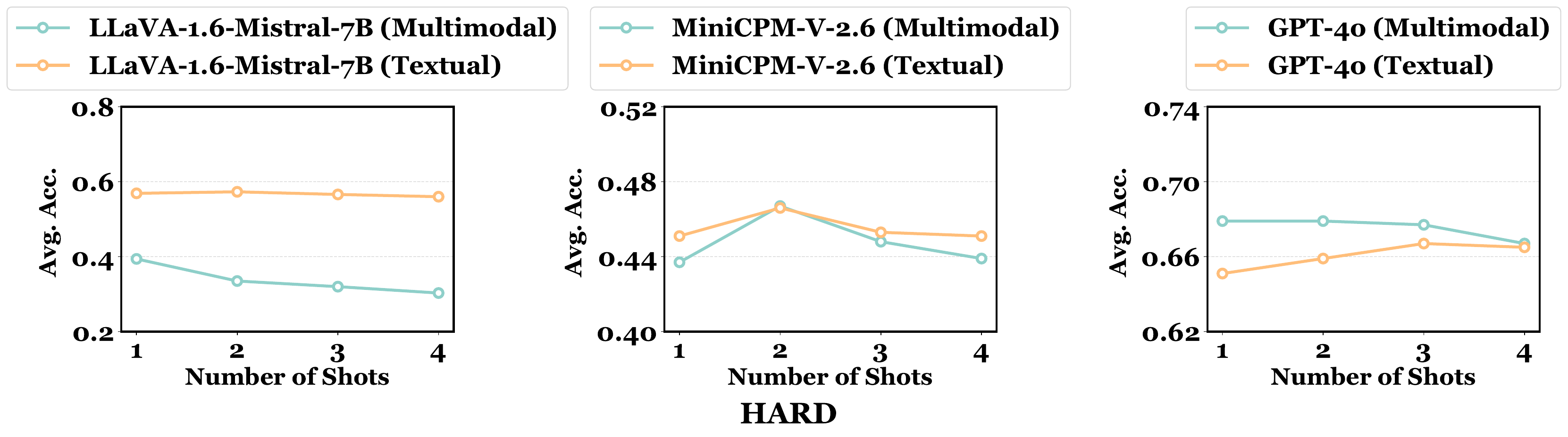}
    \end{subfigure}

    \begin{subfigure}[b]{\linewidth}
        \centering
        \includegraphics[width=\linewidth]{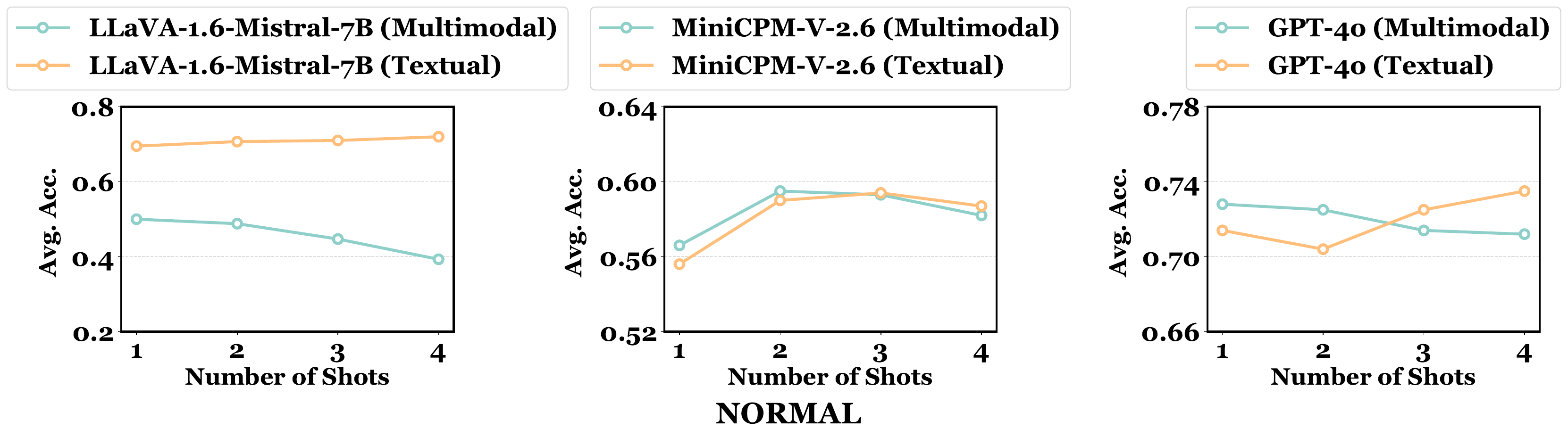}
    \end{subfigure}
    \caption{The relationship between the average accuracy of three MLLMs across five image-input Pairwise Comparison tasks and the number of shots. \textbf{HARD}: The performance on MiCEval-\textsc{Hard}; \textbf{NORMAL}: The performance on MiCEval-\textsc{Normal}.}
    \label{fig:textaul}
\end{figure*}

Table~\ref{tab:pair-normal} presents the results in MiCEval-\textsc{Normal}. The results of the exploration experiments for few-shot types are illustrated in Figure~\ref{fig:textaul}. The F1 metrics for each class on some important tasks under two different evaluation settings, as shown in Table~\ref{tab:f1-hard} and Table~\ref{tab:f1-normal}.

\begin{table*}[h]
  \begin{center}
    \resizebox{\textwidth}{!}{
    \begin{tabular}{l|ccc|cc|cc|cc|cc|cc}
    \toprule
      \multicolumn{1}{c|}{\multirow{2}{*}{Model}} &\multicolumn{3}{c|}{Step Type} & \multicolumn{2}{c|}{Description Relevance} & \multicolumn{2}{c|}{Description Correct}
      & \multicolumn{2}{c|}{Logic Relevance}& \multicolumn{2}{c|}{Logic Correct}& \multicolumn{2}{c}{Informativeness}\\ 
      
      ~ & Description & Reasoning & Both & {Both} & {Others} & Correct & Incorrect & Relevant & Irrelevant & Correct & Incorrect & {Infor...} & {Uninfo...}\\
       \midrule
        \multicolumn{13}{c}{Zero-shot} \\
        \midrule
        LLaVA-1.6-Mistral-7B & 0.727 & 0.661 & 0.028 & 0.649 & 0.010 & 0.211 & 0.170 & 0.946 & 0.354 & 0.462 & 0.510 & 0.959 & 0.000 \\
        MiniCPM-V-2.6 & 0.075 & 0.200 & 0.077 & 0.826 & 0.110 & 0.839 & 0.340 & 0.959 & 0.289 & 0.775 & 0.353 & 0.912 & 0.028 \\
        LLaVA-1.6-Yi-34B & 0.641 & 0.181 & 0.086 & 0.716 & 0.110 & 0.884 & 0.180 & 0.974 & 0.358 & 0.791 & 0.417 & 0.956 & 0.167 \\
        GPT-4o & 0.849 & 0.574 & 0.048 & 0.879 & 0.100 & 0.851 & 0.140 & 0.947 & 0.388 & 0.741 & 0.455 & 0.834 & 0.297 \\
        Qwen-VL-Max & 0.589 & 0.540 & 0.075 & 0.938 & 0.050 & 0.814 & 0.240 & 0.964 & 0.396 & 0.809 & 0.226 & 0.952 & 0.255 \\
        Gemini-1.5-Pro & 0.895 & 0.914 & 0.165 & 0.912 & 0.240 & 0.860 & 0.220 & 0.960 & 0.427 & 0.665 & 0.451 & 0.442 & 0.160 \\
        Llama-3.2-11B-Vision-Instruct & 0.163 & 0.216 & 0.095 & 0.899 & 0.020 & 0.347 & 0.040 & 0.918 & 0.213 & 0.567 & 0.504 & 0.495 & 0.174 \\
        \midrule
        \multicolumn{13}{c}{Few-shot} \\
        \midrule
        LLaVA-1.6-Mistral-7B & 0.349 & 0.625 & 0.035 & 0.483 & 0.060 & 0.435 & 0.090 & 0.965 & 0.364 & 0.790 & 0.389 & 0.894 & 0.168 \\
        MiniCPM-V-2.6 & 0.133 & 0.124 & 0.075 & 0.592 & 0.130 & 0.848 & 0.370 & 0.926 & 0.300 & 0.790 & 0.297 & 0.865 & 0.162 \\
        LLaVA-1.6-Yi-34B & 0.792 & 0.769 & 0.128 & 0.297 & 0.080 & 0.851 & 0.100 & 0.968 & 0.422 & 0.788 & 0.260 & 0.962 & 0.031 \\
        GPT-4o & 0.908 & 0.897 & 0.188 & 0.625 & 0.080 & 0.843 & 0.190 & 0.940 & 0.410 & 0.757 & 0.493 & 0.871 & 0.325 \\
        Qwen-VL-Max & 0.571 & 0.257 & 0.064 & 0.403 & 0.050 & 0.843 & 0.060 & 0.959 & 0.429 & 0.869 & 0.224 & 0.848 & 0.293 \\
        Llama-3.2-11B-Vision-Instruct & 0.000 & 0.000 & 0.085 & 0.420 & 0.010 & 0.772 & 0.070 & 0.769 & 0.167 & 0.757 & 0.486 & 0.000 & 0.128 \\
      \bottomrule
    \end{tabular}    }
    \caption{Per-label F1 metrics of each task under zero-shot and few-shot evaluations on MiCEval-\textsc{Hard}.}
    \label{tab:f1-hard}
  \end{center}
\end{table*}

\begin{table*}[h]
\scriptsize
  \begin{center}
    \resizebox{\linewidth}{!}{
    \begin{tabular}{l|ccc|cc|cc|cc|cc|cc}
    \toprule
      \multicolumn{1}{c|}{\multirow{2}{*}{Model}} &\multicolumn{3}{c|}{Step Type} & \multicolumn{2}{c|}{Description Relevance} & \multicolumn{2}{c|}{Description Correct}
      & \multicolumn{2}{c|}{Logic Relevance}& \multicolumn{2}{c|}{Logic Correct}& \multicolumn{2}{c}{Informativeness}\\ 
      
      ~ & Description & Reasoning & Both & {Both} & {Others} & Correct & Incorrect & Relevant & Irrelevant & Correct & Incorrect & {Infor...} & {Uninfo...}\\
       \midrule
       \multicolumn{14}{c}{Zero-Shot} \\
       \midrule
        LLaVA-1.6-Mistral-7B & 0.824 & 0.367 & 0.000 & 0.660 & 0.010 & 0.200 & 0.080 & 0.928 & 0.133 & 0.656 & 0.309 & 0.924 & 0.047 \\
        MiniCPM-V-2.6 & 0.371 & 0.053 & 0.007 & 0.854 & 0.050 & 0.890 & 0.260 & 0.974 & 0.000 & 0.870 & 0.000 & 0.879 & 0.114 \\
        LLaVA-1.6-Yi-34B & 0.774 & 0.111 & 0.025 & 0.676 & 0.110 & 0.922 & 0.100 & 0.981 & 0.154 & 0.893 & 0.097 & 0.913 & 0.450 \\
        GPT-4o & 0.919 & 0.489 & 0.040 & 0.868 & 0.070 & 0.888 & 0.140 & 0.968 & 0.400 & 0.854 & 0.200 & 0.674 & 0.364 \\
        Qwen-VL-Max & 0.637 & 0.242 & 0.000 & 0.960 & 0.000 & 0.861 & 0.160 & 0.975 & 0.133 & 0.898 & 0.036 & 0.878 & 0.513 \\
        Gemini-1.5-Pro & 0.915 & 0.689 & 0.051 & 0.963 & 0.150 & 0.900 & 0.170 & 0.966 & 0.250 & 0.791 & 0.226 & 0.312 & 0.269 \\
        Llama-3.2-11B-Vision-Instruct & 0.605 & 0.020 & 0.011 & 0.856 & 0.000 & 0.244 & 0.030 & 0.914 & 0.000 & 0.720 & 0.176 & 0.411 & 0.281 \\
       \midrule
       \multicolumn{14}{c}{Few-Shot} \\
       \midrule
        LLaVA-1.6-Mistral-7B & 0.821 & 0.332 & 0.000 & 0.476 & 0.020 & 0.560 & 0.070 & 0.955 & 0.194 & 0.888 & 0.094 & 0.818 & 0.311 \\
        MiniCPM-V-2.6 & 0.499 & 0.065 & 0.011 & 0.582 & 0.020 & 0.892 & 0.290 & 0.938 & 0.138 & 0.888 & 0.094 & 0.834 & 0.147 \\
        LLaVA-1.6-Yi-34B & 0.858 & 0.529 & 0.043 & 0.268 & 0.140 & 0.893 & 0.110 & 0.977 & 0.235 & 0.891 & 0.066 & 0.927 & 0.170 \\
        GPT-4o & 0.926 & 0.718 & 0.036 & 0.697 & 0.030 & 0.875 & 0.180 & 0.950 & 0.372 & 0.833 & 0.231 & 0.726 & 0.396 \\
        Qwen-VL-Max & 0.813 & 0.216 & 0.032 & 0.610 & 0.010 & 0.876 & 0.060 & 0.959 & 0.207 & 0.893 & 0.067 & 0.754 & 0.416 \\
        Llama-3.2-11B-Vision-Instruct & 0.005 & 0.000 & 0.007 & 0.368 & 0.000 & 0.841 & 0.060 & 0.890 & 0.123 & 0.859 & 0.186 & 0.000 & 0.230 \\
      \bottomrule
    \end{tabular}
    }
    \caption{Per-label F1 metrics of each task under zero-shot and few-shot evaluations on MiCEval-\textsc{Normal}.}
    \label{tab:f1-normal}
  \end{center}
\end{table*}

\noindent \textit{$\triangleright$ The reasoning capabilities of the MCoT verifier still need improvement.} The MCoT responses in MiCEval-\textsc{Hard} involve more reasoning steps and are generally more complex compared to those in MiCEval-\textsc{Normal}. MLLM performance on MiCEval-\textsc{Normal} consistently exceeds that on MiCEval-\textsc{Hard}, suggesting that current MLLMs align more closely with human judgments on simpler reasoning tasks. This highlights their limitations in effectively validating MCoTs that require complex reasoning.

\noindent \textit{$\triangleright$ Good performance in textual few-shot evaluation can be achieved.} GPT-4o and MiniCPM showed similar performance when handling multimodal demonstrations and textual demonstrations. In contrast, LLaVA-1.6-Mistral-7B exhibited nearly 20\% improvement of accuracy in the textual few-shot evaluation compared to the multimodal few-shot evaluation.

\subsection{Additional Analysis}
\label{app3.4}
\noindent \textbf{Llama under Few-shot Evaluation.} We observed that Llama-3.2-11B-Vision-Instruct performed significantly worse in few-shot evaluation compared to zero-shot evaluation. Therefore, we conducted a statistical analysis of the outputs from Llama-3.2 on several tasks where its performance was particularly poor in the few-shot evaluation. We found that the proportion of outputs labeled as “Both” for the step type was 99.9\%. Additionally, in terms of informativeness, all outputs of Llama-3.2 were “Uninformative”. In the Description Error Types task, 74.0\% of its outputs were invalid, meaning they fell outside the defined label range. Overall, Llama-3.2 exhibited a decline in performance across every task in the few-shot evaluation, which may be related to its training process, suggesting that Llama-3.2 might lack multi-round and multi-image training.

\noindent \textbf{Description Relevance in Pairwise Comparison.} 
We observed that on both MiCEval-\textsc{Hard} and MiCEval-\textsc{Normal}, all MLLMs performed worse in description relevance during few-shot evaluation compared to zero-shot evaluation. We analyzed each model’s accuracy across the labels in the description relevance task, with results shown in Tables~\ref{tab:f1-hard} and~\ref{tab:f1-normal}. Our analysis revealed that the prediction distribution of each model shifted in different ways during few-shot evaluation. We believe this performance drop may be due to the larger number of label classes in description relevance compared to other tasks, as well as the generally weaker in-context learning abilities of MLLMs for this task.

% For a detailed breakdown of each MLLM’s performance on different labels across tasks, please refer to the App.~\ref{app3.3}.

\section{Evaluator Experiment}
\label{app4}
\subsection{Detail Experimental Setting and Metrics}
\label{app4.1}
We selected image-text matching metrics such as CLIP \cite{hessel-etal-2021-clipscore}, BLIP2-ITM, BLIP2-ITC \cite{li2023blip2}, and LLM-Score \cite{chen-etal-2024-unveiling}, along with the reasoning chain evaluation metric ReCEval \cite{prasad-etal-2023-receval}, as baselines. For each metric, we compute the scores corresponding to each step in the MCoT answer.

For the \ours evaluators, we selected LLaVA-1.6-Mistral-7B, MiniCPM-V-2.6, Llama-3.2-11B-Vision-Instruct, and GPT-4o as foundational MLLMs. These MLLMs assess all steps in the MCoT answers based on both the \ours description and reasoning step correctness. {For  $\textit{Correctness}_{type}$, we use the results of Gemini-1.5-Pro on the step type task to determine the type label for each step (best zero-shot step type performance on Sec.~\ref{sec6.1.2})}. We then calculate  {two kinds of} overall MCoT correctness (Step, \ours, Type or All). Finally, we compare the correlation between the MCoT correctness and human scores. Additionally, we evaluate the correctness scores generated by the MLLMs for each step (Step, w/o \ours) and for the overall MCoT (CoT, w/o \ours) without utilizing the \ours metrics. This is done to validate the effectiveness of \ours while minimizing the influence of the ``Step-by-Step'' evaluation method.

\noindent \textbf{Scoring Evaluation.} We use Somer’s D \cite{Somers1962ANA} as the metric for human correlation. 

\noindent \textbf{Choice Ranking.} Each question in the dataset contains at least one fully correct MCoT option and one that is not. We calculated correctness scores for each MCoT answer and selected the option with the highest score. Accuracy was then computed based on whether the selected choice was fully correct. The MCoT evaluation metrics that achieved the highest accuracy demonstrate a stronger ability to filter high-quality MCoTs. we use accuracy as the evaluation metric. An example of Choice Ranking data as shown as Table~\ref{tab:sample-cr}.

\begin{table*}[h]
\small
\begin{center}
\begin{tabularx}{\textwidth}{
>{\centering\arraybackslash}m{0.2\textwidth}
>{\centering\arraybackslash}m{0.13\textwidth}
m{0.3\textwidth}
>{\centering\arraybackslash}m{0.06\textwidth}
>{\centering\arraybackslash}m{0.18\textwidth}
>{\centering\arraybackslash}m{0.08\textwidth}
}
\toprule
\textbf{Image}&\textbf{Question} &\textbf{MCoT Answer}& \textbf{Generator} &\textbf{Ground Truth}\\
\midrule
% \rowcolor{brown!5}
\multirow{17}{=}{\includegraphics[width=2.8cm,height=2cm]{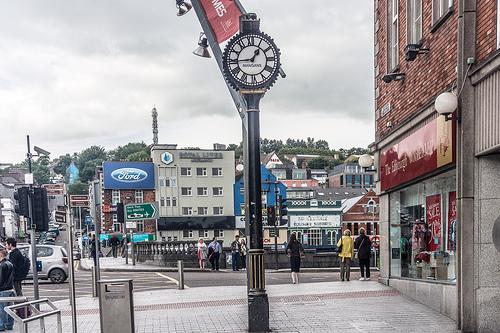}} &
\multirow{17}{=}{\centering{Why is the clock there?}}  & \cellcolor{brown!5}{(1) The clock is positioned prominently in a busy street area. It is elevated on a pole, making it visible to pedestrians and drivers. The area appears to be a commercial district with shops and businesses. So the clock is there to help people keep track of time while they are out and about.} & \cellcolor{brown!5}{GPT-4o}  & \cellcolor{brown!5}{Correct}\\
\\
~ & ~ & {(2) This is an outdoor scene during the day. A large black clock tower stands prominently against the sky. It has 2 clocks facing different directions. 1 can assume that its purpose might be to inform or remind passersby about the time.} & {Qwen-VL-Max}& {Incorrect}\\
\\
% \rowcolor{brown!5}
~ & ~ &  \cellcolor{brown!5}{(3) The clock is on a pole in the middle of a city street. People are walking by the clock and looking at it. The clock is showing the time. Therefore, the clock is there for people to check the time.} & \cellcolor{brown!5}{Instructblip-Vicuna-7B} &    \cellcolor{brown!5}{Correct}\\
\\
~ & ~ & {(4) The clock is located on a pole on the sidewalk. It is a public clock, likely placed there for pedestrians to easily see the time. The clock is situated in a busy city street with many people walking by. The clock serves as a convenient time reference for pedestrians in the area.}  & {LLaVA-1.6-Mistral-7B}  & {Correct}\\

\bottomrule
\end{tabularx}
\caption{An Example of Choice Ranking Data.}
\label{tab:sample-cr}
\end{center}
\end{table*}

\subsection{Prompt Template}
\label{app4.2}
\begin{figure*}[h]
    \centering
    \includegraphics[width=1\linewidth]{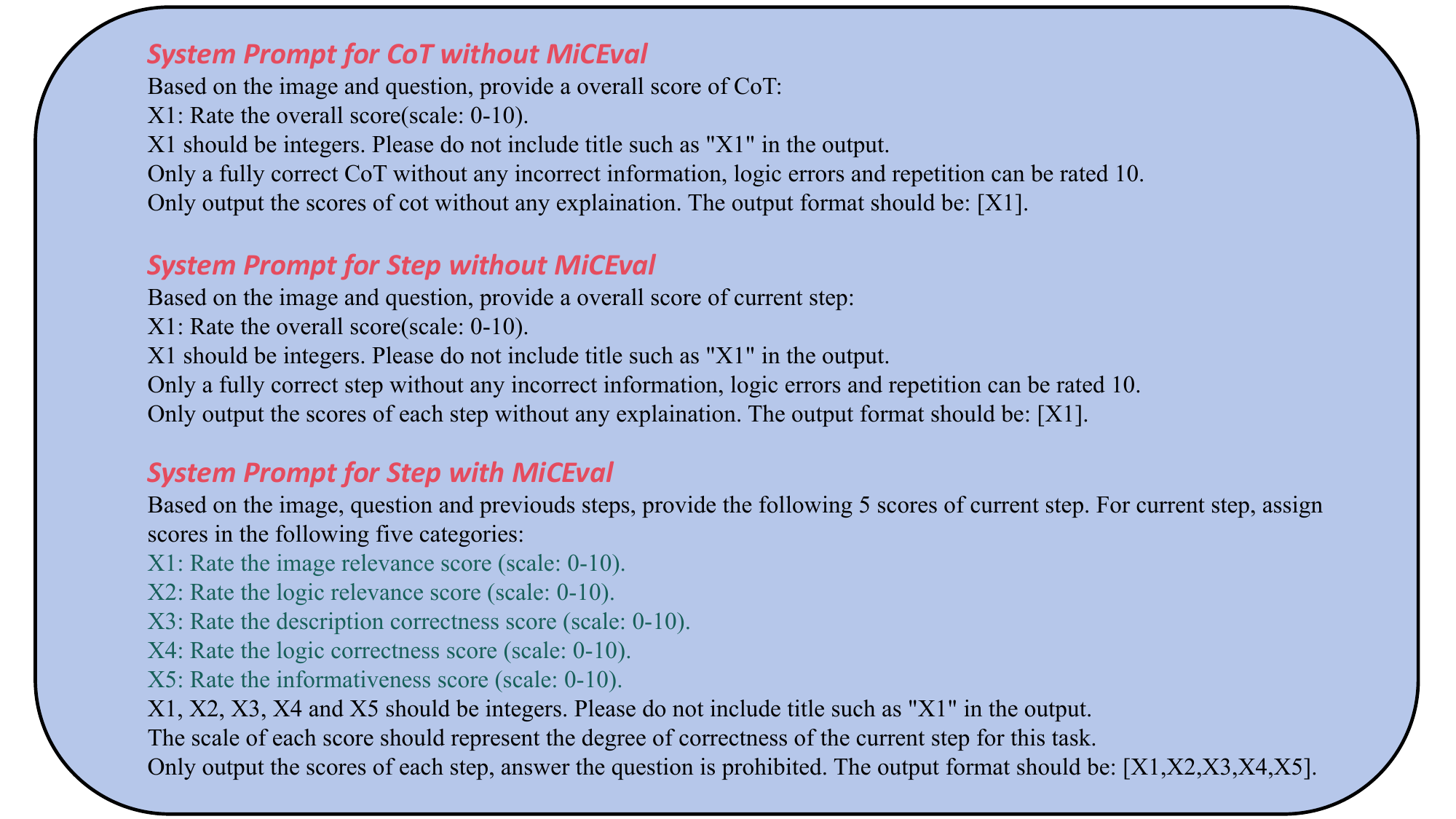}
    \caption{Different system prompts of Scoring Evaluation. \textbf{System Prompt for Step with MiCEval} is the system prompt for MLLM-based MiCEval evaluation metrics.}
    \label{fig:prompt}
\end{figure*}

In this section, the system prompt we used in each setting are illustrated in the Figure~\ref{fig:prompt}, and we provide prompt templates used in the experiments as following:

\noindent\textbf{MCoT Task}\\
\hspace*{2em}Image: [image]\\
\hspace*{2em}Question: [question]\\
\hspace*{2em}CoT: [MCoT answer]\\
\hspace*{2em}Output: [Score of MCoT]\\

\noindent\textbf{Step Task}\\
\hspace*{2em}Image: [image]\\
\hspace*{2em}Question: [question]\\
\hspace*{2em}Previous: [previous steps]\\
\hspace*{2em}Current Step: [current step]\\
\hspace*{2em}Output: [Score of the Current Step]\\

\subsection{Additional Results}
\label{app4.3} 

\noindent \textit{$\triangleright$ Step-by-step evaluation does not always help MLLMs.} Step-by-step evaluation benefits MiniCPM-V-2.6 and LLaVA-1.6-Mistral but harms Llama-3.2-11B-Vision-Instruct’s performance. We observe that, compared to directly evaluating the entire MCoT without MiCEval, step-by-step evaluation without MiCEval improves the Somer’s D score of MiniCPM-V-2.6 from 0.130 to 0.169 and LLaVA-1.6-Mistral-7B from 0.080 to 0.140 on the whole MiCEval. In contrast, Llama’s performance drops from 0.176 to 0.075. Additionally, these scores remain lower than those achieved by the MiCEval-based step-by-step evaluation method, indicating that step-by-step evaluation alone is not the primary factor driving improvements in MiCEval MLLM-based methods.

\subsection{Analysis based on Human Annotation}
\label{app4.4}
We use the step type results from human annotation to evaluate the correlation between the $\textit{Correctness}_{type}$-based MiCEval metrics and human judgments, focusing on the Description step, Reasoning step, and the overall MiCEval dataset. We also present the Spearman's Rank Correlation Coefficient \cite{ca468a70-0be4-389a-b0b9-5dd1ff52b33f} results.

The Table~\ref{tab:gt-step-somersd} and Table~\ref{tab:gt-mcot-somersd} show the results on human annotated step type labels. The Spearman's Rank Correlation Coefficient \cite{ca468a70-0be4-389a-b0b9-5dd1ff52b33f} results for different datasets, as illustrated in Table~\ref{tab:gt-spearman-step} and Table~\ref{tab:gt-spearman-dataset}. The Choice Ranking results based on human annotation is shown in Table~\ref{tab:gt-choice-ranking}.

\section{Case Study}
\begin{figure*}[htpb]
    \centering
    \includegraphics[width=0.9\linewidth]{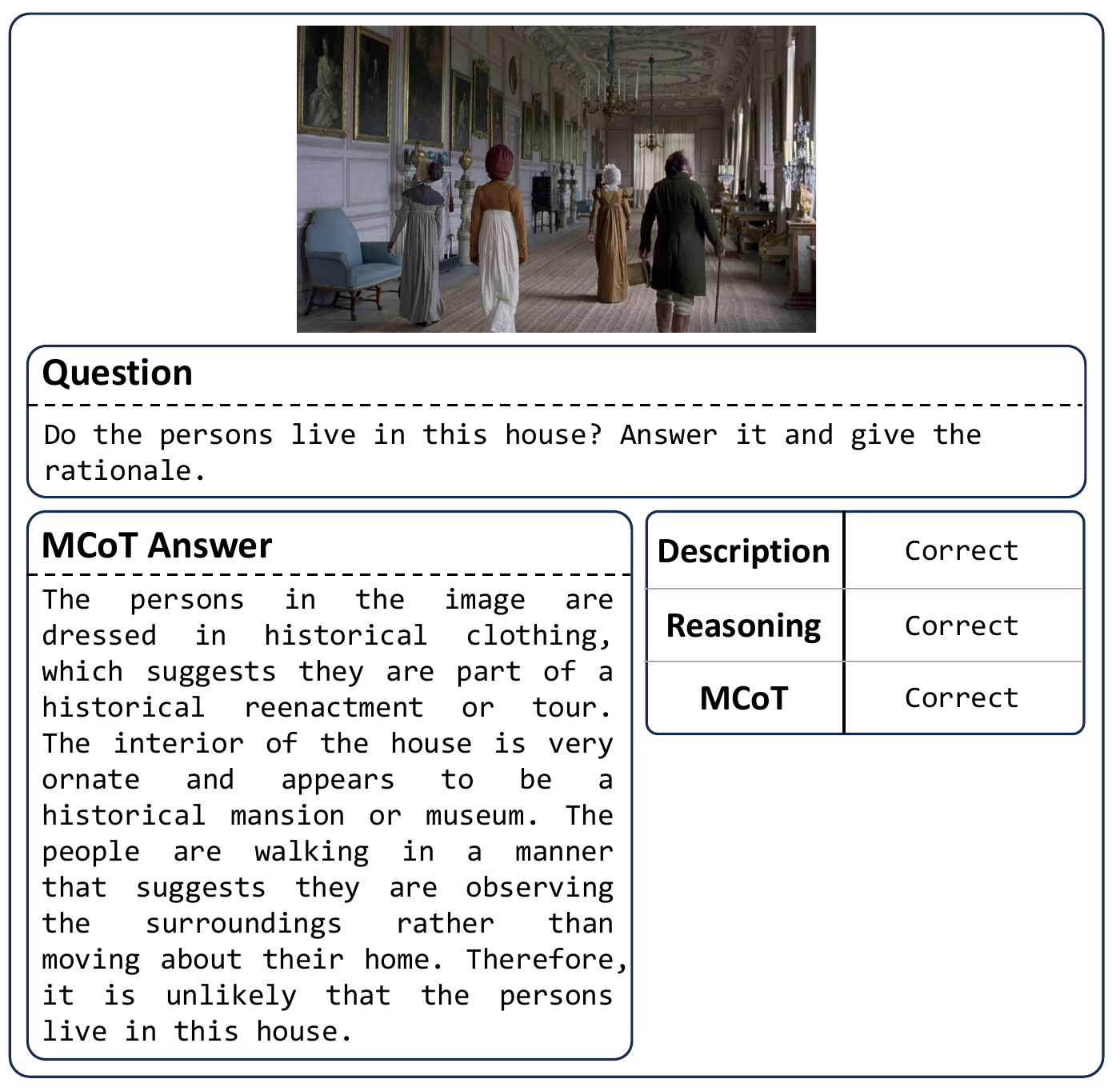}
    \caption{A $\textit{Correctness}_{type}$-based MiCEval metric high-scoring case of MiCEval-\textsc{Hard}.}
    \label{fig:type-high-hard}
\end{figure*}

\begin{figure*}[htpb]
    \centering
    \includegraphics[width=0.9\linewidth]{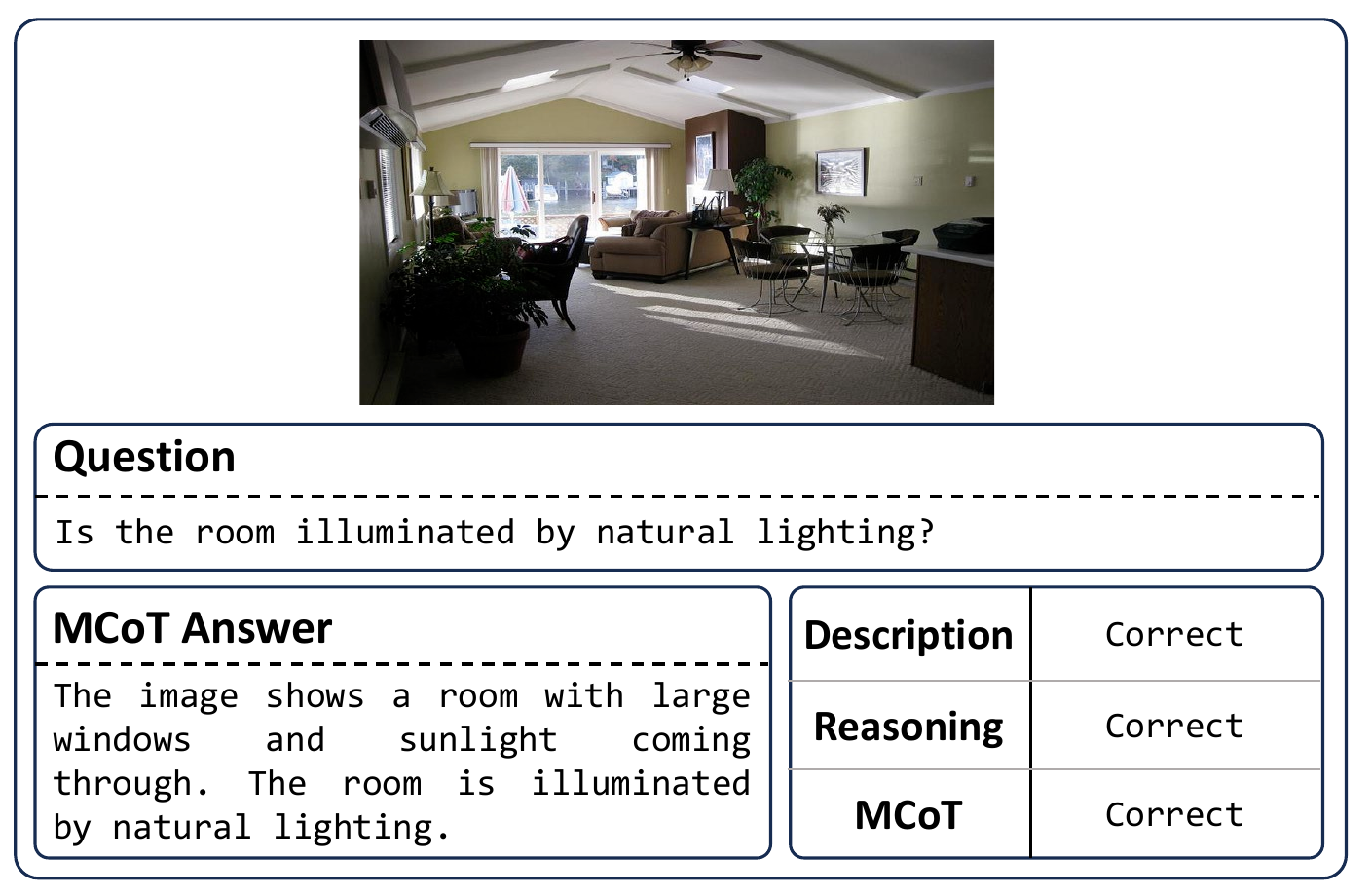}
    \caption{A $\textit{Correctness}_{type}$-based MiCEval metric high-scoring case of MiCEval-\textsc{Normal}.}
    \label{fig:type-high-normal}
\end{figure*}

\begin{figure*}[htpb]
    \centering
    \includegraphics[width=0.9\linewidth]{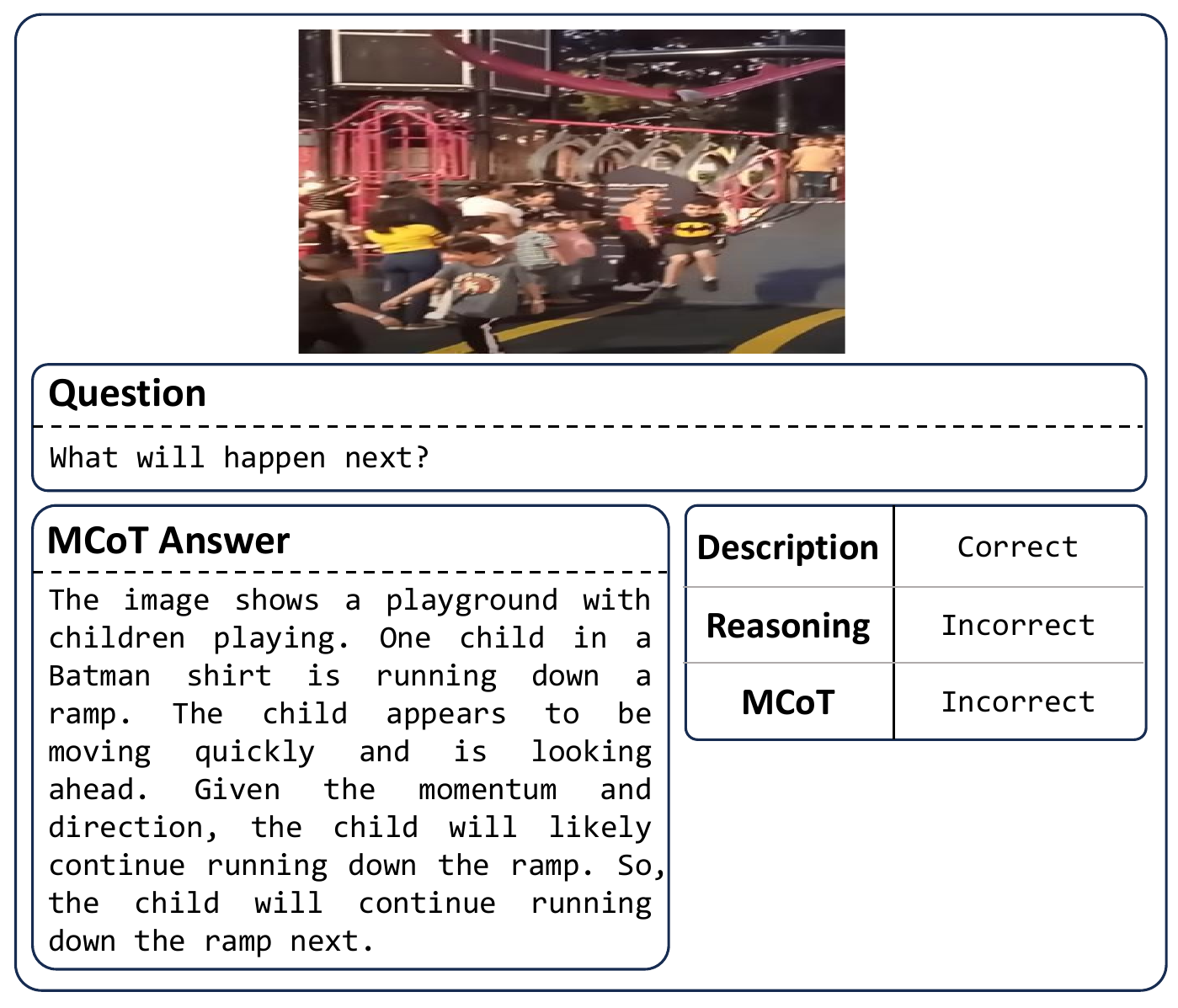}
    \caption{A $\textit{Correctness}_{type}$-based MiCEval metric low-scoring case of MiCEval-\textsc{Hard}.}
    \label{fig:type-low-hard}
\end{figure*}

\begin{figure*}[htpb]
    \centering
    \includegraphics[width=0.9\linewidth]{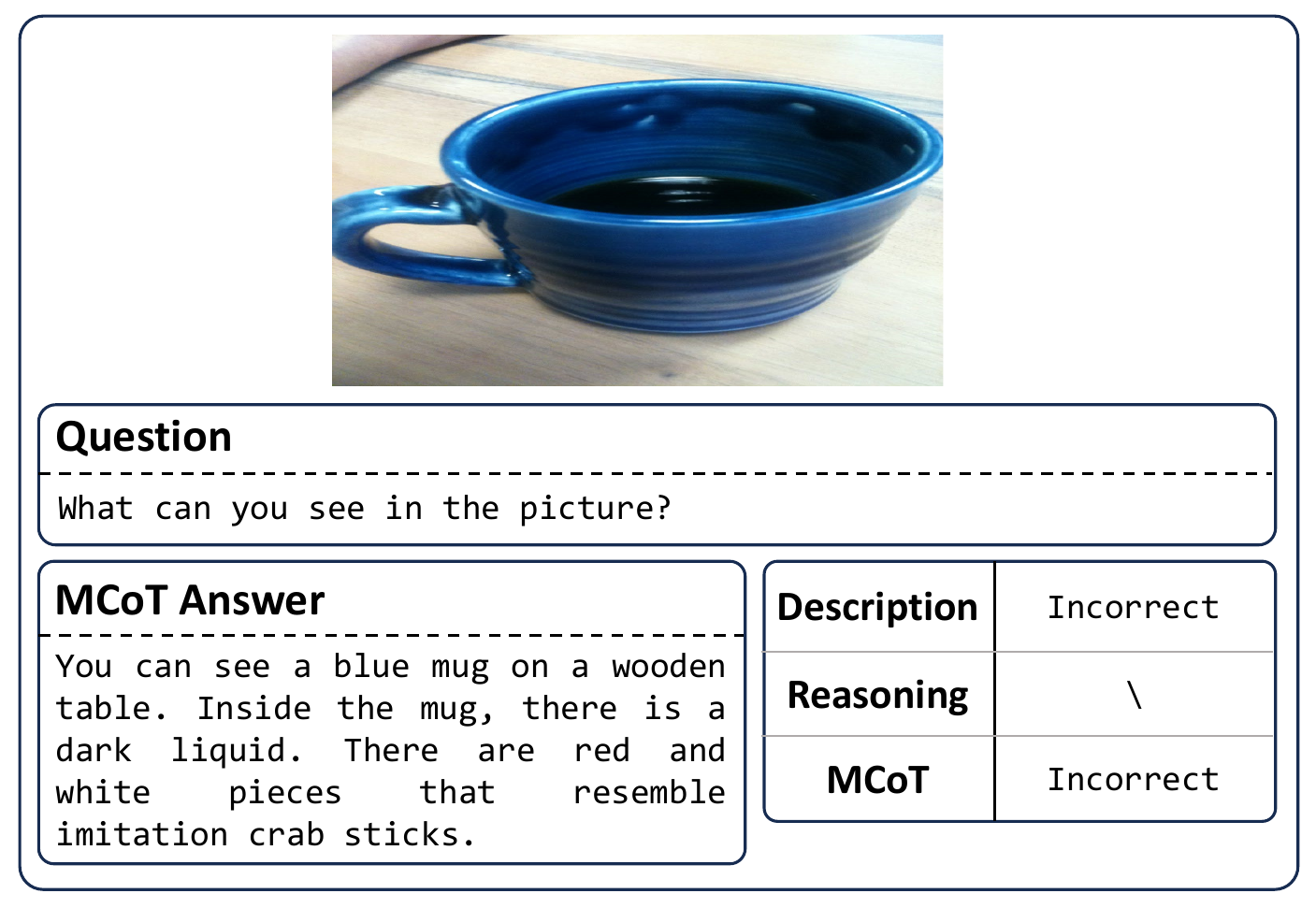}
    \caption{A $\textit{Correctness}_{type}$-based MiCEval metric low-scoring case of MiCEval-\textsc{Normal}.}
    \label{fig:type-low-normal}
\end{figure*}

\begin{figure*}[htpb]
    \centering
    \includegraphics[width=0.9\linewidth]{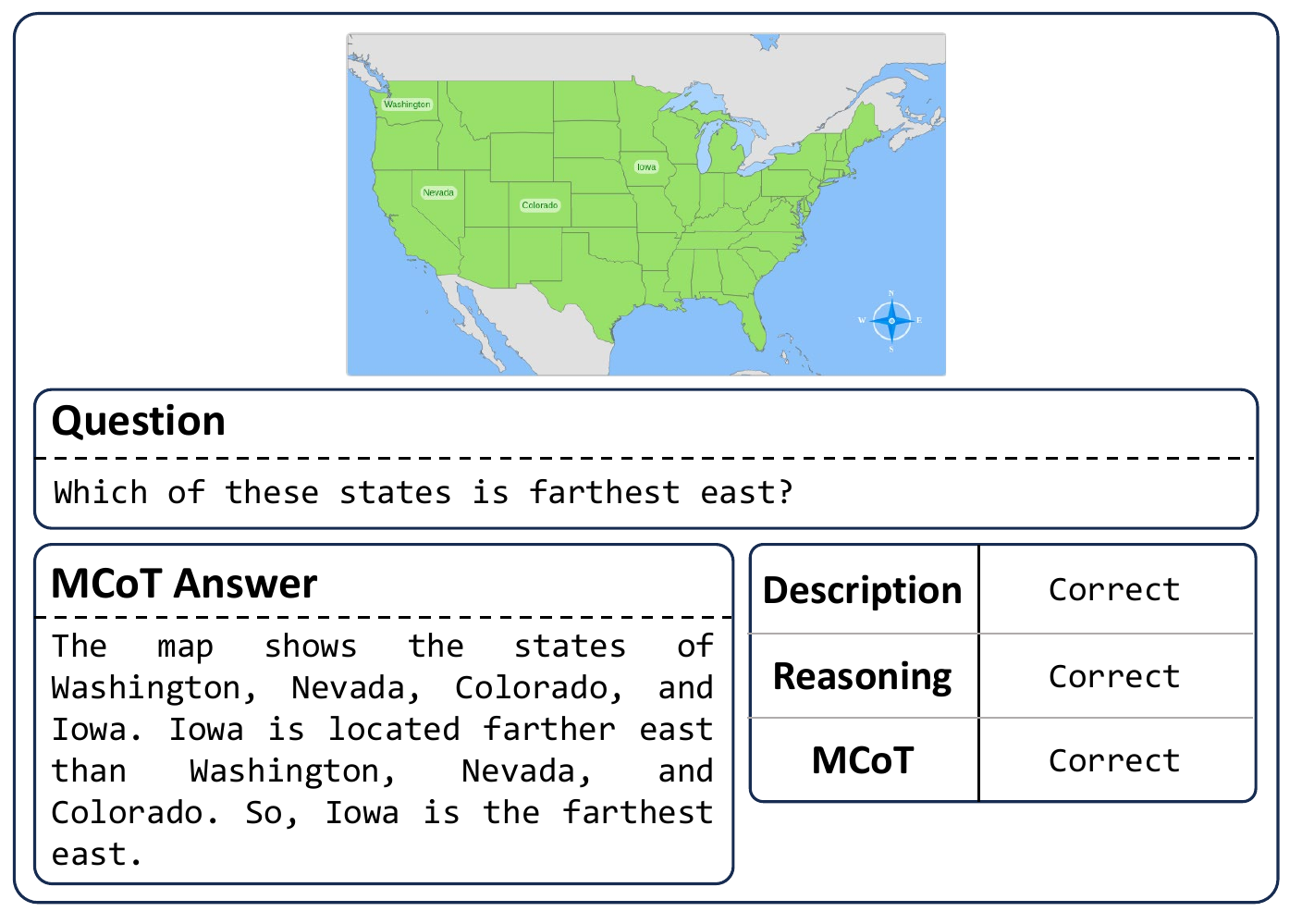}
    \caption{A $\textit{Correctness}_{all}$-based MiCEval metric high-scoring case of MiCEval-\textsc{Hard}.}
    \label{fig:all-high-hard}
\end{figure*}

\begin{figure*}[htpb]
    \centering
    \includegraphics[width=0.9\linewidth]{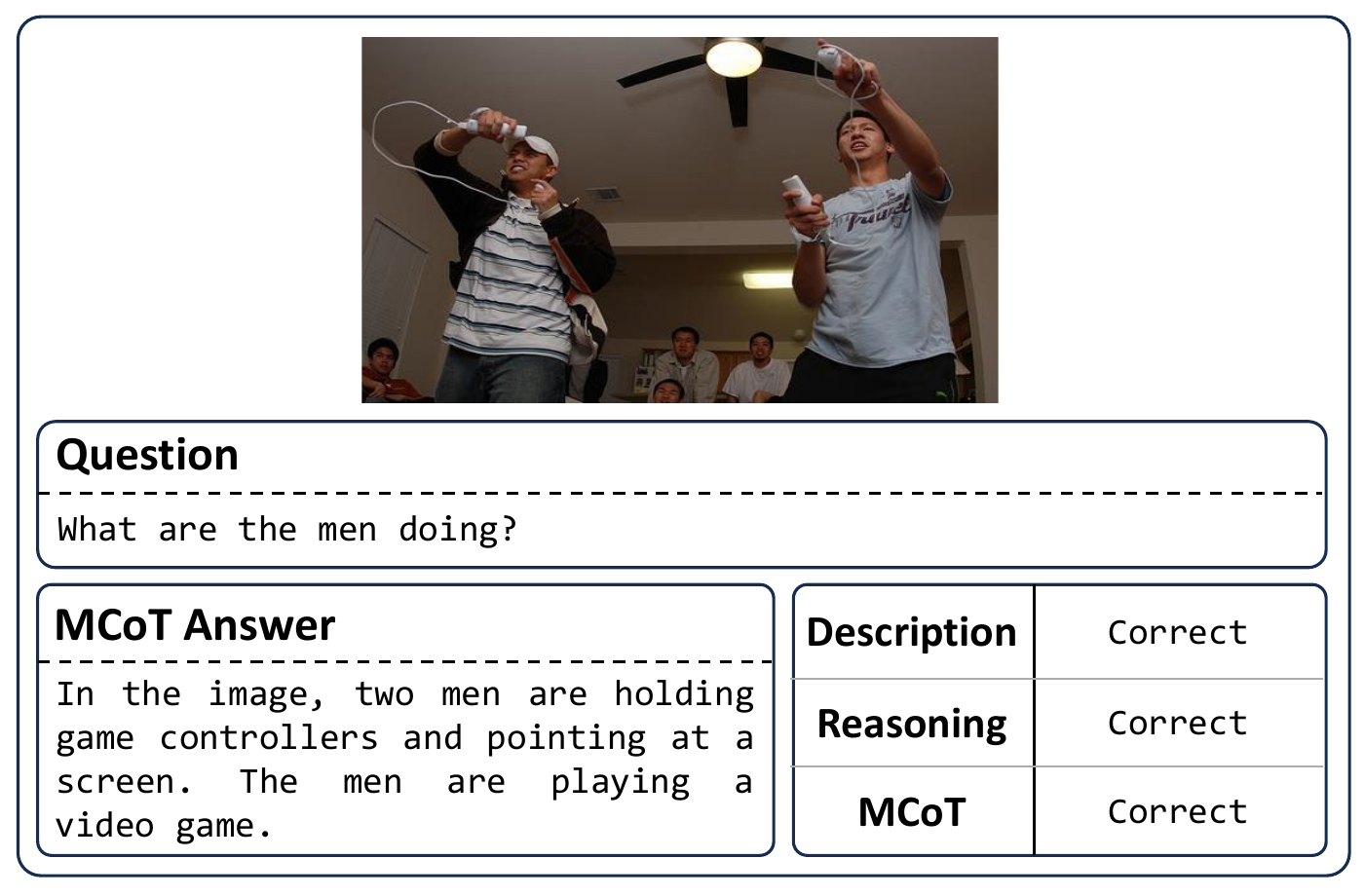}
    \caption{A $\textit{Correctness}_{all}$-based MiCEval metric high-scoring case of MiCEval-\textsc{Normal}.}
    \label{fig:all-high-normal}
\end{figure*}

\begin{figure*}[htpb]
    \centering
    \includegraphics[width=0.9\linewidth]{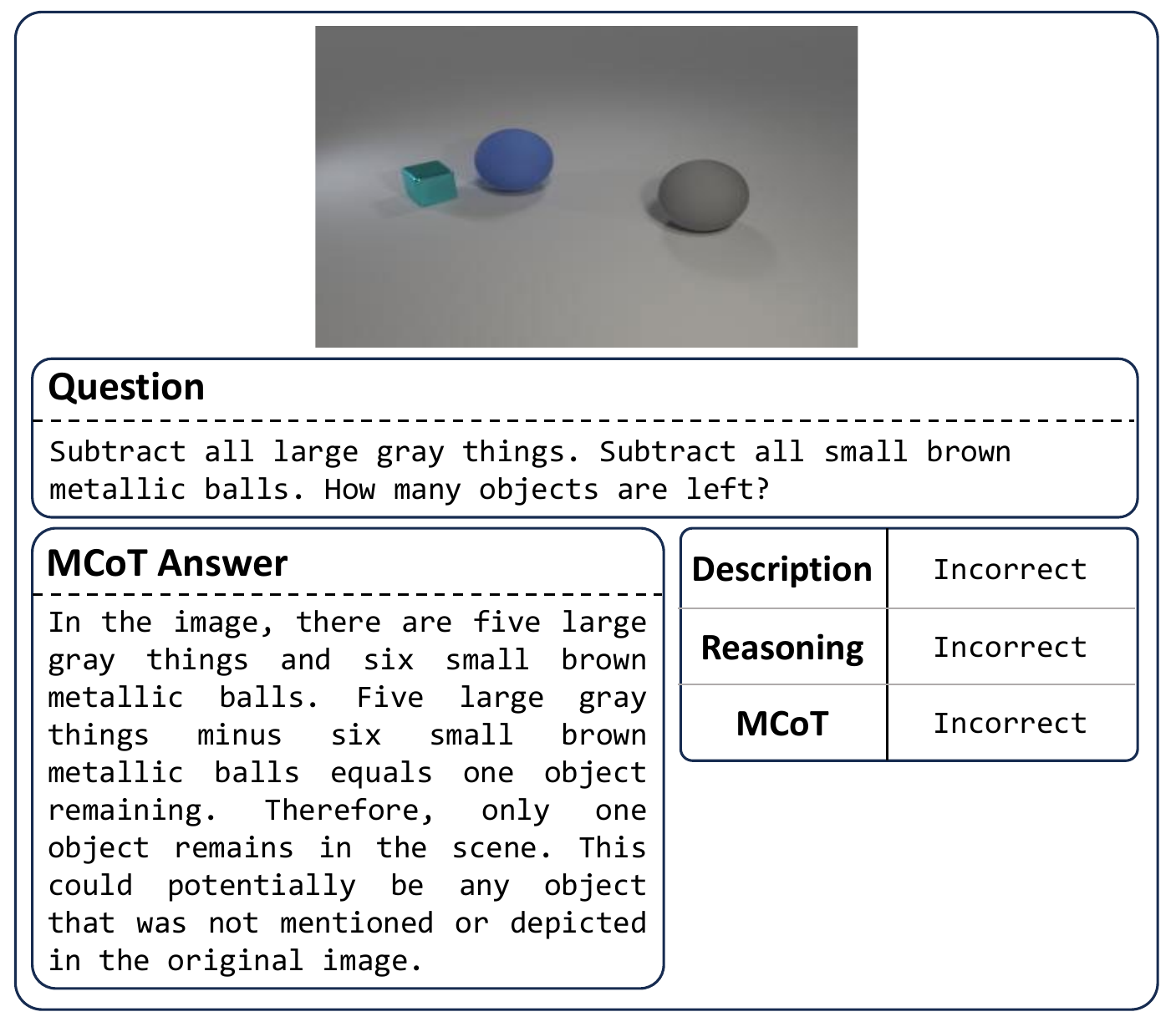}
    \caption{A $\textit{Correctness}_{all}$-based MiCEval metric low-scoring case of MiCEval-\textsc{Hard}.}
    \label{fig:all-low-hard}
\end{figure*}

\begin{figure*}[htpb]
    \centering
    \includegraphics[width=0.9\linewidth]{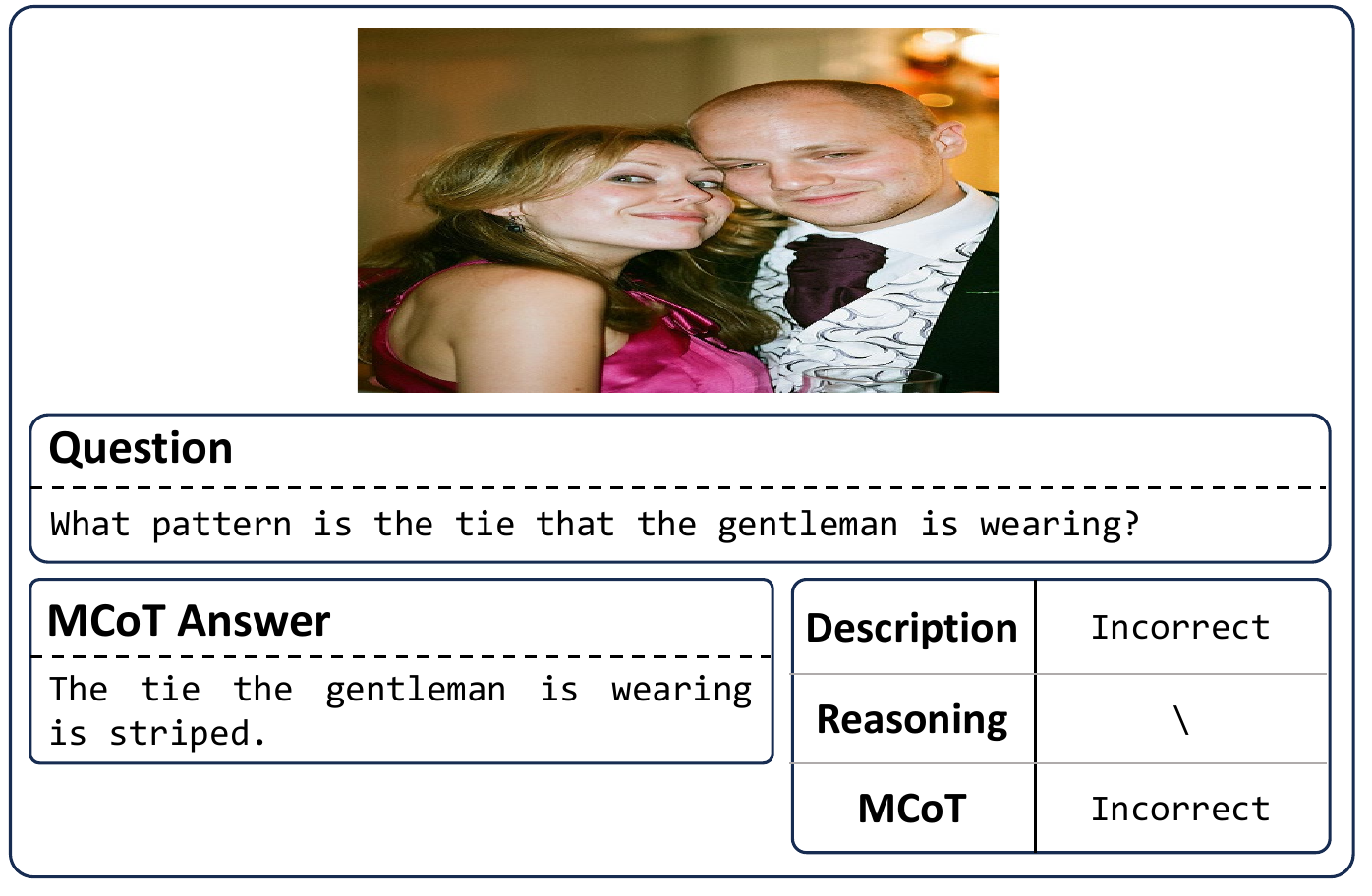}
    \caption{A $\textit{Correctness}_{all}$-based MiCEval metric low-scoring case of MiCEval-\textsc{Normal}.}
    \label{fig:all-low-normal}
\end{figure*}

In this chapter, we present examples based on two different MiCEval metrics: the high-scoring and low-scoring cases. Figures~\ref{fig:type-high-hard} to \ref{fig:type-high-normal} show the best cases based on the $\textit{Correctness}_{type}$-based MiCEval metric, while Figures~\ref{fig:type-low-hard} to \ref{fig:type-low-normal} illustrate the worst cases. The high-scoring cases for the $\textit{Correctness}_{all}$-based MiCEval metric are shown in Figures~\ref{fig:all-high-hard} to~\ref{fig:all-high-normal}, and the low-scoring cases are depicted in Figures~\ref{fig:all-low-hard} to \ref{fig:all-low-normal}.

\clearpage

\begin{table*}[h!]
  \begin{center}
    \begin{tabularx}{\textwidth}{l|X|
    >{\raggedright\arraybackslash}X}
    \toprule
\textbf{Dataset} & \textbf{Description} & \textbf{Question} \\
\midrule
{Visual7W} & {A dataset designed for grounded VQA.} & Where could this be? \\
~ & ~ & What is in front of the building? \\
\midrule
{VSR} & Visual spatial reasoning VQA. & Is the following statement true or false? Statement: The laptop is at the left side of the chair. \\
~ & ~ & Is the following statement true or false? Statement: The train is behind the motorcycle. \\
\midrule
{ScienceQA} & VQA centers around science topics. & Which of the following fossils is oldest? \\
            &                                      & Which property do these three objects have in common? \\
\midrule
{MMVP} & A dataset aims to detect obvious visual differences in images. & Is the school bus driving towards or away from the camera? \\
       &   & In the picture, are the elderly people moving to the left or to the right? \\
\midrule
{MMstar} & A dataset designed with challenging VQA.  & What might the woman in the image be doing? \\
         &   & How many types of fruits are there in the image? \\
\midrule
{MM-Vet} & A dataset featuring diverse question types and corresponding answers. & What might the woman in the image be doing? \\
         &   & How many types of fruits are there in the image? \\
        
\midrule
{VQAv2} & A dataset with questions covering various aspects of the image. & What type of bird is on the car? \\
        &                                               & What is this form of transportation? \\
\midrule
{Vizwiz} & A dataset with photos lacking clarity and exhibiting suboptimal quality collected from users. & What is this color please? \\
         &                                                   & What is this box? \\

    \bottomrule
    \end{tabularx}
    \caption{Detailed descriptions and randomly sampled questions of 8 source datasets.}
    \label{tab:sample-source}
  \end{center} 
\end{table*}

\begin{table*}[h]
% textwidth = 16.5cm
\centering
\small
\rowcolors{2}{white!8}{brown!5}
\begin{tabularx}{\textwidth}{
 >{\centering\arraybackslash}m{0.15\textwidth} 
>{\centering\arraybackslash}m{0.08\textwidth}
m{0.18\textwidth}
m{0.36\textwidth}
m{0.1\textwidth} 
}
\toprule
\textbf{Image} & \textbf{Source Dataset} & \textbf{Question} & \textbf{MCoT Answer}&\textbf{Error} \\
\midrule

\includegraphics[width=3cm,height=2cm]{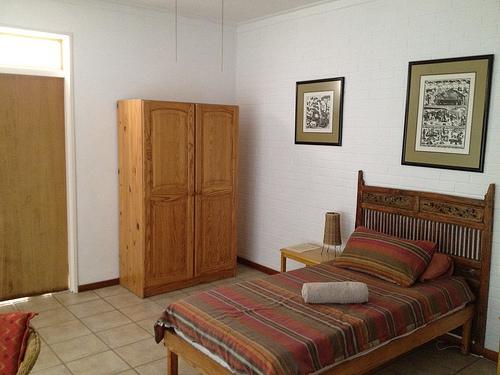} 
& {Visual7w} 
& When is this room used? 
&  {In the bedroom, there is a bed covered by a striped quilt. On both sides of the headboard hang two pictures. A wooden cabinet stands next to the door. Near the window, there is an armchair placed against the wall.}
& {Entity False (step 4)} \\
\midrule

\includegraphics[width=3cm,height=2cm]{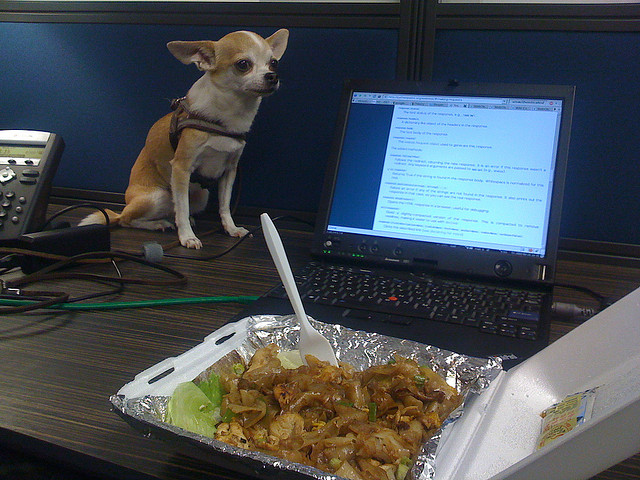} 
&{VSR}
& Is the following statement true or false? Statement: The dog is at the right side of the laptop. 
& {A small brown chihuahua sits behind an open black laptop computer. 
The laptop screen displays text that reads “The responsibilities include...”. To the left of the laptop, there's a telephone receiver. On the desk where these objects rest, there appears to be food from takeout containers.}
& {Image Relevant (step2, step3, step4)} \\
\midrule

\includegraphics[width=3cm,height=2cm]{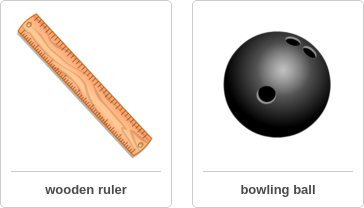} 
&{ScienceQA}
& Which property do these two objects have in common?
& {Both the wooden ruler and the bowling ball are objects. They are both made of wood. The wooden ruler is a tool used for measuring. The bowling ball is a sports equipment used for playing bowling. Both the wooden ruler and the bowling ball are made of wood and have different functions. }
& {Inter-step Incorrect (step2, step 5)} \\
\midrule

\includegraphics[width=3cm,height=2cm]{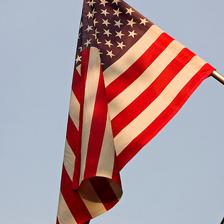} 
&{MMVP}
& Is the wind blowing the flag?
& {The flag in the image is not hanging straight down. It appears to be extended outward and has a wavy appearance. This suggests that there is some force acting on it. Therefore, the wind is blowing the flag.}
& {Inter-step Incorrect (step 4)} \\
\midrule

\includegraphics[width=3cm,height=2cm]{figures/appendix/hard_mmvet.jpg} 
&{MM-Vet}
& Are the trees taller than the giraffes? 
& {The giraffes are taller than the trees. The trees are shorter than the giraffes. Therefore, the giraffes are taller than the trees.}
& {Uninformative (step 3)} \\
\midrule

\includegraphics[width=3cm,height=2cm]{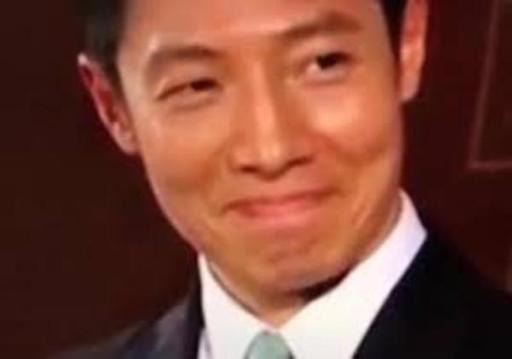} 
&{MMstar}
& What will happen next?
& {The man in the image is frowning and pouting. He appears to be upset and unhappy. Therefore, he might cry next.}
& {Attribute False (step 1), Inter-step Incorrect (step2, step3)} \\
\midrule
\end{tabularx}
\setlength{\belowcaptionskip}{0.2cm}
\caption{Randomly sampled questions from MiCEval-\textsc{Hard}.}
\label{tab:hard-sample}
\end{table*}

\begin{table*}[h]
% textwidth = 16.5cm
\centering
\small
\rowcolors{2}{white!8}{brown!5}
\begin{tabularx}{\textwidth}{
 >{\centering\arraybackslash}m{0.15\textwidth} 
>{\centering\arraybackslash}m{0.08\textwidth}
m{0.18\textwidth}
m{0.36\textwidth}
m{0.1\textwidth} 
}
\toprule
\textbf{Image} & \textbf{Source Dataset} & \textbf{Question} & \textbf{MCoT Answer}&\textbf{Error} \\
\midrule

\includegraphics[width=3cm,height=2cm]{figures/appendix/normal_v7w1.jpg} 
& {Visual7w} 
& What color are the middle suitcases?
&  {In the image, there are several suitcases in the trunk of a car. The middle suitcases are red.}
& {Attribute False (step 2)} \\
\midrule

\includegraphics[width=3cm,height=2cm]{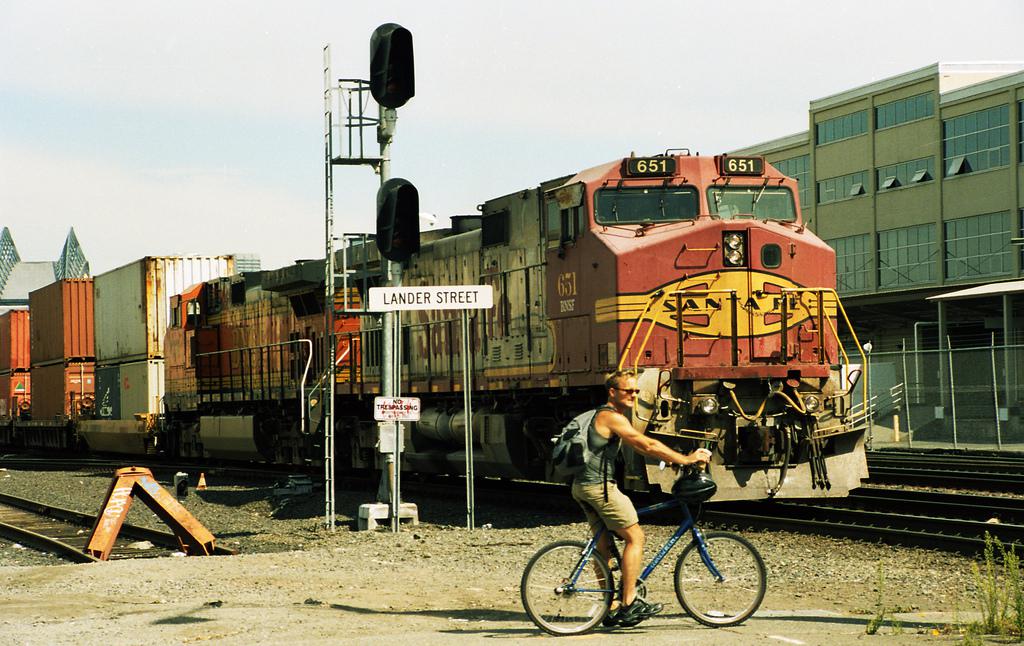} 
&{Visual7W}
& What is growing on the right? 
& {There are buildings on the right side of the image. There are trees growing next to the buildings. }
& {Entity False (step 2)} \\
\midrule

\includegraphics[width=3cm,height=2cm]{figures/appendix/normal_vqav1.jpg} 
&{VQAv2}
& Does the cat belong there?
& {The image shows a cat relaxing comfortably in a home setting. The cat belongs there.}
& {Inter-step Incorrect (step 2)} \\
\midrule

\includegraphics[width=3cm,height=2cm]{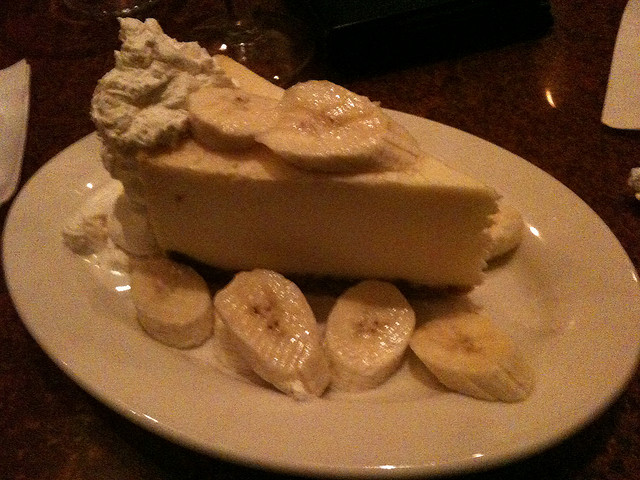} 
&{VQAv2}
& What type of fruit is sliced and laying around the cheesecake?
& {There's a plate containing food. On that plate, several banana slices are visible. These bananas have been cut into chunks or slices. Therefore, it appears as if they're being used to garnish or accompany the dessert.}
& {Image Relevant (step 4)} \\
\midrule

\includegraphics[width=3cm,height=2cm]{figures/appendix/normal_viz1.jpg} 
&{Vizwiz}
& {Is is a box? Is it a box?}
& {The image shows a wooden structure with three visible sides. The structure resembles the inside of a box. This is a box.}
& {Inter-step Incorrect (step 2, step 3)} \\
\midrule

\includegraphics[width=3cm,height=2cm]{figures/appendix/normal_viz2.jpg} 
&{Vizwiz}
& {What is this?}
& {The image shows a can of Progresso chicken and noodle soup. The can has a blue label with white text. This is a can of Progresso chicken and noodle soup.}
& {Uninformative (step 3)} \\
\midrule

\end{tabularx}
\setlength{\belowcaptionskip}{0.2cm}
\caption{Randomly sampled questions from MiCEval-\textsc{Normal}.}
\label{tab:normal-sample}
\end{table*}

\begin{table*}[h]
\centering
\begin{tabular}{l|rr}
    \toprule
\textbf{Metrics} & \textbf{Description Step} & \textbf{Reasoning Step}\\
 \midrule
CLIP & 0.051 & 0.031  \\
BLIP2-ITM & 0.035 & -0.003  \\
BLIP2-ITC & 0.071 & 0.056  \\
ReCEval & -0.012 & -0.040 \\
LLM-Score & 0.130 & 0.141 \\
 \midrule
MiniCPM-V-2.6 (CoT, w/o MiCEval) & - & -  \\
LLaVA-1.6-Mistral-7B (CoT, w/o MiCEval) & - & -  \\
Llama-3.2-11B-Vision-Instruct (CoT, w/o MiCEval) & - & -  \\
GPT-4o (CoT, w/o MiCEval) & - & -  \\
 \midrule
MiniCPM-V-2.6 (Step, w/o MiCEval) & 0.110 & 0.103 \\
LLaVA-1.6-Mistral-7B (Step, w/o MiCEval) & 0.109 & -0.025 \\
Llama-3.2-11B-Vision-Instruct (Step, w/o MiCEval) & 0.005 & 0.126  \\
GPT-4o (Step, w/o MiCEval) & 0.168 & 0.298  \\
 \midrule
MiniCPM-V-2.6 (Step, MiCEval) & 0.240 & 0.183 \\
LLaVA-1.6-Mistral-7B (Step, MiCEval) & 0.098 & -0.044\\
Llama-3.2-11B-Vision-Instruct (Step, MiCEval) & 0.115 & 0.098 \\
GPT-4o (Step, MiCEval) & \textbf{0.240} & \textbf{0.319}\\

    \bottomrule
\end{tabular}
        \setlength{\belowcaptionskip}{0.2cm}
    \caption{Human-annotation-based Somer's D results on Description Step, Reasoning Step.}
    \label{tab:gt-step-somersd}
\end{table*}

\begin{table*}[h]
\centering
\begin{tabular}{l|ccc}
\toprule
    \textbf{Metric} & \textbf{NORMAL} 
    & \textbf{HARD} &\textbf{\textsc{MiCEval}} \\
 \midrule
    CLIP & 0.079 & 0.019 & 0.060 \\
    BLIP2-ITM & 0.088 & -0.065 & 0.031 \\
    BLIP2-ITC & 0.072 & -0.006 & 0.075 \\
    ReCEval & -0.006 & 0.015 & 0.040 \\
    LLM-Score & 0.078 & 0.210 & 0.162 \\
 \midrule
    MiniCPM-V-2.6 (CoT, w/o MiCEval) & 0.154 & 0.123 & 0.130 \\
    LLaVA-1.6-Mistral-7B (CoT, w/o MiCEval) & 0.109 & -0.024 & 0.080 \\
    Llama-3.2-11B-Vision-Instruct (CoT, w/o MiCEval) & 0.090 & 0.218 & 0.176 \\
    GPT-4o (CoT, w/o MiCEval) & 0.154 & 0.256 & 0.208 \\
 \midrule
    MiniCPM-V-2.6 (Step, w/o MiCEval) & 0.112 & 0.123 & 0.169 \\
    LLaVA-1.6-Mistral-7B (Step, w/o MiCEval) & 0.083 & 0.178 & 0.140 \\
    Llama-3.2-11B-Vision-Instruct (Step, w/o MiCEval) & 0.031 & 0.090 & 0.075 \\
    GPT-4o (Step, w/o MiCEval) & 0.178 & \textbf{0.282} & 0.257 \\
 \midrule
    MiniCPM-V-2.6 (Step, MiCEval) & \textbf{0.260} & 0.198 & 0.279 \\
    LLaVA-1.6-Mistral-7B (Step, MiCEval) & 0.080 & 0.142 & 0.126 \\
    Llama-3.2-11B-Vision-Instruct (Step, MiCEval) & 0.158 & 0.170 & 0.192 \\
    GPT-4o (Step, MiCEval) & 0.256 & 0.264 & \textbf{0.281} \\
\bottomrule
\end{tabular}
   \setlength{\belowcaptionskip}{0.2cm}
    \caption{Human-annotation-based Somer's D results on MiCEval dataset.}
    \label{tab:gt-mcot-somersd}
\end{table*}

\begin{table*}[h]
\centering
\begin{tabular}{l|rr}
        \toprule
    \textbf{Model} & \textbf{Description Step} & \textbf{Reasoning Step} \\
    \midrule
    CLIP & 0.105 & 0.055 \\
    BLIP2-itm & 0.073 & -0.005 \\
    BLIP2-itc & 0.148 & 0.100 \\
    ReCEval & -0.025 & -0.072 \\
    LLM-Score & 0.245 & 0.232 \\
      \midrule
    MiniCPM-V-2.6 (CoT, w/o MiCEval) & -- & -- \\
    LLaVA-1.6-Mistral-7B (CoT, w/o MiCEval) & -- & -- \\
    Llama-3.2-11B-Vision-Instruct (CoT, w/o MiCEval) & -- & -- \\
    GPT-4o (CoT, w/o MiCEval) & -- & -- \\
      \midrule
    MiniCPM-V-2.6 (Step, w/o MiCEval) & 0.197 & 0.158 \\
    LLaVA-1.6-Mistral-7B (Step, w/o MiCEval) & 0.178 & -0.035 \\
    Llama-3.2-11B-Vision-Instruct (Step, w/o MiCEval) & 0.006 & 0.124 \\
    GPT-4o (Step, w/o MiCEval) & 0.276 & \textbf{0.408} \\
      \midrule
    MiniCPM-V-2.6 (Step, MiCEval) & 0.249 & 0.153 \\
    LLaVA-1.6-Mistral-7B (Step, MiCEval) & 0.162 & -0.051 \\
    Llama-3.2-11B-Vision-Instruct (Step, MiCEval) & 0.194 & 0.135 \\
    GPT-4o (Step, MiCEval) & \textbf{0.365} & 0.367 \\
            \bottomrule
\end{tabular}
        \setlength{\belowcaptionskip}{0.2cm}
    \caption{Spearman's Rank Correlation Coefficient on Description Step, Reasoning Step.}
    \label{tab:gt-spearman-step}
\end{table*}

\begin{table*}[h]
\centering
\begin{tabular}{lccc}
\toprule
\textbf{Metrix} & \textbf{NORMAL} & \textbf{HARD} & \textbf{MiCEval} \\
\midrule
CLIP & 0.148 & 0.033 & 0.105 \\
BLIP2-itm & 0.165 & -0.114 & 0.054 \\
BLIP2-itc & 0.134 & -0.011 & 0.131 \\
ReCEval & -0.011 & 0.026 & 0.069 \\
LLM-Score & 0.144 & 0.369 & 0.279 \\
\midrule
MiniCPM-V-2.6 (CoT, w/o MiCEval) & 0.227 & 0.182 & 0.186 \\
LLaVA-1.6-Mistral-7B (CoT, w/o MiCEval) & 0.117 & -0.028 & 0.086 \\
Llama-3.2-11B-Vision-Instruct (CoT, w/o MiCEval) & 0.130 & 0.281 & 0.234 \\
GPT-4o (CoT, w/o MiCEval) & 0.238 & 0.384 & 0.305 \\
\midrule
MiniCPM-V-2.6 (Step, w/o MiCEval) & 0.204 & 0.211 & 0.289 \\
LLaVA-1.6-Mistral-7B (Step, w/o MiCEval) & 0.146 & 0.303 & 0.232 \\
Llama-3.2-11B-Vision-Instruct (Step, w/o MiCEval) & 0.045 & 0.141 & 0.110 \\
GPT-4o (Step, w/o MiCEval) & 0.287 & \textbf{0.409} & 0.381 \\
\midrule
MiniCPM-V-2.6 (Step, MiCEval) & 0.263 & 0.259 & 0.320 \\
LLaVA-1.6-Mistral-7B (Step, MiCEval) & 0.130 & 0.221 & 0.192 \\
Llama-3.2-11B-Vision-Instruct (Step, MiCEval) & 0.266 & 0.263 & 0.300 \\
GPT-4o (Step, MiCEval) & \textbf{0.408} & 0.383 & \textbf{0.413} \\
\bottomrule
\end{tabular}
\setlength{\belowcaptionskip}{0.2cm}
\caption{Spearman's Rank Correlation Coefficient on MiCEval dataset.}
\label{tab:gt-spearman-dataset}
\end{table*}

\begin{table*}[htb]
\centering
\begin{tabular}{l|c}

\toprule
\textbf{Model} & \textbf{Acc.} \\
\midrule
CLIP & 0.700 \\
BLIP2-ITM & 0.657 \\
BLIP2-ITC & 0.657 \\
ReCEval & 0.600 \\
LLM-Score & 0.614 \\
\midrule
MiniCPM-V-2.6 (CoT, w/o MiCEval) & 0.686 \\
LLaVA-1.6-Mistral-7B (CoT, w/o MiCEval) & 0.686 \\
Llama-3.2-11B-Vision-Instruct (CoT, w/o MiCEval) & 0.729 \\
GPT-4o (CoT, w/o MiCEval) & 0.757\\
\midrule
MiniCPM-V-2.6 (Step, w/o MiCEval) & 0.700 \\
LLaVA-1.6-Mistral-7B (Step, w/o MiCEval) & 0.571 \\
Llama-3.2-11B-Vision-Instruct (Step, w/o MiCEval) & 0.757 \\
GPT-4o (Step, w/o MiCEval) & 0.771 \\
\midrule
MiniCPM-V-2.6 (Step, MiCEval) & \textbf{0.800} \\
LLaVA-1.6-Mistral-7B (Step, MiCEval) & 0.586 \\
Llama-3.2-11B-Vision-Instruct (Step, MiCEval) & 0.786 \\
GPT-4o (Step, MiCEval) & \textbf{0.800} \\

\bottomrule
\end{tabular}
    
\setlength{\belowcaptionskip}{0.2cm}
\caption{Accuracy based on human-annotation labels of different evaluation metrics on Choice Ranking.}
\label{tab:gt-choice-ranking}
\end{table*}

\begin{figure*}[htpb]
    \centering
    \includegraphics[width=0.9\linewidth]{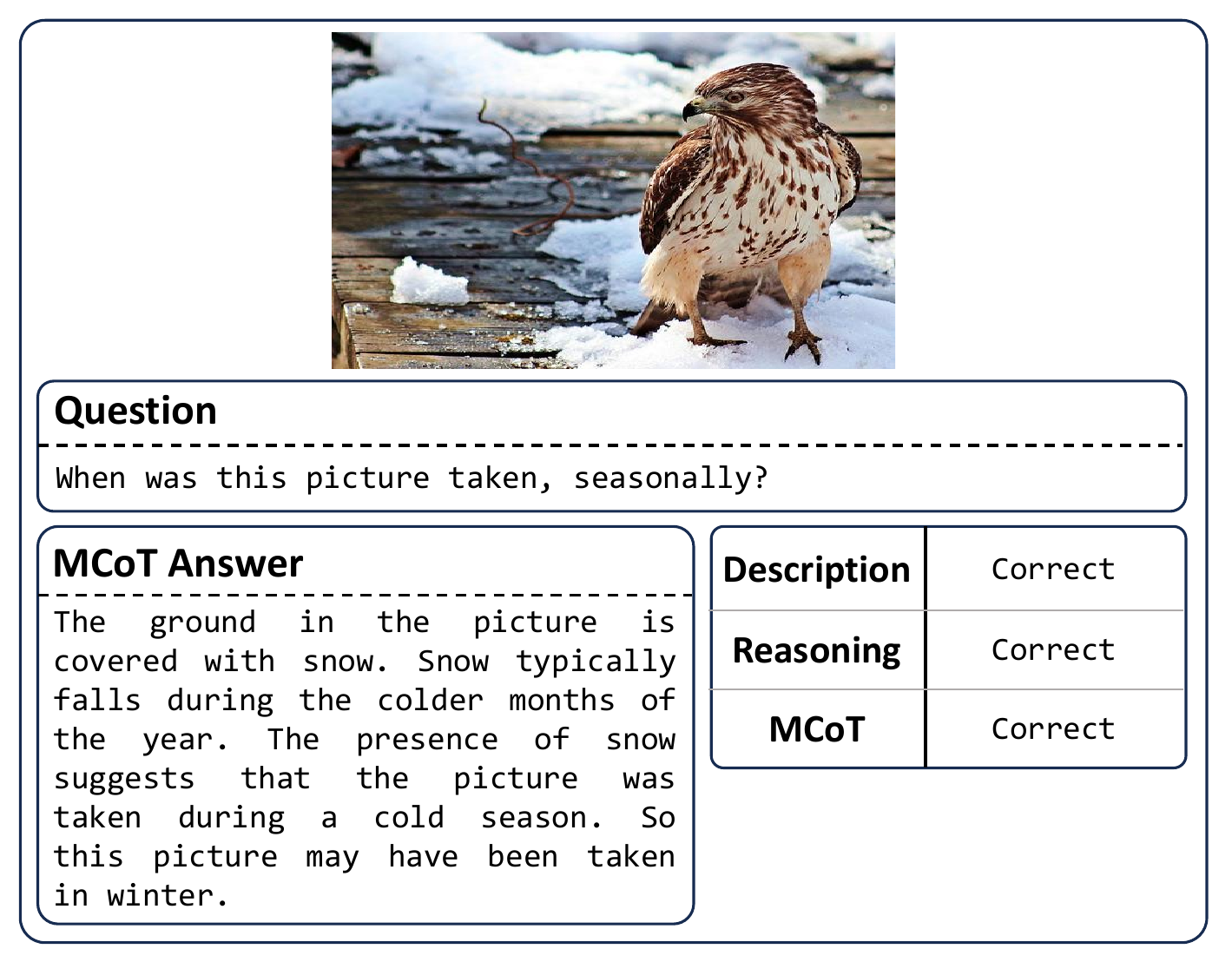}
    \caption{A $\textit{Correctness}_{type}$-based MiCEval metric high-scoring case of MiCEval-\textsc{Hard}.}
\end{figure*}

\begin{figure*}[htpb]
    \centering
    \includegraphics[width=0.9\linewidth]{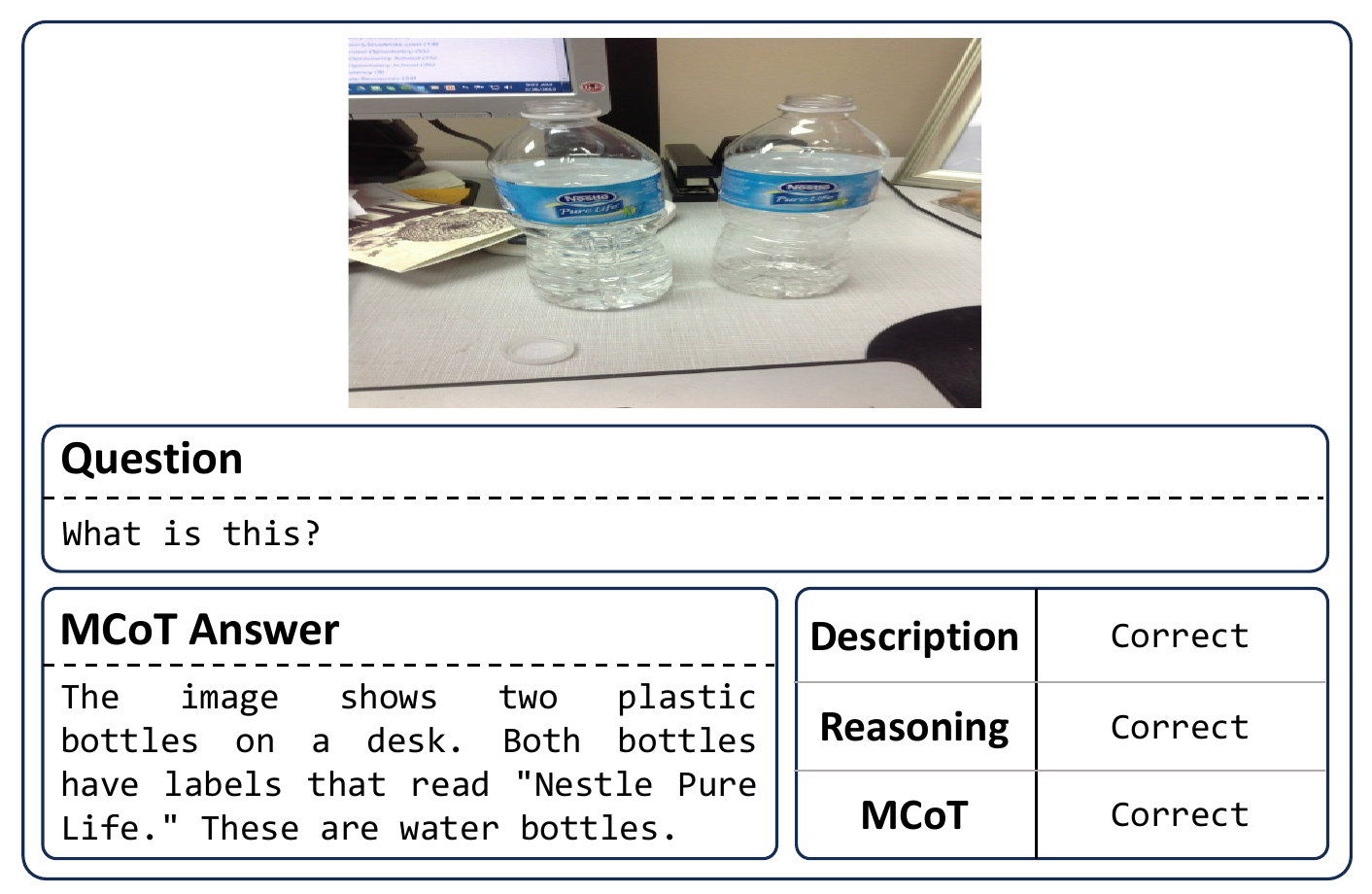}
    \caption{A $\textit{Correctness}_{type}$-based MiCEval metric high-scoring case of MiCEval-\textsc{Normal}.}
\end{figure*}

\begin{figure*}[htpb]
    \centering
    \includegraphics[width=0.9\linewidth]{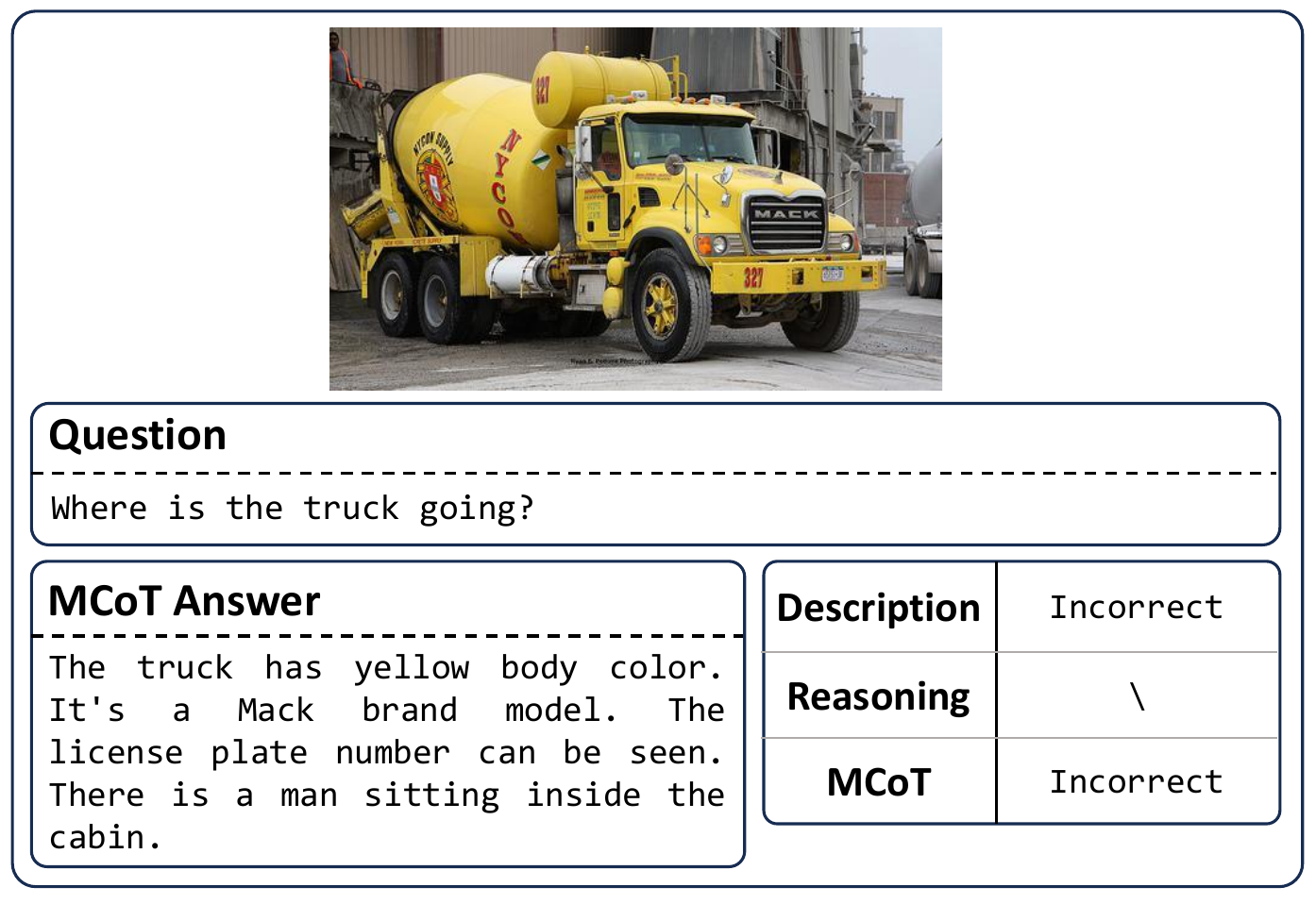}
    \caption{A $\textit{Correctness}_{type}$-based MiCEval metric low-scoring case of MiCEval-\textsc{Hard}.}
\end{figure*}

\begin{figure*}[htpb]
    \centering
    \includegraphics[width=0.9\linewidth]{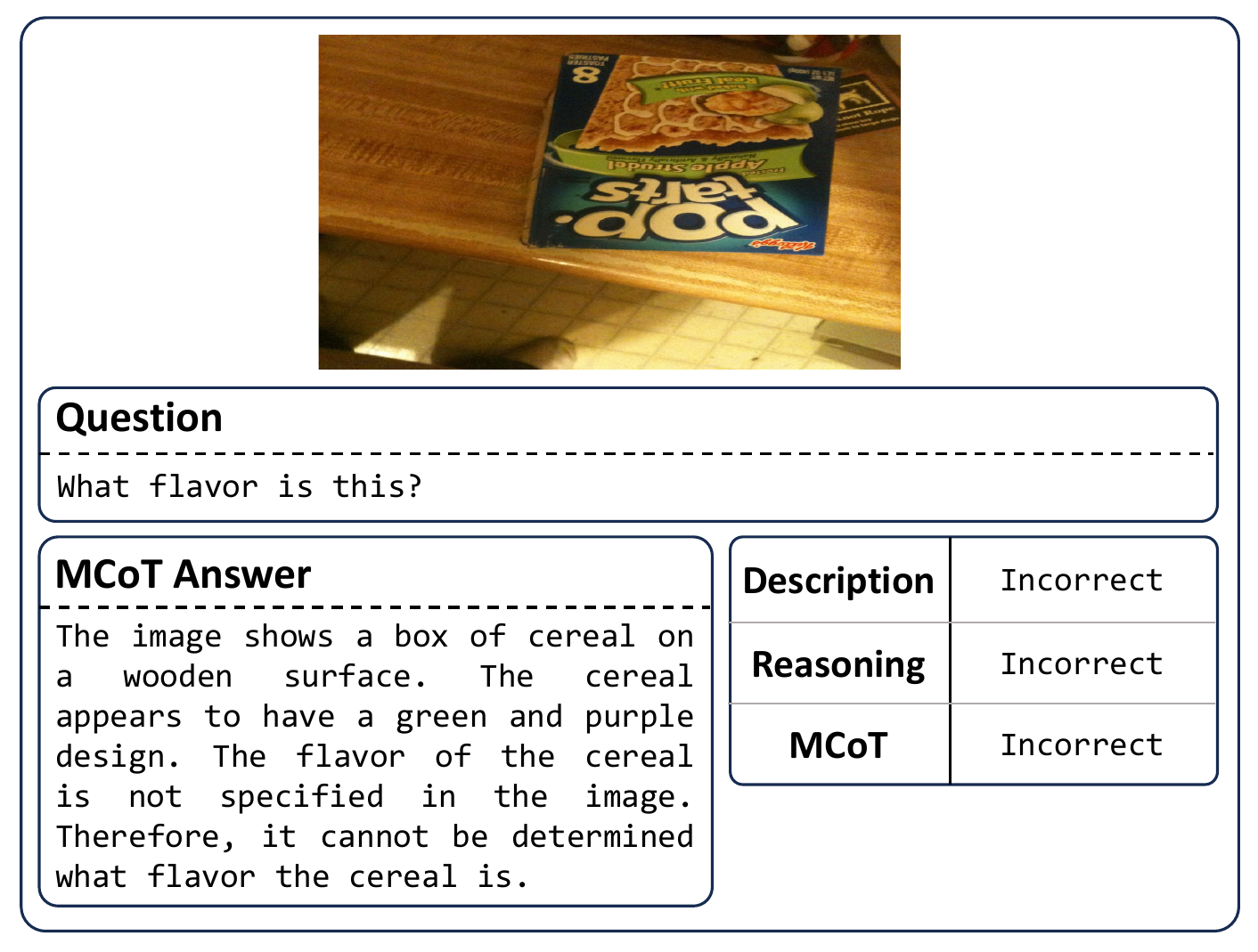}
    \caption{A $\textit{Correctness}_{type}$-based MiCEval metric low-scoring case of MiCEval-\textsc{Normal}.}
\end{figure*}

\begin{figure*}[htpb]
    \centering
    \includegraphics[width=0.9\linewidth]{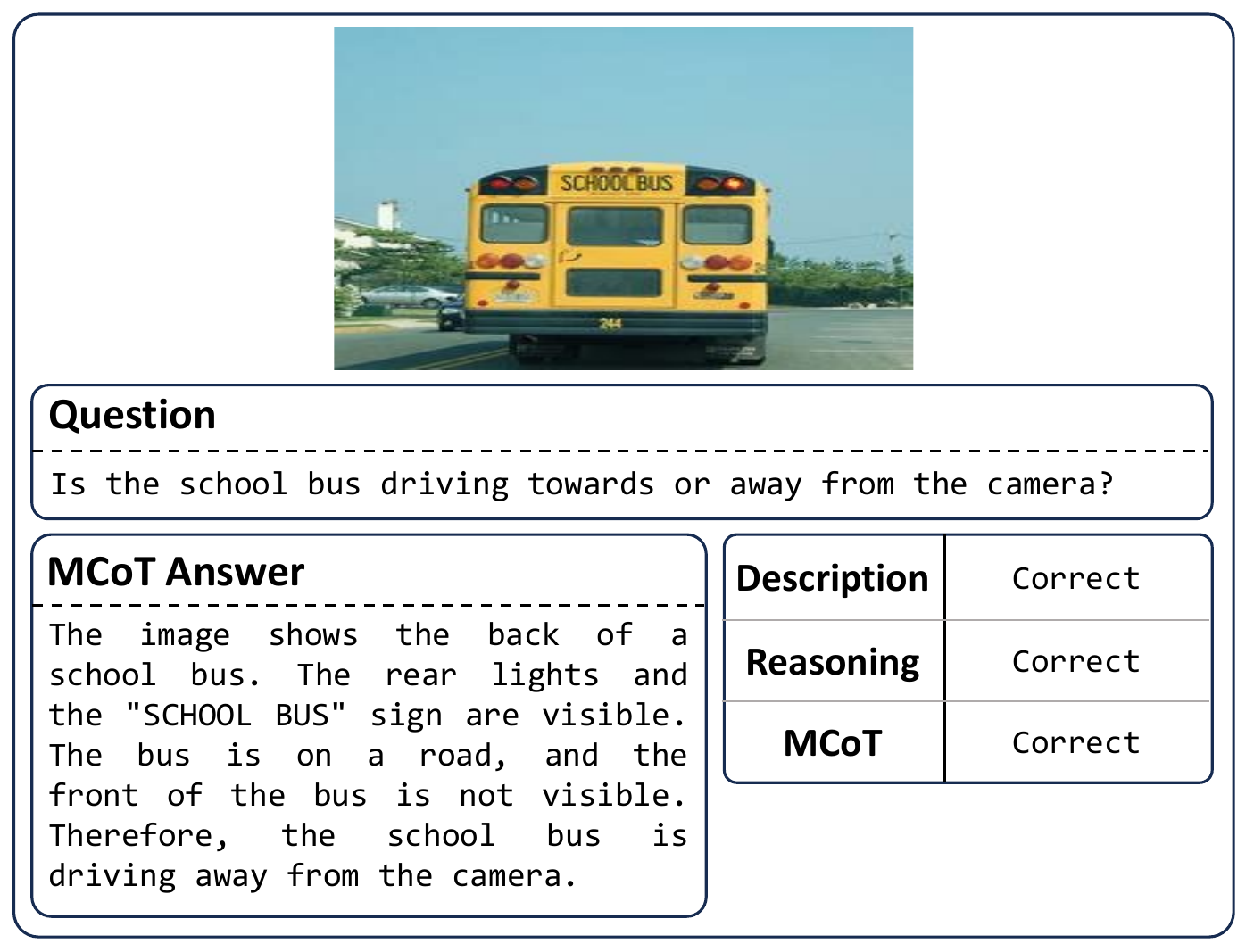}
    \caption{A $\textit{Correctness}_{all}$-based MiCEval metric high-scoring case of MiCEval-\textsc{Hard}.}
\end{figure*}

\begin{figure*}[htpb]
    \centering
    \includegraphics[width=0.9\linewidth]{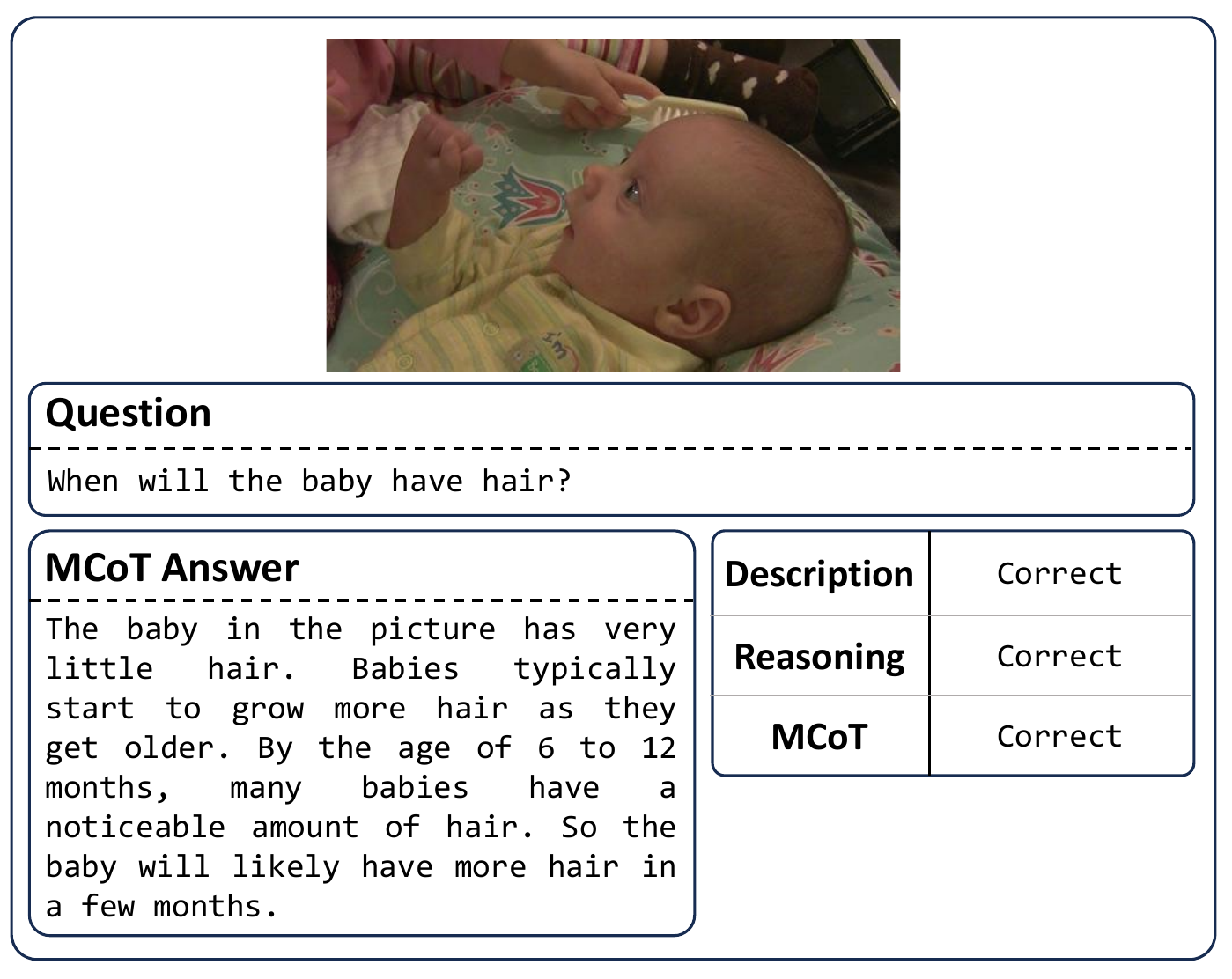}
    \caption{A $\textit{Correctness}_{all}$-based MiCEval metric high-scoring case of MiCEval-\textsc{Normal}.}
\end{figure*}

\begin{figure*}[htpb]
    \centering
    \includegraphics[width=0.9\linewidth]{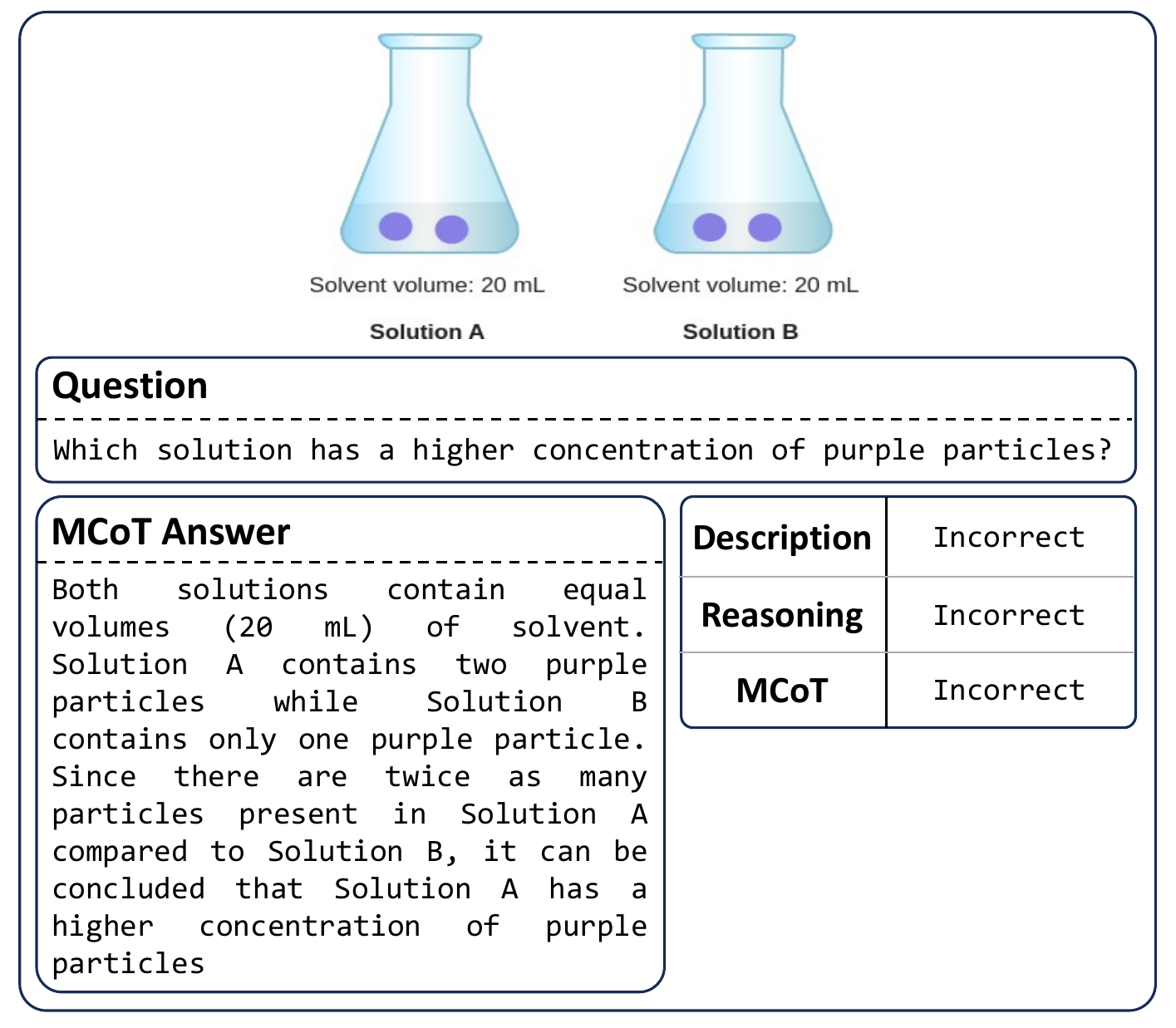}
    \caption{A $\textit{Correctness}_{all}$-based MiCEval metric low-scoring case of MiCEval-\textsc{Hard}.}
\end{figure*}

\begin{figure*}[htpb]
    \centering
    \includegraphics[width=0.9\linewidth]{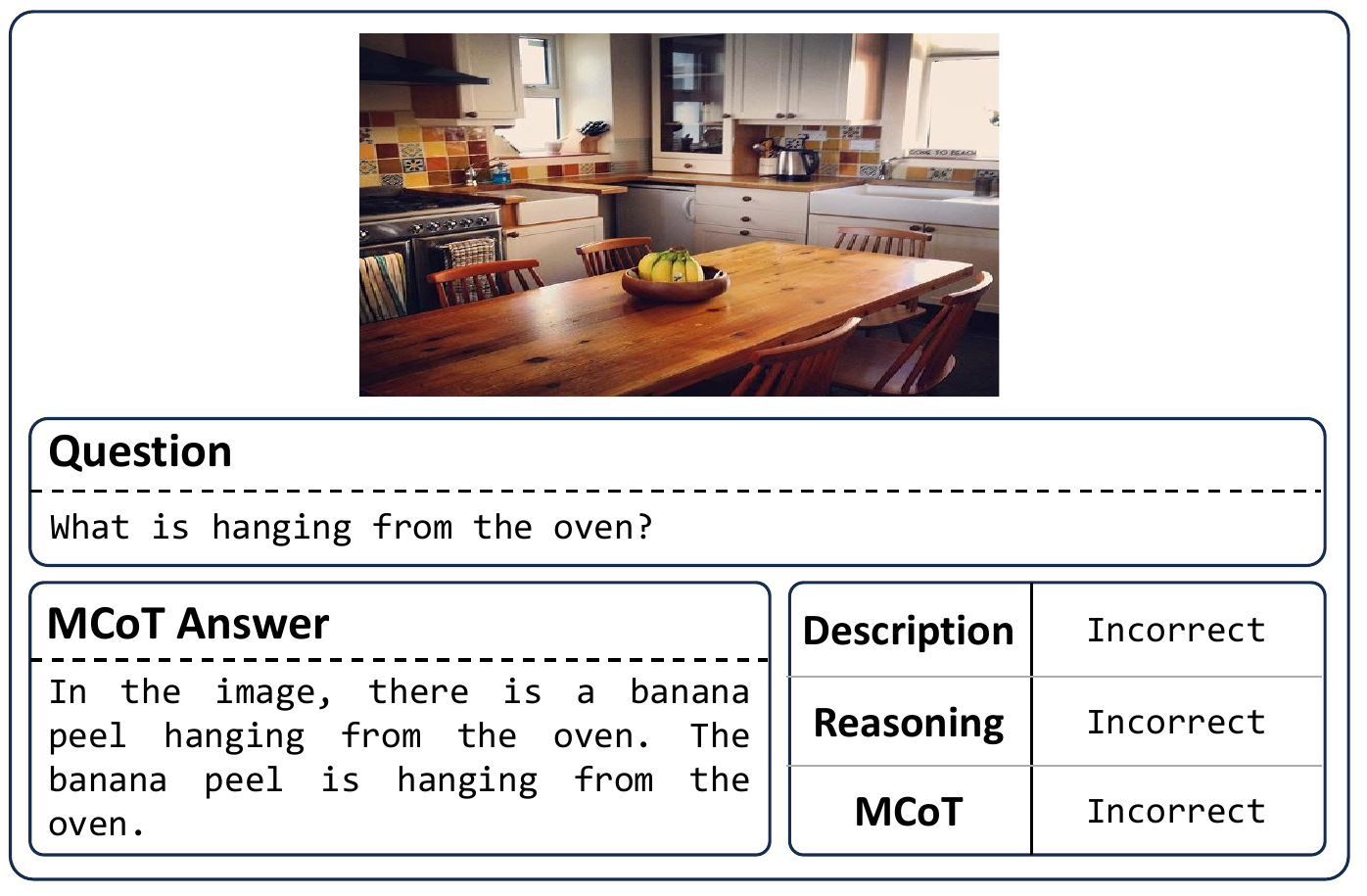}
    \caption{A $\textit{Correctness}_{all}$-based MiCEval metric low-scoring case of MiCEval-\textsc{Normal}.}
\end{figure*}

\end{document}